\PassOptionsToPackage{sort,longnamesfirst}{natbib}
\documentclass[useAMS,usenatbib,onecolumn]{biom}
\let\cite\citep

\usepackage[utf8]{inputenc}
\usepackage[T1]{fontenc}

\newcommand{\SM}[1]{Web Appendix #1}
\usepackage{siunitx}
\usepackage{multirow}
\usepackage{colortbl,etoolbox,siunitx,xcolor}
\usepackage{mathtools}
\usepackage{url}
\usepackage{hyperref}
\usepackage{amsmath}
\usepackage{amsfonts}
\usepackage{arydshln}
\usepackage{float} 
\usepackage{xr}
\definecolor{verylighgray}{gray}{0.85}
\definecolor{navyblue}{RGB}{0,0,128}

\def\multiset#1#2{\ensuremath{\left(\kern-.3em\left(\genfrac{}{}{0pt}{}{#1}{#2}\right)\kern-.3em\right)}}
\newtheorem{proposition}[theorem]{Proposition}%

\begin{document}

\title[]{LaMe: a Joint Latent Mediation Framework for High-Dimensional Systems}

\author{
Ga\"el~Poux-M\'edard$^{1,3,*}$\email{pouxg@iarc.who.int}, 
Kaniav~Kamary$^{2}$, 
Yohann~De~Castro$^{3,4}$, 
and Vivian~Viallon$^{1}$ \\
$^{1}$Nutrition and Metabolism Branch, International Agency for Research on Cancer, WHO, Lyon, France \\
$^{2}$Institut Camille Jordan, INSA Lyon, Centrale Lyon, Universit\'e Lyon 1, CNRS, Villeurbanne, France \\
$^{3}$Institut Camille Jordan, École Centrale Lyon, CNRS UMR 5208, Lyon, France \\
$^{4}$Institut Universitaire de France
}

\begin{abstract}
Mediation analysis seeks to understand how an exposure influences an outcome through intermediate variables. However, as increasingly rich datasets become available, hundreds to thousands of potential mediators make these systems difficult to model and hard to interpret. Latent-variable approaches offer a natural way to accommodate these high-dimensional settings. 
However, existing latent mediation methods face limitations, including overparameterization, biased inference, or limited scalability.
We develop LaMe, a joint linear-Gaussian framework that summarizes a high-dimensional exposure-mediator-outcome system through a small number of latent variables. We identify natural direct and indirect effects under an explicit assumption that the manifest measurements fully capture the latent mediators, and characterize conditions under which the latent representation effectively restricts the observed cross-covariance structure.
Simulations assess recovery of observed association parameters, natural effects, and manifest features associated with latent mediation pathways across a range of challenging settings. Four real-world applications demonstrate competitive predictive performance with compact latent representations, including single-dimensional representations of biomarker and proteomic measurements. 
Together, these results establish LaMe as a practical framework for estimating and summarizing multivariate mediation systems under explicit structural and measurement assumptions.
\end{abstract}

\begin{keywords}
joint modeling; latent mediators; mediation analysis; probabilistic inference
\end{keywords}

\maketitle

\section{Introduction}
Deciphering how complex exposures exert their effects through intermediary processes is a cornerstone of modern epidemiology. In the NHANES data considered here, for example, lifestyle characteristics may directly influence glycohemoglobin levels, but might also affect blood biomarkers which in turn cause these variations. 
However, the continued emergence of large-scale, high-dimensional datasets poses new challenges to established mediation analysis methods. 
Analyses conducted on biological data are particularly affected, as molecular profiling can yield hundreds or thousands of correlated molecular measurements, only a subset of which may be relevant to the mediating process linking an exposure to an outcome.
Latent variable models have been introduced to extend classical mediation models to these settings \cite{Derkach2019HighDimMed,Albert2016LatentMediation,Lindquist2023MLHighDimMediation,Chen2017DirectionsofMediationLatentMed}. This approach brings a number of improvements over non-latent mediation techniques \cite{Zeng2021SurveyMediationTechniques,Zhu2024SurveyMediationModels} (e.g. based on OLS, LASSO or ElasticNet): noise decomposition, information compression, interpretability and robustness of the estimates.

However, existing latent mediation approaches face several limitations. Stepwise methods estimate latent scores separately from subsets of the observed data and use the resulting scores in subsequent regressions on remaining observed variables, potentially inferring latent variables that are irrelevant to mediation systems \cite{Lindquist2023MLHighDimMediation,Chen2017DirectionsofMediationLatentMed}. Joint approaches avoid this issue by estimating latent variables along with their relation to observed variables, but may be overparameterized, prone to overfitting, limited in scalability, and typically assume univariate exposures, univariate outcomes, and the conditional independence among manifest variables given the latent mediators whereas biological measurement are generally correlated \cite{Derkach2019HighDimMed,Albert2016LatentMediation}. 
To address these limitations, we introduce the \textbf{Latent Mediation (LaMe)} framework, which jointly models multivariate exposures, outcomes and high-dimensional manifest variables through a small number of latent mediators without assuming diagonality of noise covariances, while explicitly accounting for direct exposure-outcome effects.

This approach makes three main methodological contributions. 
First, LaMe is novel in that it infers latent mediators from a joint probabilistic model with multivariate exposures and outcomes, high-dimensional manifest variables, and non-diagonal residual covariance matrices. As a result, it avoids biases associated with estimating latent scores separately from the rest of the system as in stepwise procedures and estimates all its parameters based on all available information rather than subsets of observed variables. We characterize when the representation imposes a genuine restriction on cross-block relationships, distinguishing dimension reduction from a reparameterization of unrestricted regression.
Second, LaMe connects joint probabilistic estimation methods \cite{Derkach2019HighDimMed,Albert2016LatentMediation} with representations in which latent mediators are exactly recoverable from the manifest measurements \cite{Lindquist2023MLHighDimMediation,Chen2017DirectionsofMediationLatentMed,Zhao2020SparsePCA}. Under this explicit restriction and suitable rank conditions, the aggregate natural direct and indirect effects are statistically identified from the observed data. Throughout this article, identifiability refers to statistical identification from the observed-data distribution, rather than causal identification of natural effects. We also introduce a score to assess which manifest variables are associated with exposures and informative about outcomes through the latent mediators, independently of how the latent variables are represented.
Third, we derive closed-form EM updates for two equivalent likelihood factorizations, allowing computational efficiency to be improved when manifest measurements and exposures have substantially different dimensions. Simulations assess parameter recovery, effect recovery under the exact recoverability assumption, and feature ranking; real-world applications assess predictive performance and the compactness of the learned representations. These results position LaMe as a useful methodological option.



\begin{figure}
    \centering
    \includegraphics[width=\textwidth]{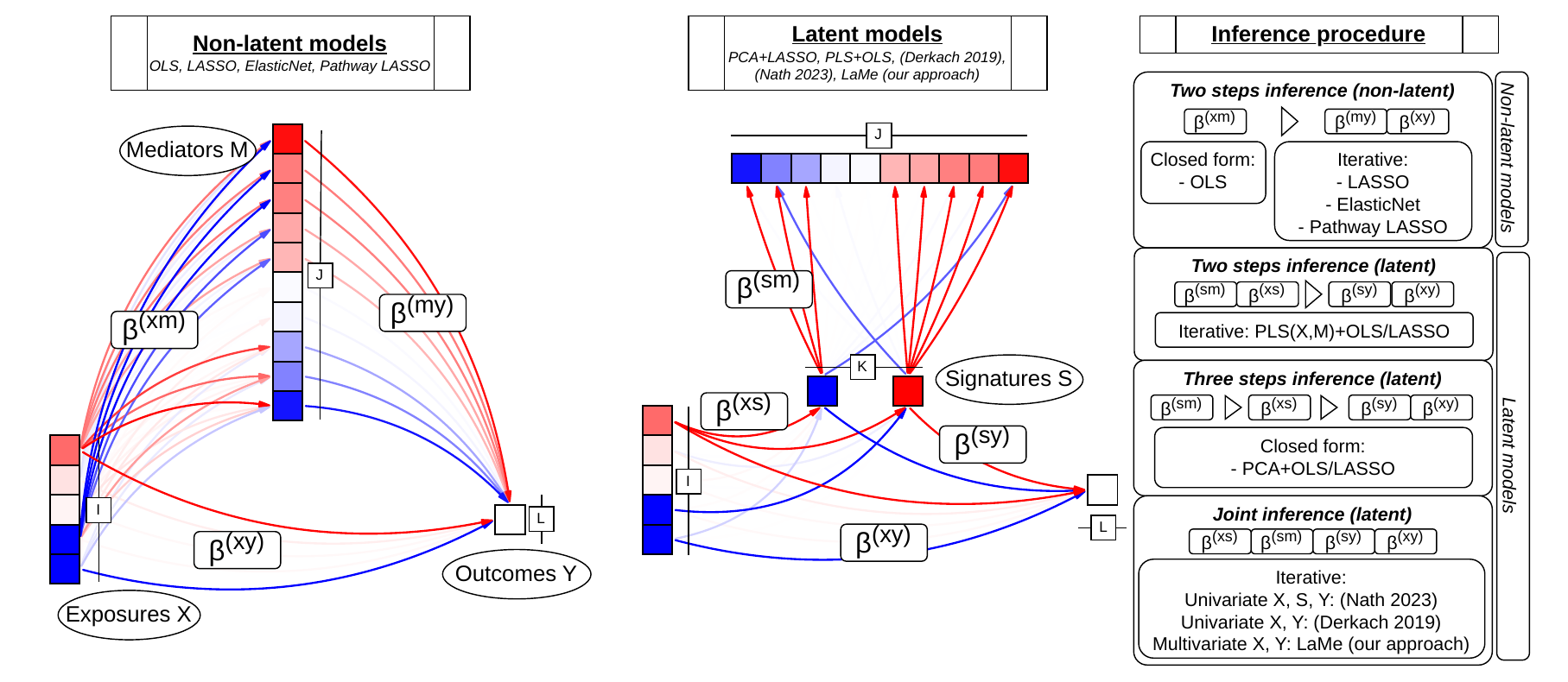}
    \caption{\textbf{Non-latent (left-panel) and latent (middle-panel) mediation systems, and overview of several associated inference procedures (right-panel)} --- (left and middle panels) Variables $X$, $M$, $S$ and $Y$ are specified in the ovals, structural coefficients matrices $\beta$ are shown in rounded rectangles, and the system's dimensions are given in white squares. Arrows represent causal relations between variables. For variables with a causal path to the outcome, squares' color represents the total causal effect of a unit intervention on the corresponding variable on the outcome~$Y$ under the model specified in Eq.~\eqref{eq:latent-mediation-equations}--in this case, the sum of all causal paths leading to $Y$; in the latent graph, the colors of the manifest-variable squares represent their conditional associations with $Y$, not causal effects. For both squares and arrows, blue represents a negative relation, red represents a positive relation, and color intensity represents its strength. (right-panel) Common optimization approaches to mediation systems and the order in which they infer parameters.
    Alt text: three-panel schematic comparing non-latent and latent mediation systems and their inference strategies. Exposures influence outcomes directly and through latent mediators; red and blue arrows indicate positive and negative relationships. The right panel groups methods by two-step, three-step, or fully joint estimation.
    }
    \label{fig-mediation-models}
\end{figure}

\section{Background}
Let $N>1$ be the number of observations and, for each observation, let $I\geq 1$ be the number of exposures, $J\geq 1$ the number of manifest variables, and $L\geq 1$ the number of outcomes. Then, let $X\in \mathbb{R}^{N \times I}$ denote the exposure matrix, $M\in \mathbb{R}^{N \times J}$ the manifest variables matrix and $Y\in \mathbb{R}^{N \times L}$ the outcome matrix. 

\subsection{Non-latent models}
In their simplest form, linear mediation models are specified by the following set of structural causal equations 
(see Fig.~\ref{fig-mediation-models}-left):
\begin{align}
    \label{eq:mediation-equations}
    \left\{
    \begin{aligned}
    X &= \varepsilon_x \\
    M &= X \beta^{(xm)} + \varepsilon_m \\
    Y &= X \beta^{(xy)} + M \beta^{(my)} + \varepsilon_y
    \end{aligned} \right.
\end{align}
The random vectors $\varepsilon_{x,n}$, $\varepsilon_{m,n}$ and $\varepsilon_{y,n}$ corresponding to the $n^{th}$ observation are usually assumed to be mean-zero Gaussians and mutually independent. Formally, for each observation $n$, $\varepsilon_{x,n} \sim \mathcal{N}(0,\Sigma_{x})$, $\varepsilon_{m,n} \sim \mathcal{N}(0,\Sigma_{m|x})$ and $\varepsilon_{y,n} \sim \mathcal{N}(0,\Sigma_{y|x,m})$; these terms account for noise and/or exogenous variables. Coefficient matrices are~$\beta^{(xm)} \in \mathbb{R}^{I \times J}, \beta^{(xy)} \in \mathbb{R}^{I \times L}, \beta^{(my)} \in \mathbb{R}^{J \times L}$ and noise covariances are~$\Sigma_{m|x} \in \mathbb{R}^{J \times J}, \Sigma_{y|x,m} \in \mathbb{R}^{L \times L}$. We consider that all variables are centered. 
As shown in \SM{\ref{sec-SM:observed-con}}, the inclusion of confounders $C$ between $X$, $M$ and $Y$ is straightforward and is therefore omitted here for clarity.

According to recent surveys \cite{Rijnhart2021MediationInResearch,Zhu2024SurveyMediationModels,Zeng2021SurveyMediationTechniques}, most high-dimensional mediation analyses published in the literature rely on non-latent linear mediation models, under which the parameters of Eq.~\eqref{eq:mediation-equations} are estimated using sequential regression procedures combined with penalization. Classical mediation analysis is based on ordinary least squares (OLS) estimation of $\beta^{(xm)}$, $\beta^{(my)}$ and $\beta^{(xy)}$ 
, while extensions to high-dimensional settings incorporate regularization to select the relevant mediators or improve estimation or prediction accuracy. Typical examples include the LASSO,
Ridge 
and ElasticNet regressions. Another example is the Pathway LASSO model \cite{Zhao2021PathwayLASSO}, which introduces a nonlinear penalty targeting entire indirect paths rather than individual coefficients. 
These approaches allow for straightforward coefficient estimation, quickly converge to global optima, and provide readily interpretable results. On the other hand, they heavily rely on hyperparameter tuning and may yield unstable and selection-inconsistent estimates in the presence of strongly correlated mediators or high-dimensional regimes, where mediator selection can become effectively arbitrary
\cite{Wang2018PrecisionLASSO,Ye2021LASSOVarSelectionHighDim}. 

\subsection{Latent models}
\subsubsection{Latent mediators}
\paragraph{Structural equations} To overcome some of these limitations, latent models have been suggested, which assume the existence of unobserved variables, denoted by $S \in \mathbb{R}^{N \times K}$, where $K$ is the number of latent dimensions. Following the latent mediation formulations of \cite{Derkach2019HighDimMed,Albert2016LatentMediation}, LaMe is specified by the following structural equations (see Fig.~\ref{fig-mediation-models}-middle):
\begin{align}
    \label{eq:latent-mediation-equations}
    \left\{
    \begin{aligned}
    X &= \varepsilon_x \\
    S &= X \beta^{(xs)} + \varepsilon_s \\
    M &= S \beta^{(sm)} + \varepsilon_m \\
    Y &= X \beta^{(xy)} + S \beta^{(sy)} + \varepsilon_y
    \end{aligned} \right.
\end{align}
The coefficient matrices of the model are $\beta^{(xs)} \in \mathbb{R}^{I \times K}$, $\beta^{(sm)} \in \mathbb{R}^{K \times J}$, $\beta^{(xy)} \in \mathbb{R}^{I \times L}$, $\beta^{(sy)} \in \mathbb{R}^{K \times L}$. The noise vectors $\varepsilon_{x,n}, \varepsilon_{s,n}, \varepsilon_{m,n}$ and $\varepsilon_{y,n}$ corresponding to the $n^{th}$ observation are assumed to be mean-zero Gaussians and mutually independent. Formally, for each observation $n$, $\varepsilon_{x,n} \sim \mathcal{N}(0,\Sigma_{x})$, $\varepsilon_{s,n} \sim \mathcal{N}(0,\Sigma_{s|x})$, $\varepsilon_{m,n} \sim \mathcal{N}(0,\Sigma_{m|s})$ and $\varepsilon_{y,n} \sim \mathcal{N}(0,\Sigma_{y|x,s})$, for some noise covariance matrices $\Sigma_{x} \in \mathbb{R}^{I \times I}$, $\Sigma_{m|s} \in \mathbb{R}^{J \times J}$, $\Sigma_{s|x} \in \mathbb{R}^{K \times K}$, $\Sigma_{y|x,s} \in \mathbb{R}^{L \times L}$.

To simplify the exposition, we omit observed confounders (say $C$) from the subsequent derivations. As shown in \SM{\ref{sec-SM:observed-con}}, the inclusion of confounders between $X$, $S$ and $Y$ is straightforward and requires only replacing the exposure variable with a composite variable $\xi:= [X,C]$, and appropriately partitioning the estimated coefficients corresponding to $X$ and $C$.

\paragraph{Natural effects} 
Under sequential ignorability, namely $\{ Y(x,s), S(x') \} \perp X | C=c$ and $Y(x,s) \perp S(x') | X=x', C=c$ where $Y(x,s)$ and $S(x')$ denote the counterfactual outcome and mediator under the corresponding interventions, the natural direct and indirect effects associated with a one-unit increase in the exposures are given by \cite{Derkach2019HighDimMed,Albert2016LatentMediation}:
\begin{align}
    \label{eq:NIE-NDE-expression}
    NDE = \beta^{(xy)} \qquad ; \qquad NIE = \beta^{(xs)}\beta^{(sy)},
\end{align}

\subsubsection{Stepwise approaches}
Stepwise approaches to latent mediation generally define $S$ as a deterministic linear combination of manifest variables. These methods typically derive latent scores from an initial model, for instance from Principal Component Analysis (PCA), Partial Least Squares (PLS) or their variants; then, these scores enter as observed mediators in the subsequent estimation of a non-latent mediation model \cite{Zhao2020SparsePCA,Kawano2018TwoStepsPCAPCRMediation,Cai2023TwoStepsLatentMediation,Hu2024TwoStepsLatentMediation}.

However, most existing stepwise approaches do not use all available data to derive their latent scores. For instance, approaches based on PCA optimize $S$ to explain the variance of $M$, and approaches based on PLS optimize $S$ to explain the covariance between $X$ and $M$, or $M$ and $Y$, regardless of whether the resulting scores best characterize mediation between $X$ and $Y$. 
Conversely, a correctly specified model accounting for the joint variance and covariance of $(X,M,Y)$ does not require latent scores to be selected using only a subset of the observed relationships.

\subsubsection{Joint modeling}
A joint approach that maximizes a likelihood corresponding to the full structural model may therefore be preferable.

The latent approaches of \cite{Chen2017DirectionsofMediationLatentMed,Lindquist2023MLHighDimMediation} both consider an univariate exposure and outcome, and both estimate the first latent direction and the associated path coefficients from a joint objective based on all observed data.
In \cite{Chen2017DirectionsofMediationLatentMed}, the model extracts subsequent directions through approximate deflation while keeping earlier directions fixed. Consequently, the retained directions generally differ from those estimated by maximizing the joint likelihood of a full multivariate latent model; in particular, the sequential objective does not include cross-score residual covariances. Finally, the authors remark that inference may become cumbersome when $J \gg 1$ without resorting to additional data augmentation techniques.
The model introduced in \cite{Lindquist2023MLHighDimMediation}  extends this approach beyond linear relationships to model $S$ as an arbitrary deterministic function of $M$. However, this extension is limited to a single latent variable. Moreover, score selection and the estimated direct–indirect decomposition are sensitive to the scale of $Y$, as shown in \SM{\ref{sec-SM:LindquistObj}}. Finally, standardization of the univariate score is imposed algorithmically rather than incorporated into the objective function, complicating the analytical characterization of the resulting estimator. 

Joint probabilistic formulations, on the other hand, explicitly account for invariance to latent rescaling. In \cite{Albert2016LatentMediation}, the authors formulate a joint probabilistic latent model with a single latent mediator. The approach was later extended by \cite{Derkach2019HighDimMed} to handle multivariate latent mediators.
In that work, the authors assume a univariate exposure and outcome, and constrain the conditional noise covariance of manifest variables given latent mediators $\Sigma_{m|s}$ to be diagonal. 
In this setting, however, we show in \SM{\ref{sec-SM:DerkachFA}} that the non-regularized and correctly specified model from \cite{Derkach2019HighDimMed} is equivalent to Factor Analysis of $M$ followed by linear regressions on the resulting $S$ when $K>1$. Additionally, the diagonal constraint on $\Sigma_{m|s}$ imposes conditional independence of manifest variables given the latent mediators, which is not necessarily true when $M$ represents biomarkers.
Finally, Condition~1 in \cite{Derkach2019HighDimMed} is required for identification of the effects, but the proposed model does not guarantee that this condition holds.

\section{Methods}
\label{sec:LaMe}
We develop LaMe to extend existing approaches by jointly modeling all observed variables, allowing non-diagonal noise covariance matrices and multivariate exposures, outcomes and latent information bottleneck. Under the exact recoverability assumption for $S$, the framework also permits identification of natural effects.

We first detail the exact recoverability and information bottleneck assumptions made by our approach, then we develop a general inference procedure for an auxiliary model that does not assume exact recoverability, and we finally show in Proposition~\ref{prop:identifiability} that exact recoverability can be recovered through a likelihood-equivalent reparametrization of the model post-estimation. This proposition further establishes that the parametric representations of natural direct and indirect effects are statistically identified under this condition.

\subsection{Assumptions}
\subsubsection{Exact recoverability}
Following score-based mediation approaches, which define latent mediators as linear combinations of manifest variables \cite{Lindquist2023MLHighDimMediation,Chen2017DirectionsofMediationLatentMed,Zhao2020SparsePCA}, we assume the same kind of deterministic relationship holds within our probabilistic framework. 
Specifically, we will work under what we call the exact recoverability assumption, under which $S = \mathbb{E}[S|M] := M \beta^{(ms)}$. 
We emphasize that this equality does not imply that the measurements $M$ are causes of $S$; they remain the manifest variables causally influenced by $S$, with $S$ causally influenced by $X$. 
This requirement is equivalent to:
\begin{align}
    \label{eq:assumption-gauge}
    \Sigma_{m|s} \beta^{(ms)}&=0\\ \notag
    \text{where } 
    \beta^{(ms)} &= \underbrace{((\beta^{(sm)})^T\Sigma_{s}\beta^{(sm)} + \Sigma_{m|s})^{-1}}_{\Sigma_m^{-1}} (\beta^{(sm)})^T \underbrace{(\Sigma_{s|x} + (\beta^{(xs)})^T \Sigma_{x} \beta^{(xs)})}_{\Sigma_s}
\end{align}

We point out that the exact-recoverability approximation becomes increasingly accurate as the number of manifest variables $J$ grows, making it well suited to high-dimensional settings. 
To see this, consider the difference between the unconstrained latent variable $S$ in Eq.~\ref{eq:latent-mediation-equations} and its exactly recoverable counterpart $S^*=M \beta^{(ms)}$. Assuming that additional manifest variables provide non-vanishing, non-redundant information about $S$, the measurement precision $\beta^{(sm)}\Sigma_{m|s}^{-1}(\beta^{(sm)})^T$ grows linearly with $J$. Then, we show in \SM{\ref{sec-SM:exact-recov-limit}} that $\mathbb{E}[||S^* - S||^2] = \mathcal{O}(J^{-1})$, meaning that $S \overset{J \to \infty}{\to} S^*$.
Thus, as the number of informative manifest variables increases, the latent variable $S$ of the unconstrained model in Eq.~\ref{eq:latent-mediation-equations} asymptotically converges to the latent mediator that is exactly recoverable from $M$.
Although exact recoverability is an idealization for finite $J$, it therefore becomes a reasonable approximation when sufficiently many informative manifest variables are available, supporting its use in high-dimensional settings.

We show in Proposition~\ref{prop:identifiability} below that the joint model specified in Eq~\eqref{eq:latent-mediation-equations} can always be formulated to satisfy exact recoverability.

\subsubsection{Information bottleneck}
Introducing low-dimensional latent variables addresses several limitations of non-latent least-squares mediation models. When the number of observations is finite, sampling noise can introduce spurious association patterns; by imposing a low-rank structure on the cross-covariance between $X$ and $M$, and between $M$ and $Y$, latent models alleviate this by forcing the model to focus on the most relevant covariation patterns, typically preventing overfitting the noise and improving robustness and generalization of structural parameters $\beta^{(xs)}, \beta^{(sm)}, \beta^{(xy)}, \beta^{(sy)}$. In this sense, the latent representation acts as an information bottleneck between observed variables \cite{Chechik2005InformationBottleneck}. 
Likewise, information bottlenecks can suppress weaker associations between observed variables, allowing the system to be approximated by a simplified representation that only captures its most defining patterns of covariation, which is a recurring objective of dimension-reduction approaches to mediation analysis.

Under the assumptions from Eq.~\eqref{eq:latent-mediation-equations} (linear relations, full noise covariances and mutually independent noises), a necessary and sufficient condition for an information bottleneck can be formulated as (see \SM{\ref{sec-SM:bottleneck-condition-derivation}} for details):
\begin{align}
    \label{eq:K-information-bottleneck}
    K <\, &\mathrm{rank}\left(\begin{aligned}\mathrm{Cov}(X,M)\\\mathrm{Cov}(Y,M|X)\end{aligned}\right)
    &\text{where } &\mathrm{rank}\left(\begin{aligned}\mathrm{Cov}(X,M)\\\mathrm{Cov}(Y,M|X)\end{aligned}\right) \leq \min(I+L, J),
\end{align}
where $\mathrm{Cov}(\cdot,\cdot)$ denotes an observed covariance. 
Conversely, $K \geq \min(J, I+L)$ is a sufficient condition ensuring that every possible cross-covariance matrix can be represented exactly. Then, the observable regression coefficients between $X$, $M$ and $Y$ estimated from the latent model in Eq.~\eqref{eq:latent-mediation-equations} are a reparametrization of those estimated by a non-latent OLS model, as shown in \SM{\ref{sec-SM:bottleneck-condition-derivation}}, limiting the interest of the approach.
Note that assuming a diagonal conditional noise covariance $\Sigma_{m|s}$ does not affect the condition above.
In practice, the latent dimension $K$ can be selected via cross-validation or information criteria such as AIC or BIC to balance model complexity and goodness-of-fit.

\subsection{LaMe model}


\subsubsection{Likelihood}
Starting from Eq.~\eqref{eq:latent-mediation-equations}, $\theta = \{ \beta^{(xs)}$, $\beta^{(sm)}$, $\beta^{(xy)}$, $\beta^{(sy)}$, $\Sigma_{x}$, $\Sigma_{s|x}$, $\Sigma_{m|s}$, $\Sigma_{y|x,s} \}$ denote the full set of model parameters, and $p(\theta)$ its prior distribution. 
Assuming that mean-zero Gaussian noises are mutually independent in Eq.~\eqref{eq:latent-mediation-equations}, a natural factorization of the joint distribution is
\begin{align}
    \label{eq:joint-distribution-vX}
    p(X, S, M, Y, \theta) 
    &= p(X|\theta) p(S|X,\theta)p(M|S,\theta)p(Y|X,S,\theta) p(\theta)
\end{align}
\noindent
Recall that all the distributions above are Gaussian, with parameters specified as, for example, $p(S|X,\theta) = \mathcal{N}(S;X \beta^{(xs)}, \Sigma_{s|x})$.
In the following derivations, we assume a flat prior $p(\theta)$ on all parameters, so that $p(X, S, M, Y, \theta) \propto p(X, S, M, Y | \theta)$. 
Other priors that regularize the parameters, together with an extension to log-normal variables, are discussed in \SM{\ref{sec-SM:model-features}}.

The latent nature of $S$ prevents direct optimization of~\eqref{eq:joint-distribution-vX}. We can, however, compute the incomplete likelihood by marginalizing the complete likelihood over all possible values of $S$. Using the complete likelihood expression~\eqref{eq:joint-distribution-vX}, a standard Gaussian integration formula and grouping terms together, we find
\begin{align}
    \label{eq:marginal-likelihood}
    p(X,M,Y| \theta) =& \int_S \underbrace{e^{-\frac{1}{2}\langle S^T S, A\rangle+\langle S, B \rangle + C}}_{p(S,X,M,Y| \theta)} dS = e^{\frac{1}{2}\langle B^TB, A^{-1} \rangle} e^{C}(2\pi)^{\frac{NK}{2}}| A |^{-\frac{N}{2}} \\ \notag
    {\rm where\ }A:=& \Sigma_{s|x}^{-1} + \beta^{(sm)}\Sigma_{m|s}^{-1}(\beta^{(sm)})^T + \beta^{(sy)}\Sigma_{y|x,s}^{-1}(\beta^{(sy)})^T \\\notag
    {\rm and\ }B :=& X\beta^{(xs)}\Sigma_{s|x}^{-1} + M\Sigma_{m|s}^{-1}(\beta^{(sm)})^T + (Y- X\beta^{(xy)}) \Sigma_{y|x,s}^{-1} (\beta^{(sy)})^T.
\end{align}
Above, $| A |$ denotes the determinant of $A$ and $\langle A, B \rangle := \sum_{i,j} A_{ij} B_{ij} $ is the Frobenius product. The expression of the normalization term $C$ is provided in \SM{\ref{sec-SM:full-Mstep}}.

The expectation-maximization algorithm then first computes the expectation of latent variables given the current parameter estimates $\theta^{(t)}$ (E-step), and then derives the new optimal parameter set $\theta^{(t+1)}$ given the expectation of latent variables computed in the previous step (M-step). This procedure is iterated until convergence.

\subsubsection{E-step}
Given the parameter estimates at the t$^{th}$ iteration, $\theta^{(t)}$, the E-step computes the conditional distribution of the latent variables $S$ and evaluates expectations of functions of $S$ with respect to this distribution.

We show in \SM{\ref{sec-SM:full-Estep}} that, for any candidate distribution $q_{\theta^{(t)}}(S)$, the observed log-likelihood is lower-bounded as:
\begin{align}
    \label{eq:likelihood-Estep}
    \log p(X,M,Y| \theta)& \geq \int_S q_{\theta^{(t)}}(S) \log \frac{p(S,X,M,Y| \theta)}{q_{\theta^{(t)}}(S)}dS
\end{align}

Then, it can be shown that this lower-bound is equal to $\log p(X,M,Y| \theta^{(t)})$, and therefore maximal, when:
\begin{align}
    \label{eq:E-step}
    q_{\theta^{(t)}}(S) &= p(S|X,M,Y,\theta^{(t)}) = \frac{p(S,X,M,Y| \theta^{(t)})}{p(X,M,Y| \theta^{(t)})}
\end{align}
A full derivation of this result can be found in \SM{\ref{sec-SM:full-Estep}}. 

We finally use $q_{\theta^{(t)}}(S)$ to compute the moments of $S$ under the current parameter estimates before proceeding to the M-step, recalling that $N$ is the number of observations:
\begin{align}
    \label{eq:moments-S}
    \overline{S}^{(t)} &= 
    B A^{-1} \quad; \quad 
    \overline{S^T S}^{(t)} = 
    N A^{-1} + A^{-1}B^TBA^{-1}
\end{align}

\subsubsection{M-step}
The M-step then computes the parameter set $\theta^{(t+1)}$ that maximizes the lower-bound in Eq.~\eqref{eq:likelihood-Estep} with $q_{\theta^{(t)}}(S)$ from Eq.~\eqref{eq:E-step}.
Taking $\beta^{(sm)}$ as an illustration:
\begin{align}
    \label{eq:example-Mstep}
    \partial_{\beta^{(sm)}} \int_S q_{\theta^{(t)}}(S) \log p(S,X,M,Y| \theta)dS &= - \overline{S^T S}^{(t)} \beta^{(sm)}\Sigma_{m|s}^{-1} + (\overline{S}^{(t)})^T M\Sigma_{m|s}^{-1} = 0 \\
    \notag
    \notag
    \Leftrightarrow   \beta^{(sm)}&= (\overline{S^T S}^{(t)})^{-1} (\overline{S}^{(t)})^T M
\end{align}
where $\overline{S}^{(t)}$ and $\overline{S^T S}^{(t)}$ are obtained from Eq.~\eqref{eq:moments-S}. 
Detailed derivations for other parameters are provided in \SM{\ref{sec-SM:full-Mstep}}.

The EM algorithm alternates between the E-step, where $\overline{S}^{(t)}$ and $\overline{S^T S}^{(t)}$ are computed given the current parameter estimates $\theta^{(t)}$, and the M-step, where the parameter estimates are updated to $\theta^{(t+1)}$ given $\overline{S}^{(t)}$ and $\overline{S^T S}^{(t)}$, until convergence of the incomplete likelihood. The EM algorithm guarantees a non-decreasing likelihood and convergence to a stationary point \cite{Neal1998ConvergenceEM}. Note that although we consider all parameters $\theta$ to be random variables, the EM optimization algorithm yields point estimates rather than their distribution. 

At this point, $S$ may not be exactly recoverable from $M$ as assumed in Eq.~\eqref{eq:assumption-gauge}. In the next section, we show how to satisfy this assumption through a likelihood-equivalent reparametrization of the EM solution.

\subsection{Identifiability}
The parametric representation of the observed moments recalled in \SM{\ref{sec-SM:bottleneck-condition-derivation}} shows that the observed likelihood in Eq.~\eqref{eq:marginal-likelihood} is invariant to some transformations of the latent space; the exact recoverability property limits these to invertible linear transformations, which makes natural effects identifiable.

\begin{proposition}[Statistical identification of the natural effects]
\label{prop:identifiability}
The natural direct and indirect effect through $S$ as defined in Eq.~\eqref{eq:NIE-NDE-expression}, namely $\beta^{(xy)}$ and $\beta^{(xs)}\beta^{(sy)}$, are uniquely identified from the joint distribution of $(X,M,Y)$ under the exact recoverability assumption $S=\mathbb{E}[S|M]$ when the bottleneck condition from Eq.~\eqref{eq:K-information-bottleneck} is satisfied.
\end{proposition}
\begin{proof}[(Proposition~\ref{prop:identifiability})]
\paragraph{Likelihood-equivalent reparametrizations}
Let $H_M \in \mathbb{R}^{K \times K}$ and $H_Y \in \mathbb{R}^{K \times K}$ denote invertible matrices. Then, we define the elements of a set of transformed parameters denoted by $\theta^*$ expressed in terms of $\theta$, $H_M$ and $H_Y$ as:
\begin{align}
    \label{eq:signal-noise-invariance}
    \beta^{(xs)}_{*} = \beta^{(xs)} H_M  \quad &; \quad 
    \Sigma_{s|x}^* = H_M^{T} \Sigma_{s|x} H_Y  \\  \notag
    \beta^{(sm)}_{*} = H_M^{-1} \beta^{(sm)} \quad &; \quad 
    \Sigma_{m|s}^* = \Sigma_{m|s} + (\beta^{(sm)})^T \Sigma_{s|x} \beta^{(sm)} - (\beta^{(sm)}_{*})^T \Sigma_{s|x}^* \beta^{(sm)}_{*}  \\  \notag
    \beta^{(sy)}_{*} = H_Y^{-1} \beta^{(sy)} \quad &; \quad 
    \Sigma_{y|x,s}^* = \Sigma_{y|x,s} + (\beta^{(sy)})^T \Sigma_{s|x} \beta^{(sy)} - (\beta^{(sy)}_{*})^T \Sigma_{s|x}^* \beta^{(sy)}_{*}  \\  \notag
    \beta^{(xy)}_{*} = \beta^{(xy)} + \beta^{(xs)}& \beta^{(sy)} - \beta^{(xs)}_{*} \beta^{(sy)}_{*} \quad ; \quad 
    \Sigma_x^* = \Sigma_x   
\end{align}
Finally, in the absence of additional assumptions, the relation $P(X,M,Y | \theta) = P(X,M,Y|\theta^*)$ holds. We demonstrate in \SM{\ref{sec-SM:exhaustive-reparam}} that no other likelihood-equivalent reparametrizations exist when the bottleneck condition from Eq.~\eqref{eq:K-information-bottleneck} is satisfied.


\paragraph{Parametrizations with exact recoverability}
From the definition in Eq.~\eqref{eq:assumption-gauge} and noting that $\Sigma_{m}^* = \Sigma_{m}$ (see \SM{\ref{sec-SM:add-deriv-id}}), we derive an expression of $H_M H_Y^{-1}$ such that they define a reparametrization that satisfies the exact recoverability assumption, meaning $\Sigma_{m|s}^* \Sigma_{m}^{-1} (\beta^{(sm)}_*)^T \Sigma_{s}^* = 0$, as:
\begin{align*}
    H_M H_Y^{-1} &= \left( (\beta^{(sm)}\Sigma_{m}^{-1} (\beta^{(sm)})^T)^{-1} - (\beta^{(xs)})^T \Sigma_{x} \beta^{(xs)} \right)^{-1} \Sigma_{s|x}
\end{align*}
Note that the relation above is defined as long as $\beta^{(sm)}\Sigma_{m}^{-1} (\beta^{(sm)})^T$ is non-singular, which is necessarily true when the bottleneck condition from Eq.~\eqref{eq:K-information-bottleneck} is satisfied.
We demonstrate in \SM{\ref{sec-SM:add-deriv-id}} that the expression above also guarantees the positive semi-definiteness of all transformed noise covariances $\Sigma_{s|x}^*$, $\Sigma_{m|s}^*$ and $\Sigma_{y|x,s}^*$.
Therefore, any choice of an invertible $H_M$, along with the corresponding $H_Y$ from the equation above, guarantees the assumption from Eq.~\eqref{eq:assumption-gauge} is satisfied.

\paragraph{Invariance of natural effects under exact recoverability}
Note that $\beta^{(sm)}\Sigma_{m}^{-1} (\beta^{(sm)})^T = \Sigma_{s}^{-1}$ (see \SM{\ref{sec-SM:add-deriv-id}}) when the assumption in Eq.~\eqref{eq:assumption-gauge} is verified.
Injecting it in the expression above yields $H_M H_Y^{-1}=I$.
Identification of the natural effects coefficients under the exact recoverability assumption immediately follows:
\begin{align*}
    \beta^{(xy)}_* = \beta^{(xy)} + \beta^{(xs)} (I-  H_M H_Y^{-1}) \beta^{(sy)} = \beta^{(xy)} ; \quad  \beta^{(xs)}_* \beta^{(sy)}_* = \beta^{(xs)} H_M H_Y^{-1} \beta^{(sy)} = \beta^{(xs)} \beta^{(sy)}
\end{align*}
\end{proof}
We point out that assumptions other than exact recoverability may also make the NDE and NIE identifiable, although they may not provide similar convergence guarantees when $J \to \infty$; for instance, alternative choices may allow $S$ to be a noisy function of $M$ rather than deterministic. This question lies outside the scope of the present article and is left for future work.

\subsection{Alternative factorization}
The causal model specified in Eq.~\eqref{eq:latent-mediation-equations} implies two conditional independence relationships, $M \perp X | S$ and $Y \perp M | X,S$, which allow the joint distribution of $(X,S,M,Y)$ to be factorized equivalently in two ways, denoted \textbf{LaMe-vX} and \textbf{LaMe-vM} hereafter:
\begin{align}
    \notag
    p(X, S, M, Y | \theta) &= p(X|\theta) p(S|X, \theta)p(M|S, \theta)p(Y|X, S, \theta) \quad \text{(LaMe-vX)} \\
    \label{eq:joint-distribution-vM}
    &= p(M|\theta) p(S|M, \theta)p(X|S, \theta)p(Y|X, S, \theta) \quad \text{(LaMe-vM)}
\end{align}
Both factorizations describe the same family of joint distributions and can therefore be used interchangeably for parameter estimation, with parameters from either representation recoverable from the other.
Note that this reformulation is not necessarily equivalent in the presence of regularization, orthogonality, or structural constraints --for instance, a diagonal conditional noise covariance in \cite{Derkach2019HighDimMed} prevents such equivalence. However, it holds for any reparametrization based on Eq.~\eqref{eq:signal-noise-invariance}, including those assuming exact recoverability.

A first remark is that alternative factorizations may not correspond to any relevant causal system. For instance, the coefficient $\beta^{(sx)}$ in LaMe-vM has no causal interpretation in a mediation analysis setting; however, as shown in \SM{\ref{sec-SM:Inverting-parameters}}, it can be used to retrieve the relevant causal path coefficient $\beta^{(xs)}$. 
Second, structural parameters differ depending on the factorization. 
Third, optimizing the joint model using one factorization instead of the other can make the EM algorithm orders of magnitude faster and the estimation more accurate when $J \gg I$: using the LaMe-vM factorization, $\Sigma_{x|s}$ and $\Sigma_{s|m}$ must be updated ($\mathcal{O}(NI(K+I)+NK(K+J))$) at each iteration whereas $\Sigma_{m}$ needs to be updated ($\mathcal{O}(NJ^2)$) only once. We numerically support this result in \SM{\ref{sec-SM:scalability}}.
The full table of the update equations for both models is available in \SM{\ref{sec-SM:full-Mstep}}.



\section{Results}

For both the LaMe-vX and LaMe-vM factorizations (see Eq.~\eqref{eq:joint-distribution-vM}), we stop the EM loop once the relative change in the log-likelihood remains below $10^{-5}$ for ten consecutive iterations. This procedure is repeated 50 times with different random initializations, and the run with the highest likelihood is selected. We consider both unregularized models and models with an $\ell_1$ penalty on the regression coefficients linking $S$ and $M$.

\subsection{Baselines}
We evaluate LaMe against 
\textbf{OLS}, 
OLS with $\ell_1$ regularizer (\textbf{LASSO}), 
OLS with $\ell_1$ and $\ell_2$ regularizers (\textbf{ElasticNet}), 
PCA on the manifest variables $M$ followed by OLS or LASSO on the resulting ($X,S,Y$) latent mediation system (\textbf{PCA+OLS} and \textbf{PCA+LASSO}), 
PLS on $X \rightarrow M$ followed by OLS on $[X, S] \rightarrow Y$ (brackets denote column-wise concatenation) (\textbf{PLS(X,M)+OLS}), 
PLS on $X \rightarrow [M, Y]$ followed by OLS on $X \rightarrow Y_{res}$ (\textbf{PLS(X,M,Y)+OLS}), 
OLS on the direct path $X \rightarrow Y$ followed by PLS on the indirect path $X \rightarrow [M, Y_{res}]$ (\textbf{OLS+PLS(X,M,Y)}) (where $Y_{res}$ are the residuals of $Y$ that are not explained by the model used in the first step), 
an over-parametrized version of LaMe-vX (\textbf{OVP-LaMe}), and 
an adaptation of the latent mediation model described in \cite{Lindquist2023MLHighDimMediation} (Joint Alternate Optimization - Gaussian case \textbf{JAOG}). 
A discussion of our adaptation of \cite{Lindquist2023MLHighDimMediation} (originally designed for univariate latent mediators) and why we consider OVP-LaMe rather than the model of \cite{Derkach2019HighDimMed} is provided in \SM{\ref{sec-SM:BaselinesDerkLind}}. 

All experiments were run on an Intel Core Ultra 7 165U CPU; a scalability comparison is provided in \SM{\ref{sec-SM:scalability}}.

\begin{figure}
    \centering
    \includegraphics[width=\textwidth]{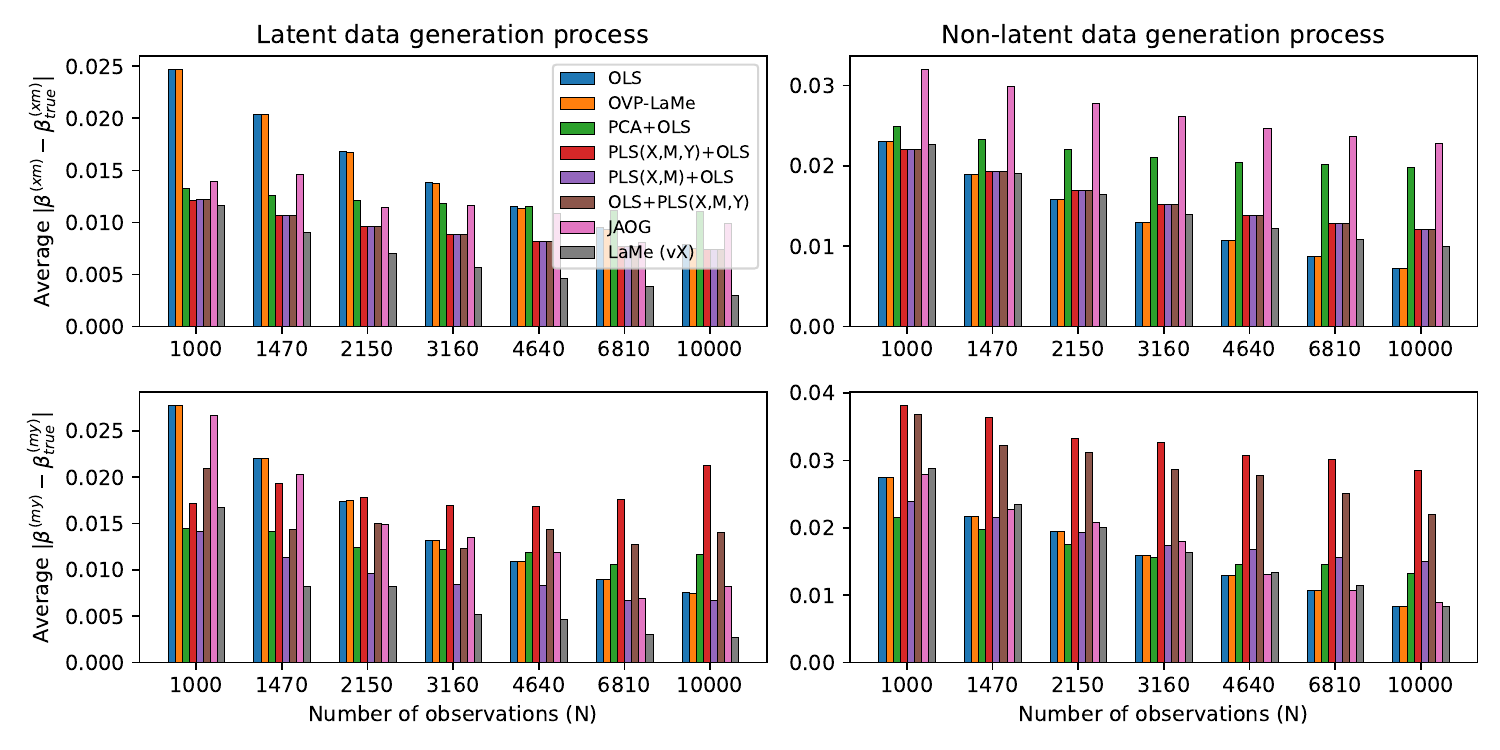}
    \caption{\textbf{LaMe accurately retrieves underlying parameters} --- LaMe and baselines when $I=100$, $J=100$, $L=1$, and a signal-to-noise ratio from $S$ to $M$ and $Y$ equal to $0.1$ for varying $N$. \textbf{(Top)} Mean absolute error of the estimated coefficients $\beta^{(xm)}$ relative to the true coefficients $\beta^{(xm)}_{true}$ | \textbf{(Bottom)} Mean absolute error of the estimated coefficients $\beta^{(my)}$ relative to the true coefficients $\beta^{(my)}_{true}$ | \textbf{(left)} Data generated using a latent process ($K=10$, see Eq.~\eqref{eq:latent-mediation-equations}) with the true $K$ provided to all latent models | \textbf{(right)} Data generated using a non-latent process (see Eq.~\eqref{eq:mediation-equations}) with $K=\frac{\min(J, I+L)}{2}$ for all latent models except for OVP-LaMe where $K=\frac{I+J+L}{2}=100$.
    Alt text: on latent data, LaMe's estimates decrease with $N$ toward smaller mean absolute errors on the true parameters whereas other models do not for $N \in [1000 ; 10000]$; on non-latent data, estimates from OLS, OVP-LaMe, LaMe and JAOG decrease with $N$ toward a smaller mean absolute error on $\beta^{(my)}_{true}$ whereas other models do not for $N \in [1000 ; 10000]$.
    }
    \label{fig-XP2}
\end{figure}

\subsection{Synthetic data}
\subsubsection{Data generation}
Synthetic variables $X, M, Y$ are generated according to the linear-Gaussian structural equations specified by Eq.~\eqref{eq:latent-mediation-equations} when data are assumed to stem from a latent model, and according to Eq.~\eqref{eq:mediation-equations} when we assume the absence of a shared latent structure. The dimensions $I,J,L$ and possibly $K_{true}$ are specified for each experiment below. 
Noise covariances are rescaled so the following signal-to-noise ratios (SNR) $SNR(S|X), SNR(M|S), SNR(Y|X,S)$ (in the latent case) and $SNR(M|X), SNR(Y|X,M)$ (in the non-latent case) attain a given value (e.g. $SNR(M|S)=\frac{\operatorname{Var}(S \beta^{(sm)})}{\operatorname{Var}(\mathcal{N}(0,\Sigma_{m|s}))}=\frac{25}{100}$ means $20\%$ of the variations in $M$ are explained by the signal from $S$, and $80\%$ are due to random noise; see Eq.~\eqref{eq:latent-mediation-equations} and \SM{\ref{sec-SM:Force-SNR}}).

\subsubsection{Recovering underlying parameters}
To compare LaMe against baselines that do not assume the existence of latent mediators (e.g. OLS and OVP-LaMe), we first evaluate the capacity of each model to recover the true marginal association between observed variables $X$ and $M$ ($\beta^{(xm)}:=\nabla_X \mathbb{E}[M|X]$) and between $M$ and $Y$ ($\beta^{(my)}:=\nabla_M \mathbb{E}[Y|M]$). Not only do these coefficients provide a common basis for comparing models, they are also involved in the computation of the natural effects under the exact recoverability assumption (i.e. $NIE = \beta^{(xs)} H_M H_Y^{-1} \beta^{(sy)} = \beta^{(xm)}\beta^{(my)}$ where $H_M H_Y^{-1}$ are derived from Proposition~\ref{prop:identifiability}). A detailed assessment of LaMe's capacity to retrieve causal natural effects is developed in the next subsection. 
We compare the models' estimates with the corresponding parameters used for data generation, $\beta^{(xm)}_{true}$ and $\beta^{(my)}_{true}$, using the mean absolute error across all coefficients as the evaluation metric. 


We present results under a challenging configuration with $I=100$, $J=100$, $L=1$, signal-to-noise ratios $SNR(M|S)=SNR(Y|X,S)=0.1$ and $SNR(S|X)=1$.
Data generated from the latent mediation system in Eq.~\eqref{eq:latent-mediation-equations} uses $K_{true}=10$ latent dimensions. 
For all latent models (OVP-LaMe aside), $K$ is set to $K_{true}$ when data are generated under a latent model. When data are generated under a non-latent model, we set $K=\frac{\min(J, I+L)}{2}$ to test latent models' ability to compress a full-rank covariance matrix between $X$ and $M,Y$ through a matrix of half this rank.
Regularized models (LASSO, ElasticNet, Pathway LASSO, PCA+LASSO) are not considered because the data in this experiment were generated without sparse coefficients.

Fig.~\ref{fig-XP2} reports the mean absolute error between the inferred ($\beta^{(xm)}, \beta^{(my)}$) and the true ($\beta^{(xm)}_{true}, \beta^{(my)}_{true}$), for varying numbers of observations $N$, and for data generated under latent and non-latent models. First, LaMe consistently outperforms all baselines in the latent case (Fig.~\ref{fig-XP2}-left). In particular, because $K_{true} \ll \min(J, I+L)$, LaMe reduces estimation error through its low-rank restriction where OLS cannot. Second, and as expected, OVP-LaMe performs similarly to OLS, confirming that the violation of Eq.~\eqref{eq:K-information-bottleneck} allows OVP-LaMe to converge to a solution equivalent to the OLS solution. In particular, OVP-LaMe is prone to overfitting, just as OLS, when data are generated under a latent model. In the non-latent case, LaMe outperforms latent baselines and performs nearly as well as the non-latent models, despite the constraint $K=\frac{\min(J, I+L)}{2}$. 
Additional results for $\{I=50, J=200$\}, as well as for varying signal-to-noise ratios are presented and discussed in \SM{\ref{sec-SM:Additional-synth-XP}}. 

\subsubsection{Causal effect recovery}
\begin{figure}
    \centering
    \includegraphics[width=\textwidth]{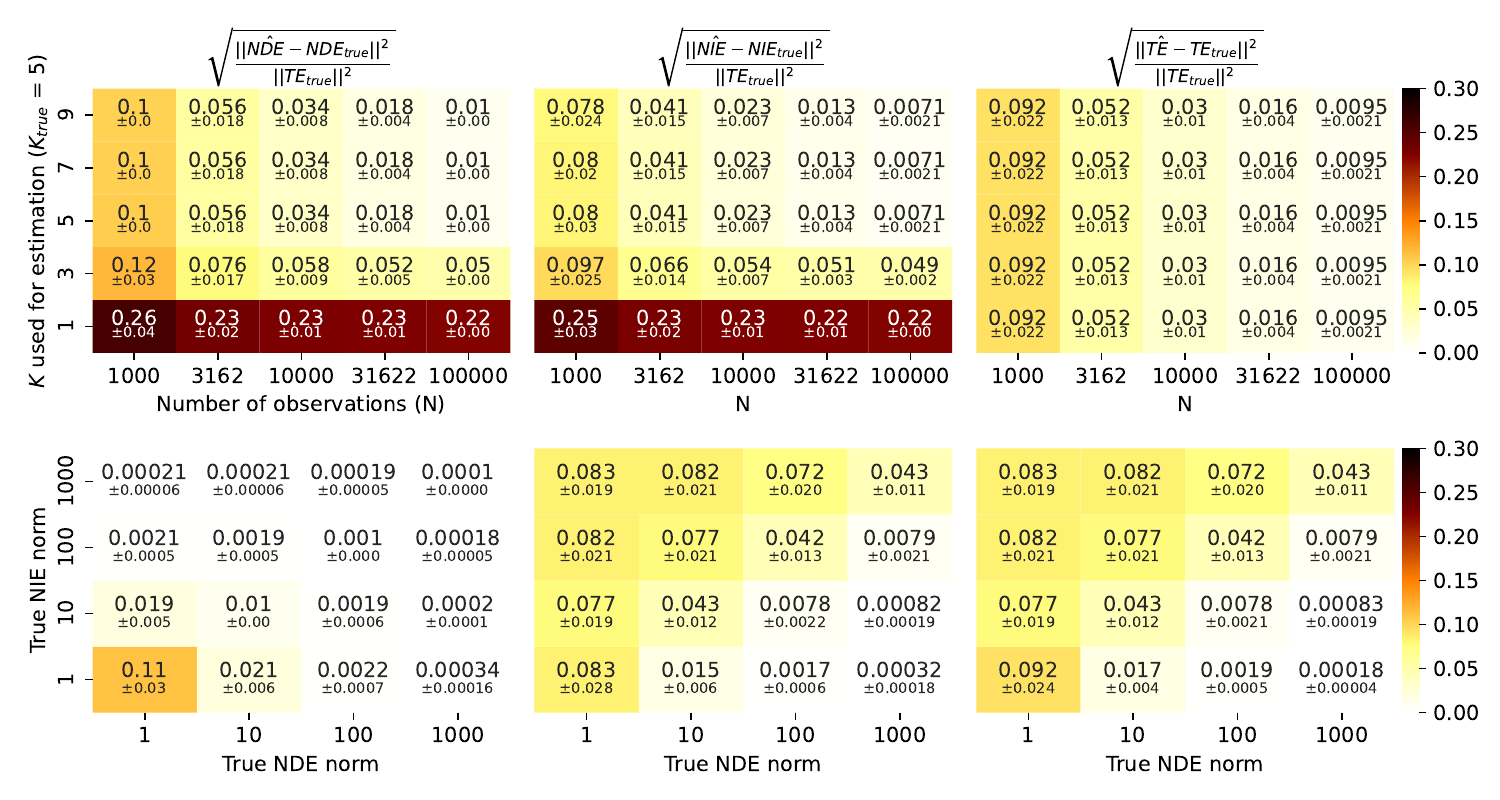}
    \caption{\textbf{LaMe accurately recovers the natural effects} --- Root mean squared error (RMSE) of estimated natural effect coefficients, normalized by the norm of the true total-effect coefficient, averaged over 100 runs, along with the associated standard deviation, for LaMe optimized with $N=1~000$ observations on data generated with $I=10$, $J=50$, $K=5$ and $L=1$, unless specified otherwise. \textbf{(Top)} Varying the number of observations and the number of latent variables used for estimation when the true number of latent variables used for data generation is fixed at $K_{true}=5$ | \textbf{(Bottom)} Varying the values of NIE and NDE coefficients norm used for data generation | Root mean squared error relative to the true total effect between the true and estimated \textbf{(Left)} natural direct effect, \textbf{(Middle)} natural indirect effect and \textbf{(Right)} total effect.
    Alt text: Heatmap representing the RMSE of LaMe with respect to the true NIE, NDE and TE path coefficients relative to the true TE. The RMSE of NIE, NDE and TE decreases with $N$, with $K-K_{true}$ (plateauing once $K \geq K_{true}$) and with $\frac{||NDE_{true}||^2}{||NIE_{true}||^2}$, with a relative RMSE remaining below 0.1 in most cases.
    }
    \label{fig-XP5-6}
\end{figure}
We recall from Eq.~\eqref{eq:NIE-NDE-expression} that $NIE=\beta^{(xs)}\beta^{(sy)}$, $NDE=\beta^{(xy)}$ and $TE=\beta^{(xs)}\beta^{(sy)}+\beta^{(xy)}$ \cite{Derkach2019HighDimMed,Albert2016LatentMediation}.
We observe that LaMe correctly retrieves the ground-truth natural effects coefficients overall in Fig.~\ref{fig-XP5-6}, when optimized with $N=1~000$ observations on data generated under the exact recoverability assumption (see Proposition~\ref{prop:identifiability}) with $I=10$, $J=50$, $K=5$ and $L=1$, unless specified otherwise. The normalization using the true total effect provides a shared comparison ground while avoiding artificial errors when either the NDE or NIE coefficient is small.
Fig.~\ref{fig-XP5-6}-top in particular shows that natural effects estimation error is stable when $K$ is larger than the true number of latent dimensions $K_{true}$ used for data generation. When $K<K_{true}$ however, the NDE and NIE estimates are not accurate, as the under-parametrized model must express effects through a smaller number of latent variables. Taken together, these findings hint at a possible model selection strategy based on parameter stability rather than information criteria. Estimation error also decreases over the sample sizes considered. Finally, note that the direct path ensures that TE is equal to the OLS regression coefficient $X$ and $Y$, namely $TE=(X^T X)^{-1} X^T Y$, explaining why it does not depend on $K$.
Fig.~\ref{fig-XP5-6}-bottom shows that LaMe estimates the NDE, NIE and TE coefficients more accurately when the norm of the NDE coefficient matrix exceeds that of the NIE coefficient matrix: in this case, most of the observed covariance between $X$ and $Y$ is expressed through the NDE, whose estimation does not require a fine approximation of the unobserved variable $S$. Conversely, this explains why the relative error on the NDE, NIE and TE coefficients increases when the norm of the NIE coefficient is larger than that of the NDE. Overall, LaMe accurately retrieves the true natural effects across a range of direct and indirect effect magnitudes, supporting the applicability of the LaMe model to various settings. 

\subsubsection{Retrieving the manifest variables involved in mediation}
\begin{figure}
    \centering
    \includegraphics[width=\linewidth]{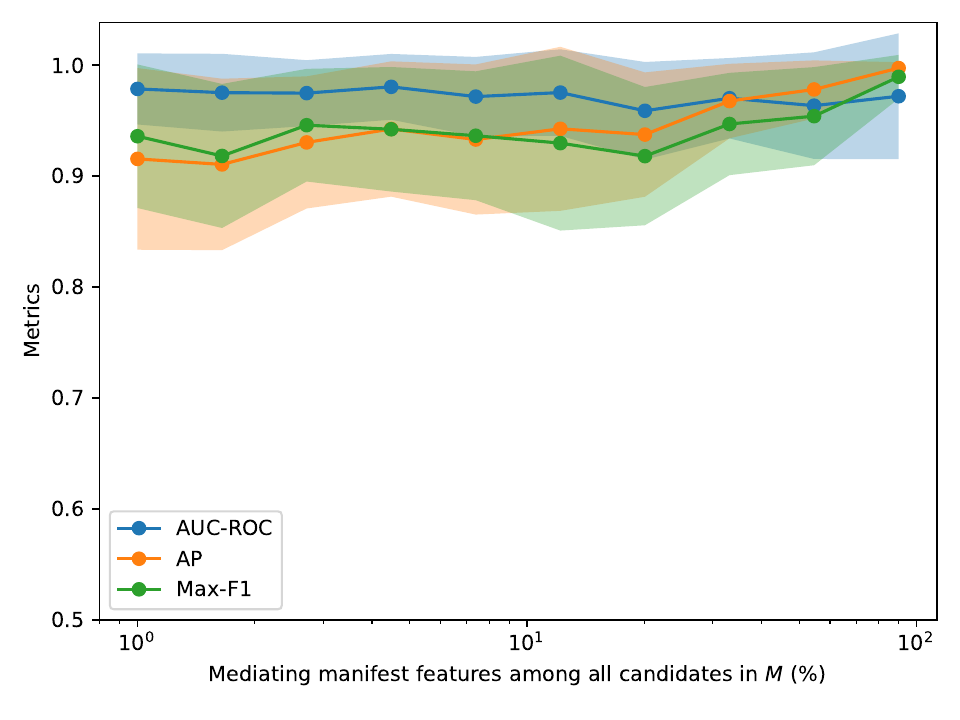}
    \caption{\textbf{LaMe accurately recovers the observed ``mediating'' set of manifest variables in $M$ among a large set of ``non-mediating'' candidates}  -- Data are generated from Eq.~\eqref{eq:latent-mediation-equations} with $I=10, J=10+\Delta J, K=2, L=1, N=5~000$ without confounding, where $\Delta J \in [1, 990]$ is the number of features in $M$ that do not contribute to mediation mechanisms (that are either not influenced by $X$, or not informative about $Y$ given $X$). The shaded area represents the standard deviation over 50 datasets generated under the same parameter setting. The AUC-ROC, AP and Max-F1 all range from 0 (worst) to 1 (best), and quantify how well the model recovers the true set of mediating manifest features among all possible candidates.}
    \label{fig:signature-recovery}
    Alt text: The three metrics remain close to 1 over the whole range of candidate mediators, with standard deviations around 0.05 across all experiments.
\end{figure}
\label{sec:mediator-selection}
We show here that LaMe efficiently identifies the entries of $M$ that are both influenced by $X$ through $S$ and informative about $Y$ through $S$ given $X$. These are not observed mediators, since there is no direct path between $M$ and $X$ or $Y$; $M$ is a descendant of the latent mediator rather than a causal intermediate on the indirect path. However, they are the closest analogue of observed mediating variables and are used as such in \cite{Derkach2019HighDimMed}; for simplicity, we refer to them as mediating manifest features.
Building on the definition proposed in \cite{Derkach2019HighDimMed}, we define a global mediating manifest feature $M_j$ as the $j^{th}$ entry of $M$ under the exact recoverability assumption such that both:
\begin{align*}
    \sum_i^I |C_{ij}^{XM}| :&= \sum_i^I |\sum_k^K \beta^{(xs)}_{ik} \beta^{(sm)}_{kj}| \neq 0   \qquad \text{and} \qquad \sum_l^L |C_{jl}^{MY}| := \sum_l^L |\sum_k^K \beta^{(ms)}_{jk} \beta^{(sy)}_{kl}| \neq 0
\end{align*}
A mediating manifest feature is therefore an entry of $M$, noted $M_j$, such that $C_j := \sum_i^I \sum_l^L |C_{ij}^{XM}| |C_{jl}^{MY}| \neq 0$.
This new definition is intended to generalize that proposed by \cite{Derkach2019HighDimMed}, in particular to make it invariant to linear transformations of the latent space and to be robust to cases where manifest variables are correlated (see \SM{\ref{sec-SM:issues-mediating-biomarkers}}). 

We generate data from Eq.~\eqref{eq:latent-mediation-equations} such that there are exactly $J=10$ observed mediating manifest variables $M_j$ with $C_j \neq 0$ and $\Delta J \in [1, 990]$ non-mediating manifest variables with $C_j = 0$. In practice, we generate full-rank parameters (which ensures that $C_j \neq 0$) for the system with $J$ mediating manifest variables, expand $\beta^{(sm)}$ and $\Sigma_{m|s}$ with $\Delta J$ columns (and rows for $\Sigma_{m|s}$) such that $C_j = 0$, and generate 5~000 observations according to the resulting set of parameters to train LaMe. We then evaluate whether the model accurately distinguishes the $\Delta J$ manifest variables with a true $C_j=0$ from the $J$ manifest variables with $C_j \neq 0$ using the area under the ROC curve (AUC-ROC), the Average Precision (AP) and the maximum F1 across classification thresholds (Max-F1). These metrics estimate how well the inferred $C_j$ (continuous value) distinguishes the nonzero entries in $C_j^{(true)}$ (binary value based on its null/non-null entries) without the need to specify a threshold between scores associated with nonzero and near-zero values of $C_j$. For AUC-ROC, AP and Max-F1, a value of 1 means the model achieves perfect separation between mediating and non-mediating variables in $M$.

The results presented in Fig.~\ref{fig:signature-recovery} show that LaMe consistently distinguishes the ten true mediating manifest features even in situations where these represent only $1\%$ of the total number of manifest variables.

\subsection{Real-world data}
\begin{table}
\caption{Out-of-sample and in-sample performance metrics of LaMe (version $X$ and $M$, with and without $\ell 1$ regularization, assuming full noise covariances) and all baseline models. The last lines report the average improvement of models' metrics over OLS across all datasets --hence the null line for OLS. NLPD is the Negative Log-Predictive Density, $R^2$ the fraction of explained variance and $AIC$ the Akaike Information Criterion. When applicable, the number of latent variables is reported as $K$. Errors correspond to the standard error across five folds. The colour scale indicates relative performance within each dataset and metric ; (orange=best). }
\label{table-res-RW-outer-red}
\centering
\noindent\makebox[\columnwidth]{\resizebox{\columnwidth}{!}{

\begin{tabular}{llSSSSS}

    \cline{1-6}

    & &\text{$NLPD(Y|X,M)$ $\downarrow$} & \text{$R^2(Y|X,M)$ $\uparrow$} & \text{AIC $\downarrow$} & \text{$K$ $\downarrow$}\\

    \cline{1-6}

    \multirow{10}{*}{\rotatebox[origin=c]{90}{\footnotesize \text{\textbf{NHANES}}}}
    & LaMe (vM) &  \cellcolor{orange!28} 1.356(20) &  \cellcolor{orange!14} 0.151(9) &  \cellcolor{orange!26} 97.404(1110) &  \cellcolor{orange!30} 1.000(1)\\
    & LaMe (vM) (reg) &  \cellcolor{orange!30} 1.355(20) &  \cellcolor{orange!30} 0.154(7) &  \cellcolor{orange!12} 98.270(1257) &  \cellcolor{orange!30} 1.000(1)\\
    & LaMe (vX) &  \cellcolor{orange!28} 1.356(20) &  \cellcolor{orange!14} 0.151(9) &  \cellcolor{orange!26} 97.404(1110) &  \cellcolor{orange!30} 1.000(1)\\
    & LaMe (vX) (reg) &  \cellcolor{orange!30} 1.355(20) &  \cellcolor{orange!29} 0.154(7) &  \cellcolor{orange!9} 98.449(1325) &  \cellcolor{orange!30} 1.000(1)\\
    & PCA+OLS &  \cellcolor{orange!0} 1.386(22) &  \cellcolor{orange!0} 0.114(23) &  \cellcolor{orange!0} 99.382(1086) &  \cellcolor{orange!0} 24.000(1)\\
    & PCA+LASSO &  \cellcolor{orange!27} 1.357(20) &  \cellcolor{orange!5} 0.150(12) &  \cellcolor{orange!0} 99.872(1077) &  \cellcolor{orange!0} 24.000(1)\\
    & PLS(X,M,Y)+OLS &  \cellcolor{orange!0} 1.502(35) &  \cellcolor{orange!0} -0.148(104) &  \cellcolor{orange!26} 97.406(1107) &  \cellcolor{orange!0} 3.600(456)\\
    & PLS(X,M)+OLS &  \cellcolor{orange!25} 1.358(20) &  \cellcolor{orange!8} 0.150(8) &  \cellcolor{orange!30} 97.147(1058) &  \cellcolor{orange!0} 3.400(358)\\
    & OVP-LaMe &  \cellcolor{orange!0} 1.389(23) &  \cellcolor{orange!0} 0.110(24) &  \cellcolor{orange!6} 98.597(1087) &  \cellcolor{orange!0} 19.000(1)\\
    & JAOG &  \cellcolor{orange!8} 1.367(19) &  \cellcolor{orange!0} 0.136(11) &  \cellcolor{orange!20} 97.761(1120) &  \cellcolor{orange!16} 1.800(179)\\

    \cline{1-6}

    \multirow{10}{*}{\rotatebox[origin=c]{90}{\footnotesize \text{\textbf{EPIC CAN}}}}
    & LaMe (vM) &  \cellcolor{orange!21} 0.724(3) &  \cellcolor{orange!22} 0.752(1) &  \cellcolor{orange!26} 899.770(1959) &  \cellcolor{orange!30} 1.000(1)\\
    & LaMe (vM) (reg) &  \cellcolor{orange!25} 0.719(3) &  \cellcolor{orange!26} 0.754(1) &  \cellcolor{orange!25} 900.111(2035) &  \cellcolor{orange!30} 1.000(1)\\
    & LaMe (vX) &  \cellcolor{orange!25} 0.720(2) &  \cellcolor{orange!25} 0.754(1) &  \cellcolor{orange!26} 899.784(1962) &  \cellcolor{orange!30} 1.000(1)\\
    & LaMe (vX) (reg) &  \cellcolor{orange!26} 0.719(3) &  \cellcolor{orange!26} 0.755(1) &  \cellcolor{orange!18} 902.733(1817) &  \cellcolor{orange!30} 1.000(1)\\
    & PCA+OLS &  \cellcolor{orange!0} 0.779(9) &  \cellcolor{orange!0} 0.738(1) &  \cellcolor{orange!0} 993.797(1271) &  \cellcolor{orange!0} 130.600(1972)\\
    & PCA+LASSO &  \cellcolor{orange!30} 0.714(3) &  \cellcolor{orange!30} 0.757(1) &  \cellcolor{orange!0} 998.631(1333) &  \cellcolor{orange!0} 130.600(1972)\\
    & PLS(X,M,Y)+OLS &  \cellcolor{orange!0} 0.791(17) &  \cellcolor{orange!0} 0.720(9) &  \cellcolor{orange!30} 898.288(1961) &  \cellcolor{orange!0} 11.600(219)\\
    & PLS(X,M)+OLS &  \cellcolor{orange!0} 0.753(8) &  \cellcolor{orange!0} 0.738(5) &  \cellcolor{orange!30} 898.268(1964) &  \cellcolor{orange!0} 11.600(219)\\
    & OVP-LaMe &  \cellcolor{orange!0} 1.392(36) &  \cellcolor{orange!0} 0.609(4) &  \cellcolor{orange!0} 1046.354(1913) &  \cellcolor{orange!0} 206.000(1)\\
    & JAOG &  \cellcolor{orange!0} 1.173(86) &  \cellcolor{orange!0} 0.622(18) &  \cellcolor{orange!27} 899.466(1745) &  \cellcolor{orange!16} 4.400(1486)\\

    \cline{1-6}

    \multirow{10}{*}{\rotatebox[origin=c]{90}{\footnotesize \text{\textbf{PISA - I}}}}
    & LaMe (vM) &  \cellcolor{orange!29} 1.068(1) &  \cellcolor{orange!27} 0.504(1) &  \cellcolor{orange!30} 102.335(98) &  \cellcolor{orange!9} 7.000(1)\\
    & LaMe (vM) (reg) &  \cellcolor{orange!29} 1.069(1) &  \cellcolor{orange!27} 0.504(1) &  \cellcolor{orange!30} 102.337(106) &  \cellcolor{orange!9} 7.000(1)\\
    & LaMe (vX) &  \cellcolor{orange!29} 1.068(1) &  \cellcolor{orange!27} 0.504(1) &  \cellcolor{orange!30} 102.336(98) &  \cellcolor{orange!9} 7.000(1)\\
    & LaMe (vX) (reg) &  \cellcolor{orange!29} 1.068(1) &  \cellcolor{orange!28} 0.504(1) &  \cellcolor{orange!30} 102.328(112) &  \cellcolor{orange!9} 7.000(1)\\
    & PCA+OLS &  \cellcolor{orange!13} 1.077(1) &  \cellcolor{orange!0} 0.496(1) &  \cellcolor{orange!29} 102.353(99) &  \cellcolor{orange!0} 19.000(1)\\
    & PCA+LASSO &  \cellcolor{orange!13} 1.077(1) &  \cellcolor{orange!0} 0.496(1) &  \cellcolor{orange!29} 102.367(99) &  \cellcolor{orange!0} 19.000(1)\\
    & PLS(X,M,Y)+OLS &  \cellcolor{orange!0} 1.438(108) &  \cellcolor{orange!0} -0.181(298) &  \cellcolor{orange!9} 103.065(202) &  \cellcolor{orange!0} 12.000(1)\\
    & PLS(X,M)+OLS &  \cellcolor{orange!0} 1.419(109) &  \cellcolor{orange!0} -0.149(308) &  \cellcolor{orange!10} 103.023(204) &  \cellcolor{orange!0} 12.000(1)\\
    & OVP-LaMe &  \cellcolor{orange!30} 1.068(1) &  \cellcolor{orange!30} 0.505(1) &  \cellcolor{orange!29} 102.365(99) &  \cellcolor{orange!0} 23.000(1)\\
    & JAOG &  \cellcolor{orange!0} 1.162(1) &  \cellcolor{orange!0} 0.402(1) &  \cellcolor{orange!0} 103.858(108) &  \cellcolor{orange!30} 1.000(1)\\

    \cline{1-6}

    \multirow{10}{*}{\rotatebox[origin=c]{90}{\footnotesize \text{\textbf{CRIME}}}}
    & LaMe (vM) &  \cellcolor{orange!6} 1.055(11) &  \cellcolor{orange!18} 0.575(4) &  \cellcolor{orange!30} 79.560(1031) &  \cellcolor{orange!18} 7.000(1)\\
    & LaMe (vM) (reg) &  \cellcolor{orange!27} 1.037(8) &  \cellcolor{orange!22} 0.578(3) &  \cellcolor{orange!0} 92.369(1104) &  \cellcolor{orange!30} 1.000(1)\\
    & LaMe (vX) &  \cellcolor{orange!6} 1.055(11) &  \cellcolor{orange!18} 0.575(4) &  \cellcolor{orange!30} 79.560(1031) &  \cellcolor{orange!18} 7.000(1)\\
    & LaMe (vX) (reg) &  \cellcolor{orange!28} 1.037(8) &  \cellcolor{orange!22} 0.578(3) &  \cellcolor{orange!0} 93.572(1722) &  \cellcolor{orange!30} 1.000(1)\\
    & PCA+OLS &  \cellcolor{orange!0} 1.135(21) &  \cellcolor{orange!0} 0.556(8) &  \cellcolor{orange!23} 81.309(1005) &  \cellcolor{orange!0} 31.600(219)\\
    & PCA+LASSO &  \cellcolor{orange!20} 1.044(14) &  \cellcolor{orange!30} 0.582(4) &  \cellcolor{orange!3} 86.625(927) &  \cellcolor{orange!0} 31.600(219)\\
    & PLS(X,M,Y)+OLS &  \cellcolor{orange!30} 1.035(7) &  \cellcolor{orange!0} 0.564(6) &  \cellcolor{orange!3} 86.660(1065) &  \cellcolor{orange!0} 24.800(2704)\\
    & PLS(X,M)+OLS &  \cellcolor{orange!28} 1.037(7) &  \cellcolor{orange!2} 0.566(5) &  \cellcolor{orange!3} 86.642(1070) &  \cellcolor{orange!0} 25.400(2586)\\
    & OVP-LaMe &  \cellcolor{orange!0} 1.149(23) &  \cellcolor{orange!0} 0.550(10) &  \cellcolor{orange!0} 90.303(991) &  \cellcolor{orange!0} 50.000(1)\\
    & JAOG &  \cellcolor{orange!0} 1.184(36) &  \cellcolor{orange!0} 0.516(22) &  \cellcolor{orange!16} 83.379(901) &  \cellcolor{orange!22} 5.000(1)\\

    \cline{1-6}

\end{tabular}

}}
\end{table}


\subsubsection{Data}
The \textbf{NHANES} dataset \cite{NHANES2021Dataset} examines whether participants' ($N=2,710$) glycohemoglobin levels (Y) are a direct consequence of lifestyle habits ($I=14$; hours of sleep, alcohol intake, sunscreen use, etc.) or are mediated by biomarkers ($J=25$; blood levels of lead, insulin, cholesterol, etc.).

The \textbf{EPIC-CAN} dataset \cite{EPIC2024Dataset} examines whether the age of a cancer event ($N=4,993$) is directly influenced by socio-demographic and lifestyle features ($I=33$; weight, height, gender, age, smoking habits, physical activity index, diet, etc.) or whether this influence is mediated by blood levels of a high-dimensional collection of candidate proteomic features measured at inclusion in the study ($J=379$). 

The \textbf{CRIME} dataset \cite{CrimeDataset} examines whether cities' ($N=1,993$) crime rates (output $Y$) are a direct consequence of population characteristics ($I=67$ representing the number of inhabitants, unemployment rate, percentage of divorce, ethnicity, etc.) or are mediated by city-level contextual factors ($J=33$ representing median rent, public transportation, number of vacant houses, etc.).

The \textbf{PISA-I} dataset comes from the PISA corpus \cite{PISA2022Dataset}. It examines whether students' ($N=52,028$) mathematics proficiency score is a direct consequence of their home environment ($I=27$; indexes quantifying home possessions, availability of electronic devices, mental health, body image, etc.) or is mediated by the school environment ($J=20$; indexes quantifying the sense of belonging, safety perception, school quality, etc.).

The complete list of variables is provided in \SM{\ref{sec-SM:RW-labels}}.
The applications evaluate whether LaMe can summarize exposure–biomarker–outcome associations using compact latent representations while retaining predictive performance; these applications are not intended to provide substantive causal mediation analyses. In particular, for EPIC-CAN, our analysis focused on individuals who developed cancer during follow-up, precluding a straightforward causal interpretation of the fitted relationships. We therefore do not report natural direct and indirect effects for these applications, and instead assess predictive performance, goodness of fit, and the ability to summarize observed relationships through parsimonious latent representations.

\subsubsection{Method}
We empirically observe that most models reach a plateau in predictive performance (the addition of new data no longer improves predictive accuracy) when trained on 80\% of the data, and converge to similar fitted solutions. We instead report the results for a small-training-sample stress test by using an outer 5-fold cross validation procedure: we train the models on 20\% of the available data, and evaluate them on the 80\% remaining observations; this is iterated five times with different training data. The results from the usual 5-fold cross validation are provided in \SM{\ref{sec-SM:RW-XP-5xCV}}. Note that this procedure serves for unbiased evaluation only, and is not involved in any hyper-parameter tuning: test data are systematically isolated from the training procedure.

We optimize LaMe for $K \in [1, 2, 3, 4, 5, 7, 10, 15, 20, 25]$ and select the version with the lowest Bayesian Information Criterion within each fold; for PCA, $K$ has been chosen using scikit-learn's maximum likelihood estimator for the number of components; for PLS and JAOG, $K$ has been chosen to maximize their respective scoring function on a validation set (10\% randomly sampled from the training set). 

\subsubsection{Results}
We report our results for LaMe and latent baselines in Table~\ref{table-res-RW-outer-red}. LaMe achieves predictive performance comparable to that of the strongest latent baselines, sometimes improving on them, while using parsimonious representations. 
Visualizations of the inferred systems are provided in \SM{\ref{sec-SM:full-mediation-graphs}}.
In particular, these experiments show that LaMe achieves competitive predictive performance with compact latent representations. These results support a favorable trade-off between predictive performance and latent dimensionality under the stated model-selection procedures. 
On the other hand, over-parametrized models such as OVP-LaMe converge to solutions that are equivalent to the OLS solution. A similar case arises for PCA-based methods here, for which the maximum likelihood estimator selected models with large $K$ --thus forfeiting any low-dimensional covariance structure. 
For regularized LaMe on CRIME, and for NHANES and EPIC-CAN, LaMe uses a single latent dimension where other latent approaches used many more to achieve comparable performance in terms of prediction and goodness-of-fit.


\section{Conclusion}
We have introduced LaMe as a general framework for joint estimation of multivariate latent mediation systems that makes the conditions for dimension reduction and identification of the effect decomposition explicit. 
The simulations demonstrate gains from low-rank estimation when the data contain shared latent structure and recovery of natural effects under the stipulated measurement model. They also show that accurate estimation of the total effect can coexist with persistent errors in its direct and indirect components when too few latent dimensions are fitted. The real-world applications complement these findings by demonstrating competitive prediction with compact latent representations. The present evidence supports LaMe as a scalable, parsimonious and accurate option for high-dimensional mediation analysis across a wide variety of situations.


Possible directions for future work include 
incorporating mixed data types such as categorical exposures, 
jointly inferring the number of latent components $K$ (e.g. through point-process priors), 
relaxing linearity assumptions and
extending the framework to survival models with censored data. 

\paragraph{Acknowledgments}
This work was supported by a grant from the French National Research Agency as part of the ``Investissements d’Avenir ExcellencES'' program from France 2030 (SHAPE-Med@Lyon; [ANR-22-EXES-0012]). 
Funding for grant [IIG\_FULL\_2022\_013] was obtained from Wereld Kanker Onderzoek Fonds (WKOF) as part of the World Cancer Research Fund International grant programme.
Where authors are identified as personnel of the International Agency for Research on Cancer/World Health Organization, the authors alone are responsible for the views expressed in this article and they do not necessarily represent the decisions, policy, or views of the International Agency for Research on Cancer/World Health Organization.

\paragraph{Conflicts of interest} The authors declare no conflict of interest.


\paragraph{Data availability statement}
The NHANES August 2021–August 2023 data are publicly available 
at \url{https://wwwn.cdc.gov/nchs/nhanes/continuousnhanes/default.aspx?Cycle=2021-2023}.
The CRIME dataset (DOI: 10.24432/C53W3X) is publicly available 
at \url{https://doi.org/10.24432/C53W3X}.
The PISA-I data were derived from the PISA 2022 database, publicly available 
at \url{https://www.oecd.org/en/data/datasets/pisa-2022-database.html}.
The data underlying EPIC-CAN is based on data from the European Prospective Investigation into Cancer and Nutrition (EPIC) and cannot be shared publicly due to privacy considerations. EPIC data are available by application to the EPIC Steering Committee (\url{https://epic.iarc.fr/access/}).


\bibliographystyle{biom} 
\bibliography{Bibliography}

@article{Rijnhart2021MediationInResearch,
    author = {Judith J. M. Rijnhart and Sophia J. Lamp and Matthew J. Valente and David P. MacKinnon and Jos W. R. Twisk and Martijn W. Heymans},
    title = {Mediation analysis methods used in observational research: a scoping review and recommendations},
    journal = {BMC Med Res Methodol},
    volume = {21},
    year = {2021},
    doi={https://doi.org/10.1186/s12874-021-01426-3}
}

@article{Lindquist2023MLHighDimMediation,
    title = {A machine learning based approach towards high-dimensional mediation analysis},
    author = {Tanmay Nath and Brian Caffo and Tor Wager and Martin A. Lindquist},
    journal = {NeuroImage},
    volume = {268},
    pages = {119843},
    year = {2023},
    issn = {1053-8119},
    doi = {https://doi.org/10.1016/j.neuroimage.2022.119843},
    url = {https://www.sciencedirect.com/science/article/pii/S1053811922009648},
}

@article{Albert2016LatentMediation,
    author = {Albert, Jeffrey M. and Geng, Cuiyu and Nelson, Suchitra},
    title = {Causal mediation analysis with a latent mediator},
    journal = {Biometrical Journal},
    volume = {58},
    number = {3},
    pages = {535-548},
    doi = {https://doi.org/10.1002/bimj.201400124},
    url = {https://onlinelibrary.wiley.com/doi/abs/10.1002/bimj.201400124},
    eprint = {https://onlinelibrary.wiley.com/doi/pdf/10.1002/bimj.201400124},
    year = {2016}
}

@article{Derkach2019HighDimMed,
    author = {Derkach, Andriy and Pfeiffer, Ruth M. and Chen, Ting-Huei and Sampson, Joshua N.},
    title = {High Dimensional Mediation Analysis With Latent Variables},
    journal = {Biometrics},
    volume = {75},
    number = {3},
    pages = {745-756},
    year = {2019},
    month = {03},
    issn = {0006-341X},
    doi = {10.1111/biom.13053},
    url = {https://doi.org/10.1111/biom.13053},
    eprint = {https://academic.oup.com/biometrics/article-pdf/75/3/745/56121846/biometrics\_75\_3\_745.pdf},
}

@inproceedings{Zhu2024SurveyMediationModels,
    author = {Zhu, Yaochen and He, Yinhan and Ma, Jing and Hu, Mengxuan and Li, Sheng and Li, Jundong},
    title = {Causal Inference with Latent Variables: Recent Advances and Future Prospectives},
    year = {2024},
    isbn = {9798400704901},
    publisher = {Association for Computing Machinery},
    address = {New York, NY, USA},
    url = {https://doi.org/10.1145/3637528.3671450},
    doi = {10.1145/3637528.3671450},
    booktitle = {Proceedings of the 30th ACM SIGKDD},
    pages = {6677–6687},
    numpages = {11},
    location = {Barcelona, Spain},
    series = {KDD '24}
}

@Inbook{Neal1998ConvergenceEM,
    author={Neal, Radford M. and Hinton, Geoffrey E.},
    title={A View of the Em Algorithm that Justifies Incremental, Sparse, and other Variants},
    bookTitle={Learning in Graphical Models},
    year={1998},
    publisher={Springer Netherlands},
    pages={355--368},
    isbn={978-94-011-5014-9},
    doi={10.1007/978-94-011-5014-9_12},
    url={https://doi.org/10.1007/978-94-011-5014-9_12}
}

@book{Bishop2006PatternReco,
    author ={Bishop, Christopher M.},
    title={Pattern Recognition and Machine Learning},
    publisher={New York: Springer},
    url={https://tinyurl.com/43ehs9x9},
    year={2006}
}

@article{Zhao2021PathwayLASSO,
    author = {Zhao, Yi and Luo, Xi},
    journal = {Statistics and its interface},
    number = {1},
    pages = {39–50},
    title = {Pathway Lasso: Pathway Estimation and Selection with High-Dimensional Mediators},
    volume = {15},
    year = {2022},
    doi={10.4310/21-sii673},
}

@article{Wang2018PrecisionLASSO,
    author = {Wang, Haohan and Lengerich, Benjamin J and Aragam, Bryon and Xing, Eric P},
    title = {Precision Lasso: accounting for correlations and linear dependencies in high-dimensional genomic data},
    journal = {Bioinformatics},
    volume = {35},
    number = {7},
    pages = {1181-1187},
    year = {2018},
    month = {09},
    issn = {1367-4803},
    doi = {10.1093/bioinformatics/bty750},
}

@article{Ye2021LASSOVarSelectionHighDim,
    author = {Ye, Z. and Zhu, Y. and Coffman, D. L.},
    title = {Variable selection for causal mediation analysis using LASSO-based methods},
    year = {2021},
    journal = {Statistical methods in medical research},
    volume = {30},
    number = {6},
    pages={1413–1427},
    doi={10.1177/0962280221997505},
}

@article{Zhao2020SparsePCA,
    title = {Sparse principal component based high-dimensional mediation analysis},
    journal = {Computational Statistics and Data Analysis},
    volume = {142},
    pages = {106835},
    year = {2020},
    issn = {0167-9473},
    doi = {https://doi.org/10.1016/j.csda.2019.106835},
    url = {https://www.sciencedirect.com/science/article/pii/S0167947319301902},
    author = {Yi Zhao and Martin A. Lindquist and Brian S. Caffo},
}

@book{Hayashi2000OLSisMLE,
    author    = {Hayashi, Fumio},
    title     = {Econometrics},
    publisher = {Princeton University Press},
    year      = {2000},
    ISBN = {9780691010182},
    pages={15},
}

@misc{PISA2022Dataset,
  author       = {{Organisation for Economic Co-operation and Development}},
  title        = {{PISA 2022 Database}},
  year         = {2023},
  howpublished = {OECD},
  url          = {https://www.oecd.org/en/data/datasets/pisa-2022-database.html},
  urldate      = {2025-06-26},
  note         = {Available at \url{https://www.oecd.org/en/data/datasets/pisa-2022-database.html}
                  (accessed June 26, 2025)}
}

@misc{NHANES2021Dataset,
  author       = {{National Center for Health Statistics}},
  title        = {{National Health and Nutrition Examination Survey,
                   August 2021--August 2023}},
  year         = {2024},
  howpublished = {U.S. Department of Health and Human Services,
                  Centers for Disease Control and Prevention},
  url          = {https://wwwn.cdc.gov/nchs/nhanes/continuousnhanes/default.aspx?Cycle=2021-2023},
  urldate      = {2025-06-26},
  note         = {Available at \url{https://wwwn.cdc.gov/nchs/nhanes/continuousnhanes/default.aspx?Cycle=2021-2023}
                  (accessed June 26, 2025)}
}

@misc{CrimeDataset,
  author       = {Redmond, Michael},
  title        = {{Communities and Crime}},
  year         = {2002},
  howpublished = {UCI Machine Learning Repository},
  doi          = {10.24432/C53W3X},
  url          = {https://doi.org/10.24432/C53W3X},
  note         = {Available at \url{https://doi.org/10.24432/C53W3X}
                  (accessed June 26, 2025)}
}

@misc{EPIC2024Dataset,
  author       = {{International Agency for Research on Cancer}},
  title        = {{European Prospective Investigation into Cancer and Nutrition (EPIC) cohort}},
  year         = {2024},
  publisher    = {International Agency for Research on Cancer, World Health Organization},
  url          = {https://epic.iarc.fr/},
  urldate      = {2025-06-26},
  note         = {(accessed June 26, 2025)}
}

@article{Chechik2005InformationBottleneck,
    author  = {Gal Chechik and Amir Globerson and Naftali Tishby and Yair Weiss},
    title   = {Information Bottleneck for Gaussian Variables},
    journal = {Journal of Machine Learning Research},
    year    = {2005},
    volume  = {6},
    number  = {6},
    pages   = {165--188},
    url     = {http://jmlr.org/papers/v6/chechik05a.html}
}

@article{Zeng2021SurveyMediationTechniques,
    title = {Statistical methods for mediation analysis in the era of high-throughput genomics: Current successes and future challenges},
    journal = {Computational and Structural Biotechnology Journal},
    volume = {19},
    pages = {3209-3224},
    year = {2021},
    issn = {2001-0370},
    doi = {https://doi.org/10.1016/j.csbj.2021.05.042},
    url = {https://www.sciencedirect.com/science/article/pii/S2001037021002233},
    author = {Ping Zeng and Zhonghe Shao and Xiang Zhou},
}

@article{Kawano2018TwoStepsPCAPCRMediation,
    title = {Sparse principal component regression for generalized linear models},
    journal = {Computational Statistics \& Data Analysis},
    volume = {124},
    pages = {180-196},
    year = {2018},
    issn = {0167-9473},
    doi = {https://doi.org/10.1016/j.csda.2018.03.008},
    url = {https://www.sciencedirect.com/science/article/pii/S0167947318300562},
    author = {Shuichi Kawano and Hironori Fujisawa and Toyoyuki Takada and Toshihiko Shiroishi},
}

@inproceedings{Anderson1956StatisticalIdentifiabilityFA,
    title={Statistical inference in factor analysis},
    author={Anderson, Theodore W and Rubin, Herman and others},
    booktitle={Proceedings of the third Berkeley symposium on mathematical statistics and probability},
    volume={5},
    pages={111--150},
    year={1956}
}

@article{Cai2023TwoStepsLatentMediation,
    author = {Cai, Xizhen and Wang, Qing and Zhu, Yeying},
    title = {Mediation analysis with latent factors using simultaneous group-wise and parameter-wise penalization},
    journal = {Stat},
    volume = {12},
    number = {1},
    pages = {630},
    doi = {https://doi.org/10.1002/sta4.630},
    url = {https://onlinelibrary.wiley.com/doi/abs/10.1002/sta4.630},
    eprint = {https://onlinelibrary.wiley.com/doi/pdf/10.1002/sta4.630},
    year = {2023}
}

@article{Hu2024TwoStepsLatentMediation,
    author = {Hu, Jia Yuan and DeSimone, Marley and Wang, Qing},
    title = {Utilizing latent connectivity among mediators in high-dimensional mediation analysis},
    journal = {Stat},
    volume = {13},
    number = {2},
    pages = {675},
    doi = {https://doi.org/10.1002/sta4.675},
    url = {https://onlinelibrary.wiley.com/doi/abs/10.1002/sta4.675},
    eprint = {https://onlinelibrary.wiley.com/doi/pdf/10.1002/sta4.675},
    year = {2024}
}

@article{Chen2017DirectionsofMediationLatentMed,
    author = {Chén, Oliver Y and Crainiceanu, Ciprian and Ogburn, Elizabeth L and Caffo, Brian S and Wager, Tor D and Lindquist, Martin A},
    title = {High-dimensional multivariate mediation with application to neuroimaging data},
    journal = {Biostatistics},
    volume = {19},
    number = {2},
    pages = {121-136},
    year = {2017},
    month = {06},
    issn = {1465-4644},
    doi = {10.1093/biostatistics/kxx027},
    url = {https://doi.org/10.1093/biostatistics/kxx027},
    eprint = {https://academic.oup.com/biostatistics/article-pdf/19/2/121/24268714/kxx027.pdf},
}

\paragraph{Supplementary Materials}
Web Appendices are available with this paper at the Biometrics website on Oxford Academic.
The Python code for performing the simulations and data analysis in this paper is available at 
\url{https://github.com/GaelPouxMedard/LaMe-repr/releases/tag/v1.1.0-paper}. Code is permanently archived in Zenodo at \url{https://doi.org/10.5281/zenodo.23264883}.

\section{Observed confounding between exposures, mediators and outcomes}
\label{sec-SM:observed-con}

Based on \cite{Derkach2019HighDimMed}, observed confounding $C$ enters latent mediation systems as follows:
\begin{align}
    \label{eq:latent-mediation-equations-con}
    \left\{
    \begin{aligned}
        C &= \varepsilon_c  \\
        X &= C \beta^{(cx)} + \varepsilon_x  \\
        S &= X \beta^{(xs)} + C \beta^{(cs)} + \varepsilon_s  \\
        M &= S \beta^{(sm)} + \varepsilon_m  \\
        Y &= X \beta^{(xy)} + S \beta^{(sy)} + C \beta^{(cy)} + \varepsilon_y
    \end{aligned} \right.
\end{align}

Observed confounding $C$ enters the non-latent mediation system as follows:
\begin{align}
    \label{eq:mediation-equations-con}
    \left\{
    \begin{aligned}
        C &= \varepsilon_c  \\
        X &= C \beta^{(cx)} + \varepsilon_x  \\
        M &= X \beta^{(xm)} + C \beta^{(cm)} + \varepsilon_m  \\
        Y &= X \beta^{(xy)} + M \beta^{(my)} + C \beta^{(cy)} + \varepsilon_y
    \end{aligned} \right.
\end{align}

Here, we demonstrate that observed exposures $X$ and observed confounders $C$ can be jointly represented as a single composite observed variable $\xi := [X,C]$ from a likelihood perspective. As such, the entire technical demonstration in the main text can be conducted by considering $\xi=[X]$ without loss of generality. Modelling confounding simply requires replacing exposures by their augmented counterpart $\xi$ of dimension $I+\dim(C)$, and performing the separation step below after parameters estimation.

For latent systems (Eq.~\eqref{eq:latent-mediation-equations-con}), we start from any joint distribution $p(X,S,M,Y,C)$ along with the conditional independence relationships corresponding to confounded LaMe (see Eq.~\eqref{eq:latent-mediation-equations-con}), namely $M \perp (X,C) \mid S$ and $Y \perp M \mid (X,C,S)$. Developing $p(X,S,M,Y,C)$ using the chain rule, we show that $X$ and $C$ can be treated jointly as a single observed variable from a statistical standpoint:
\begin{align}
    \label{eq:factorisation-observed-confounders}
    &p(X,S,M,Y,C) \\ \notag
    = &p(X,C)p(S|X,C)p(M|S)p(Y|X,C,S)  \\ \notag
    := &p(\xi)p(S|\xi)p(M|S)p(Y|\xi,S),
\end{align}
where we defined $\xi :=[X,C]$ as a composite observed variable. 
The last factorization is identical in form to that of unconfounded LaMe-vX in the main text, with $\xi$ replacing $X$; maximum likelihood estimates for all these parameters are therefore obtained from the same inference procedure, treating $\xi$ as an augmented observed exposure variable of dimension $I+\dim(C)$.

For non-latent systems (Eq.~\eqref{eq:mediation-equations-con}), the same demonstration immediately follows from the chain rule noting that $p(\xi,M,Y)= p(\xi)p(M|\xi)p(Y|\xi,M)$.

The path coefficients from $X$ and $C$ to the mediator and the outcome are then obtained by appropriate slicing of the coefficients associated with $\xi$ e.g. in the latent case $\beta^{(\xi s)}=[\beta^{(xs)}, \beta^{(cs)}]^T$ and $\beta^{(\xi y)}=[\beta^{(xy)}, \beta^{(cy)}]^T$. In both latent and non-latent cases, the path coefficient from $C$ to $X$ is obtained from the off-diagonal blocks in the inferred covariance of $\xi$, namely $\Sigma_{\xi}$. After appropriate slicing of $\Sigma_{\xi}$ in terms of exposures $X$ and confounders $C$:
\begin{align}
    \label{eq:offdiag-to-direct-path}
    \Sigma_{\xi} =
    \begin{pmatrix}
        \Sigma_{x} & \Sigma_{x c}\\
        \Sigma_{c x} & \Sigma_{c}
    \end{pmatrix},
\end{align}
then the path coefficient from $C$ to $X$ is $\beta^{(c x)} = \Sigma_{c}^{-1} \Sigma_{c x}$.

\section{Exact recoverability error}
\label{sec-SM:exact-recov-limit}
We recall that:
\begin{align*}
    \beta^{(ms)} &= \underbrace{((\beta^{(sm)})^T\Sigma_{s}\beta^{(sm)} + \Sigma_{m|s})^{-1}}_{\Sigma_m^{-1}} (\beta^{(sm)})^T \underbrace{(\Sigma_{s|x} + (\beta^{(xs)})^T \Sigma_{x} \beta^{(xs)})}_{\Sigma_s}
\end{align*}
$\beta^{(ms)} = \Sigma_m^{-1} (\beta^{(sm)})^T \Sigma_s$.

The expected difference between $S$ and $S^*$ is zero following the tower property of nested expectations:
\begin{align*}
    \mathbb{E}[S^*-S] &= \mathbb{E}[\mathbb{E}[S^*-S|M]] \\
    &= 0
\end{align*}

The variance of the difference is:
\begin{align*}
    Var(S^*-S) &= Var(S) + (1-2) Var(\mathbb{E}[S|M]) \\
    &= \Sigma_s - (\beta^{(ms)})^T\Sigma_{m}\beta^{(ms)} \\
    &= \Sigma_s (I - \beta^{(sm)}\beta^{(ms)})
\end{align*}
From the definition of $\beta^{(ms)}$:
\begin{align*}
    \beta^{(ms)} &= \underbrace{((\beta^{(sm)})^T\Sigma_{s}\beta^{(sm)} + \Sigma_{m|s})^{-1}}_{\Sigma_m^{-1}} (\beta^{(sm)})^T \Sigma_s  \\
    &=  \Sigma_{m|s}^{-1} ((\beta^{(sm)})^T \Sigma_s - (\beta^{(sm)})^T\Sigma_{s}\beta^{(sm)}\beta^{(ms)})  \\
    &= \Sigma_{m|s}^{-1} (\beta^{(sm)})^T \Sigma_s(I - \beta^{(sm)}\beta^{(ms)})  \\
    &= \Sigma_{m|s}^{-1} (\beta^{(sm)})^T Var(S^*-S)  \\
\end{align*}
Therefore:
\begin{align*}
    Var(S^*-S) &= \Sigma_s (I - \beta^{(sm)}\beta^{(ms)})  \\ 
    &= \Sigma_s (I - \beta^{(sm)}\Sigma_{m|s}^{-1} (\beta^{(sm)})^T Var(S^*-S))  \\
    &= (\Sigma_s^{-1} + \beta^{(sm)}\Sigma_{m|s}^{-1} (\beta^{(sm)})^T)^{-1}
\end{align*}

The result from the main text follows:
\begin{align}
    \mathbb{E}[||S^* - S||^2] &= Tr(Var(S^*-S)) + ||\mathbb{E}[S^*-S]||^2 \\ \notag
    &= Tr(\left( \Sigma_{s}^{-1} + \beta^{(sm)}\Sigma_{m|s}^{-1}(\beta^{(sm)})^T \right)^{-1}) + 0 \\ \notag
    &= \mathcal{O}(J^{-1}),
\end{align}

\section{Bottleneck condition}
\label{sec-SM:bottleneck-condition-derivation}
In the most general setting, we assume that $I \geq 1$, $J \geq 1$, $K \geq 1$, $L \geq 1$, that all noise covariance matrices are full, and that variables are centred on their average value so that intercepts can be omitted. We show that under these conditions, the optimal solution of a model that violates the bottleneck condition $K < \min(I+L,J)$ corresponds to a reparametrization of the Ordinary Least Squares (OLS) optimum.

It has already been established that in the fully unconstrained case, OLS is the maximum likelihood estimator for a linear model of the observed variables, and that it exactly matches the first and second population moments \cite{Hayashi2000OLSisMLE}. 
Although the latent system defined in the main text assumes that linear relations between observed variables are structured by a latent variable $S$, the relations between observed variables remain linear. Therefore, whenever the latent model can attain (a given reparametrization of) the OLS solution, this solution also corresponds to a maximum likelihood estimator of the latent model.

We therefore derive the value of $K$ for which the latent model will necessarily produce a solution equivalent to the OLS solution, while using its own parametric representation. Equivalence here means that the two models reproduce exactly the same population moments through their respective parametrizations.
Here, the latent model is identified from the population moments $\Sigma_{XX},\Sigma_{MM},\Sigma_{YY},\Sigma_{XM},\Sigma_{XY},\Sigma_{MY}$ by the following system of equations:
\begin{align}
    \label{eq:population-moments}
    \Sigma_{XX} &= \Sigma_{x} \\ \notag
    \Sigma_{MM} &= \Sigma_{m|s} + (\beta^{(sm)})^T \Sigma_{s|x} \beta^{(sm)} + (\beta^{(xs)}\beta^{(sm)})^T \Sigma_{x} \beta^{(xs)}\beta^{(sm)}  \\ \notag
    \Sigma_{YY} &= \Sigma_{y|x,s} + (\beta^{(sy)})^T \Sigma_{s|x} \beta^{(sy)} + (\beta^{(xs)}\beta^{(sy)}+\beta^{(xy)})^T \Sigma_{x} (\beta^{(xs)}\beta^{(sy)}+\beta^{(xy)}) \\ \notag
    \Sigma_{XY} &= \Sigma_x (\beta^{(xs)}\beta^{(sy)}+\beta^{(xy)}) \\ \notag
    \Sigma_{XM} &= \Sigma_x \beta^{(xs)}\beta^{(sm)} \\ \notag
    \Sigma_{MY} &= (\beta^{(sm)})^T \Sigma_{s|x} \beta^{(sy)} + (\beta^{(xs)} \beta^{(sm)})^T \Sigma_{x} (\beta^{(xs)}\beta^{(sy)}+\beta^{(xy)})
\end{align}
The OLS solution is reachable when all these equations can be exactly satisfied simultaneously. The equations involving $\Sigma_{XX}, \Sigma_{MM}, \Sigma_{YY}$, and $\Sigma_{XY}$ are straightforward, as these population moments are always reachable from the latent model's parameters in the fully unconstrained case: $\Sigma_{XX}$, $\Sigma_{MM}$ and $\Sigma_{YY}$ can always be exactly modelled by the full matrices $\Sigma_{x}$, $\Sigma_{m|s}$ and $\Sigma_{y|x,s}$ ; $\Sigma_{XY}$ can always be modelled exactly by $\beta^{(xy)}$ because its contribution to $\Sigma_{YY}$ can always be compensated by the full matrix $\Sigma_{y|x,s}$, and because it does not appear in $\Sigma_{MY}$ once $\Sigma_{XY}$ is known. The rank constraint on the joint system Eq.~\eqref{eq:population-moments} induced by the value of $K$ can therefore be expressed in terms of $\Sigma_{XM}$ and $\Sigma_{MY}$ only.

Replacing the expression for $\Sigma_{XM}$ in $\Sigma_{MY}$, then $\Sigma_{MY} = (\beta^{(sm)})^T \Sigma_{s|x} \beta^{(sy)} + (\Sigma_{XM})^T \Sigma_{XX}^{-1} \Sigma_{XY}$ ; the above joint system then reduces to:
\begin{align}
    \label{eq:rank-condition-moments}
    C := \left(
    \begin{aligned}
        \Sigma_{XX}^{-1} \Sigma_{XM}  \\
        \Sigma_{MY}^T - \Sigma_{XY}^T \Sigma_{XX}^{-1} \Sigma_{XM}
    \end{aligned}
    \right)
    &= 
    \left(
    \begin{aligned}
        \beta^{(xs)}  \\
        (\beta^{(sy)})^T \Sigma_{s|x}
    \end{aligned}
    \right) \beta^{(sm)}
\end{align}
The matrix of population moments has $I+L$ rows and $J$ columns ; its rank is therefore upper-bounded by $\min(J,I+L)$. Its parametric representation instead has rank upper-bounded by $\min (J,I+L,K)$. Therefore, a sufficient condition for it to be represented exactly by the latent model is $K \geq \min(I+L,J)$. Conversely, a necessary but not sufficient condition for an approximate solution is $K<\min (J,I+L)$.
It follows that the globally optimal solution of an unconstrained latent linear model with $K \geq \min(J,I+L)$ must necessarily be a reparametrization of the optimal OLS solution.

Note that the exact condition on $K$ for the latent model to be able to reparametrize the OLS solution is $K \geq rank(C)$, with $C$ defined above ; however, the sufficient condition based on rank upper bounds is clearer and sufficient for our demonstration, and is the only one we use in the main text.

\section{Dependence of $S$ on the scale of the outcome in sum-of-variance objectives}
\label{sec-SM:LindquistObj}
Let $\theta$ denote the set of parameters to estimate. Omitting the interprets for simplicity, the joint objective of the model in \cite{Lindquist2023MLHighDimMediation} is:
\begin{align*}
    &\arg \min_\theta ||Y - X \beta^{(xy)} - S \beta^{(sy)}||^2 + ||S - X \beta^{(xs)}||^2 \\
    \text{where } &S = \Phi(M)
\end{align*}
After profiling $S$, this can equivalently be written as an inference problem on $S$ directly:
\begin{align*}
    \arg \min_{S = \Phi(M)} Var(Y|X,S) + Var(S|X)
\end{align*}

The sensitivity of the optimal representation $S$ to the scale of $Y$ is immediate, since the transformation $Y \to cY$ makes $Var(Y|X,S) \to c^2 Var(Y|X,S)$:
\begin{align*}
    &\arg \min_{S = \Phi(M)} Var(S|X) + Var(Y|X,S) \neq \arg \min_{S = \Phi(M)} Var(S|X) + c^2 Var(Y|X,S)
\end{align*}

More importantly, this implies that $\hat{NIE}(cY) \neq c \hat{NIE}(Y)$, where $\hat{NIE}(Y)$ denotes the estimated natural indirect effect associated with a one-unit increase in the exposures; the NIE also depends on the scale of $Y$.



\section{Constrained LVMA is a factor analysis model}
\label{sec-SM:DerkachFA}
We show that under the assumptions of the latent linear model from \cite{Derkach2019HighDimMed}, namely $\Sigma_{m|s} = diag(\sigma_{m_1|s}, ..., \sigma_{m_J|s})$, $I=1$ and $L=1$, a correctly specified LVMA model (full rank coefficient matrices and positive semi-definite covariances) recovers the same latent subspace as a Factor Analysis (FA) model for $M$ whenever the latent space dimension is larger than $K=1$, that is large enough not to constrain the population cross-covariances of the system (see Sec.~\ref{sec-SM:bottleneck-condition-derivation}).

Although the authors in \cite{Derkach2019HighDimMed} work under the assumption that $\Sigma_{s|x}=I$, we chose a representation where the unconditional variance $\Sigma_{s}=I$, without loss of generality.
These formulations are equivalent due to the usual invariance of linear latent-variable models to rescaling and rotations of the latent space ; the system can be equivalently defined in terms of $\Sigma_{s}=I$ or $\Sigma_{s|x}=I$ after applying the appropriate transformations of the latent space. Here, we choose the convention where $\Sigma_{s}=I$ to simplify the derivations below.
Then, the model implies the following population moment equations:
\begin{align}
    \label{eq:XX}
    \Sigma_{XX} &= \Sigma_{x} \\ \notag
    \Sigma_{MM} &= \Sigma_{m|s} + (\beta^{(sm)})^T \underbrace{(\Sigma_{s|x} + (\beta^{(xs)})^T \Sigma_{x} \beta^{(xs)})}_{\Sigma_s=I} \beta^{(sm)} \\
    \label{eq:MM}
    &= \Sigma_{m|s} + (\beta^{(sm)})^T \beta^{(sm)}   \\
    \label{eq:YY}
    \Sigma_{YY} &= \Sigma_{y|x,s} + (\beta^{(sy)})^T \Sigma_{s|x} \beta^{(sy)} + (\beta^{(xs)}\beta^{(sy)}+\beta^{(xy)})^T \Sigma_{x} (\beta^{(xs)}\beta^{(sy)}+\beta^{(xy)}) \\
    \label{eq:XY}
    \Sigma_{XY} &= \Sigma_x (\beta^{(xs)}\beta^{(sy)}+\beta^{(xy)}) \\
    \label{eq:XM}
    \Sigma_{XM} &= \Sigma_x \beta^{(xs)}\beta^{(sm)} \\
    \label{eq:MY}
    \Sigma_{MY} &= (\beta^{(sm)})^T \Sigma_{s|x} \beta^{(sy)} + (\beta^{(xs)} \beta^{(sm)})^T \Sigma_{x} (\beta^{(xs)}\beta^{(sy)}+\beta^{(xy)})
\end{align}
Equation \eqref{eq:MM} is exactly the equation for an FA model applied to $M$, which uniquely identifies the latent subspace. The cross-moment equations can then determine the other population parameters according to this subspace without altering it.
In particular, the FA equation can be solved independently from the rest of the system when the model is correctly specified. 

As shown in Sec.~\ref{sec-SM:bottleneck-condition-derivation}, $K < rank(C)$ induce low-rank constraints, where $rank(C) \leq \min(I+L,J)$, meaning the model can only approximate the population moments $Var(M)$, $Cov(X,M)$ and $Cov(M,Y)$ through joint optimization. 
Therefore, Eq.~\eqref{eq:MM} cannot be solved analytically in isolation: its estimation results from a minimization problem defined by Eq.~\eqref{eq:MM}, Eq.~\eqref{eq:XM} and Eq.~\eqref{eq:MY}, and the resulting latent space no longer matches with that of an FA solution.

In \cite{Derkach2019HighDimMed} however, $K>I=1$ is large enough that the model is not rank-constrained. Then, $\beta^{(sm)}$ and $\Sigma_{m|s}$ can be obtained analytically from Eq.~\eqref{eq:MM} as the solution of a FA model, under suitable identification conditions \cite{Anderson1956StatisticalIdentifiabilityFA}. Once these are known, deriving a closed-form solution for $\beta^{(xs)}, \beta^{(sy)}, \beta^{(xy)}, \Sigma_{x}, \Sigma_{y|x,s}$ is straightforward from the remaining population moment equations, provided the model remains correctly specified.

\begin{figure}[H]
    \centering
    \includegraphics[width=0.45\textwidth]{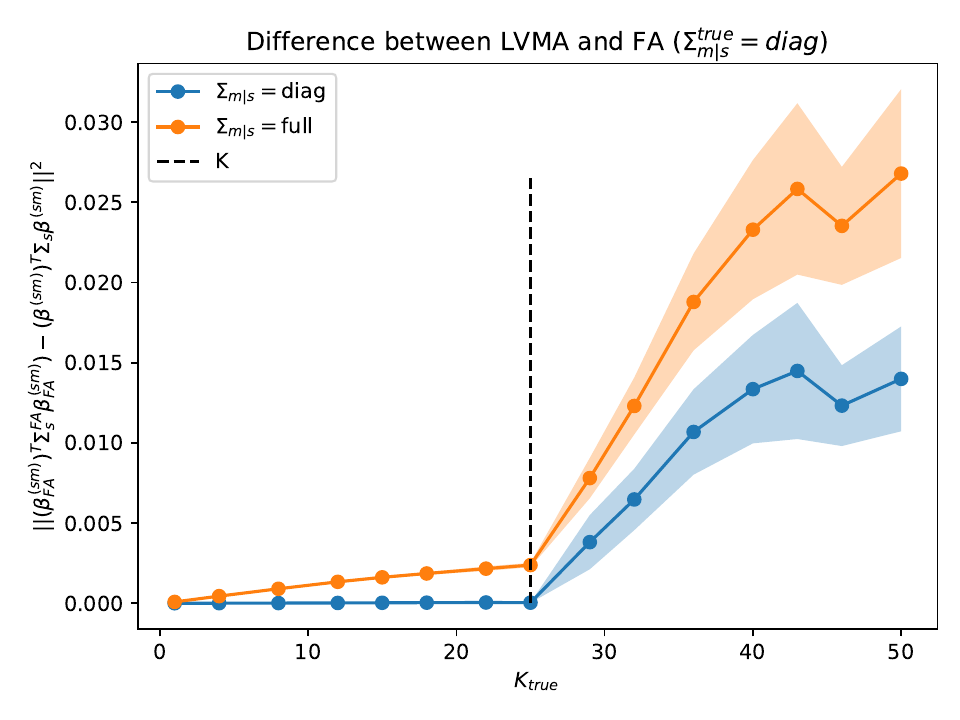}
    \includegraphics[width=0.45\textwidth]{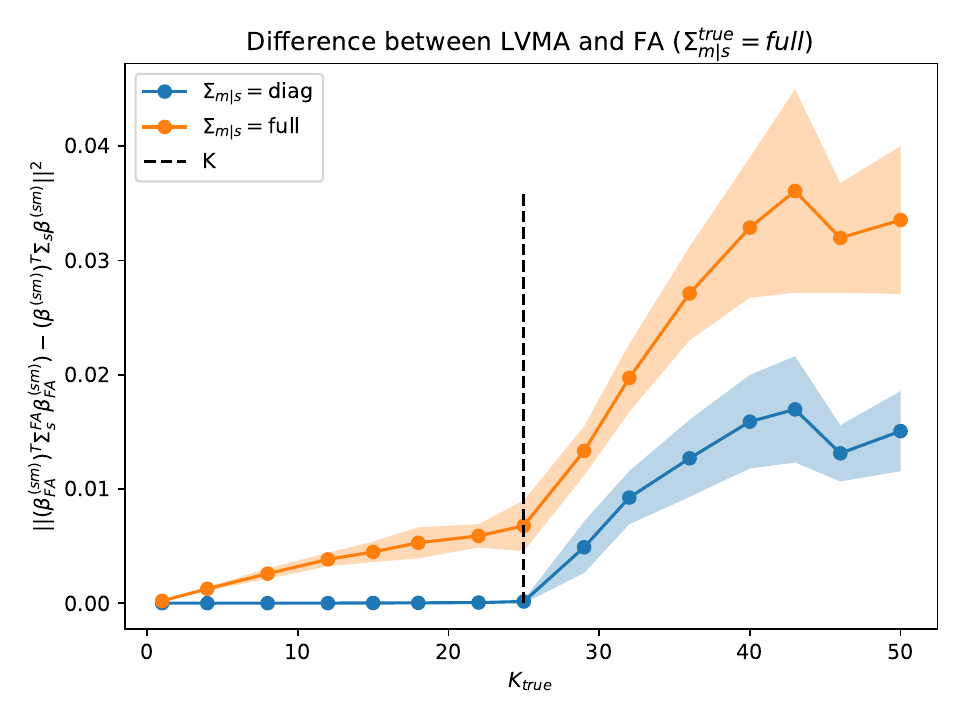}
    \caption{\textbf{LVMA is similar to a factor analysis} --- \textbf{(Left)} Data generated using a diagonal $\Sigma_{m|s}$; \textbf{(Right)} Data generated using a full $\Sigma_{m|s}$. We report the MSE between the covariance of $M$ explained by $S$ from a factor analysis model and the same quantity estimated from one of two latent models, one with diagonal conditional covariance $\Sigma_{m|s}$, and one with full conditional covariance $\Sigma_{m|s}$. We generate $N=1000$ data points with $I=J=50$, $L=1$ and $K_{true}=rank(C)$ (see Eq.~\eqref{eq:rank-condition-moments}) denoting the varying rank of the system used for data generation. We fit the two latent models with $K=25$. 
    Note that $I>1$ was not considered in LVMA, although it provides more informative visualizations for our demonstration. The dashed black line indicates when $K_{true}=K$. The shaded region indicates the standard deviation over 50 runs. This shows that forcing $\Sigma_{m|s}$ to be diagonal makes the latent model equivalent to a factor analysis model when $K \geq K_{true}$.
    }
    \label{fig-Derkach-FA}
\end{figure}

We support this observation numerically in Fig.~\ref{fig-Derkach-FA}-left, where the blue curve shows that the model in \cite{Derkach2019HighDimMed} recovers exactly the same solution as a simple FA model on $M$ when $K \geq K_{true} = rank(C)$, and starts to yield different solutions when $K < K_{true}$. The orange curve shows that the same model with unconstrained noise covariances does not yield the same solution as a FA model, regardless of the rank of $C$. This further demonstrates the importance of choosing $K$ based on the system's dimensions in a way that guarantees the model does not reduce to either FA or ordinary least squares problems.
For exhaustiveness, we also report results when both models are incorrectly specified, that is when the true $\Sigma_{m|s}$ used for data generation is full, in Fig.~\ref{fig-Derkach-FA}-right; results are similar.

\section{E-step developments}
\label{sec-SM:full-Estep}
\subsection{Jensen's inequality}
Here, we derive an expression for a lower bound for $\log p(X,M,Y| \theta)$. Let $q(S)$ denote any distribution of $S$:
\begin{align}
    \label{eq:JensenIneq}
    \log p(X,M,Y| \theta) &= \log p(X,M,Y| \theta) \underbrace{\int_S q(S)dS}_{=1} \\
    \notag
    &= \int_S q(S) \log p(X,M,Y| \theta)dS  \\
    \notag
    &= \int_S q(S) \log \left( \frac{q(S)p(S,X,M,Y| \theta)}{q(S)p(S,X,M,Y| \theta)}p(X,M,Y| \theta) \right) dS  \\
    \notag
    &= \int_S q(S) \log \frac{p(S,X,M,Y| \theta)}{q(S)}dS + \int_S q(S) \log \frac{q(S)p(X,M,Y| \theta)}{p(S,X,M,Y| \theta)}dS \\
    \notag
    &= \int_S q(S) \log \frac{p(S,X,M,Y| \theta)}{q(S)}dS + \underbrace{\int_S q(S) \log \frac{q(S)}{p(S|X,M,Y, \theta)}dS}_{KL(q(S)||p_\theta (S \vert X,M,Y)) \geq 0} \\
    \notag
    & \geq \int_S q(S) \log \frac{p(S,X,M,Y| \theta)}{q(S)}dS
\end{align}
where the Kullback-Leibler divergence is $KL(q_{\theta, X,M,Y} (\cdot)||p_\theta (\cdot \vert X,M,Y)):=\int_S q(S)\log\frac{q_{\theta,o}(s)}{p_\theta(s \vert o)}dS\geq 0$ and $q(S)$ is any distribution of $S$.

Alternatively, using Jensen's inequality ($\mathbb{E}[\log x] \leq \log \mathbb{E}[x]$) directly:
\begin{align*}
    \log p(X,M,Y| \theta) = \underbrace{\log\int_S q(S)\frac{p(S,X,M,Y| \theta)}{q(S)} dS}_{\log \mathbb{E}[\frac{p(S,X,M,Y| \theta)}{q(S)}]} \geq \underbrace{\int_S q(S) \log \frac{p(S,X,M,Y| \theta)}{q(S)}dS}_{\mathbb{E}[\log \frac{p(S,X,M,Y| \theta)}{q(S)}]}
\end{align*}

\subsection{Lower-bound maximization}
Here, we demonstrate that the lower bound in Eq.~\eqref{eq:JensenIneq} is equal to $\log p(X,M,Y| \theta)$ when $q(S) = q_{\theta,X,M,Y}(S)=\frac{p(S, X,M,Y| \theta)}{p(X,M,Y| \theta)}$ --it is maximized.
\begin{align}
    \int_S q_{\theta,X,M,Y}(S) \log \frac{p(S,X,M,Y| \theta)}{q_{\theta,X,M,Y}(S)} dS &= \int_S \frac{p(S, X,M,Y| \theta)}{p(X,M,Y| \theta)} \log \frac{p(S,X,M,Y| \theta)}{\frac{p(S, X,M,Y| \theta)}{p(X,M,Y| \theta)}} dS \\
    \notag
    &= \int_S \frac{p(S, X,M,Y| \theta)}{p(X,M,Y| \theta)} \log p(X,M,Y| \theta) dS \\
    \notag
    &= \frac{\log p(X,M,Y| \theta)}{p(X,M,Y| \theta)} \underbrace{\int_S p(S, X,M,Y| \theta) dS}_{:=p(X,M,Y| \theta)} \\
    \notag
    &= \log p(X,M,Y| \theta)
\end{align}

The equality naturally follows (because a lower bound cannot be larger than the bounded quantity):
\begin{align}
    \log p(X,M,Y| \theta) &= \int_S q_{\theta,X,M,Y}(S) \log \frac{p(S,X,M,Y| \theta)}{q_{\theta,X,M,Y}(S)}dS \\  \notag
    \text{when} \qquad q_{\theta,X,M,Y}(S)&=\frac{p(S, X,M,Y| \theta)}{p(X,M,Y| \theta)}
\end{align}

\section{Full M-step derivation}
\label{sec-SM:full-Mstep}
Intercepts are omitted for notational clarity; they can be included in the usual way by augmenting the relevant design matrix with a column of ones.
\subsection{LaMe-vX}

The solutions for the LaMe-vX model are summarized below:
\begin{subequations}
    \label{eq:Mstep-updates}
\begin{align}
    &\beta^{(xs)} = \left(X^TX\right)^{-1} X^T \overline{S} \qquad\qquad
    &&\beta^{(sm)} = \left( \overline{S^T S} \right)^{-1} \overline{S}^T M \\
    \notag
    &\beta^{(xy)} = \left(X^T (I - P_S) X \right)^{-1} X^T (I - P_S) Y  \qquad\qquad
    &&\beta^{(sy)} = \left(\overline{S^T S}\right)^{-1}\overline{S}^T ( Y - X \beta^{(xy)})
\end{align}
\begin{align}
    \Sigma_{x} &= \frac{1}{N} X^T X \\
    \notag
    \Sigma_{s|x} &= \frac{1}{N} \left( \overline{S^T S} + \left( \overline{S} - X\beta^{(xs)}\right)^T \left( \overline{S} - X\beta^{(xs)} \right) - \overline{S}^T\overline{S} \right) \\
    \notag
    \Sigma_{m|s} &= \frac{1}{N} \left((\beta^{(sm)})^T (\overline{S^T S})\beta^{(sm)} + \left(M -\overline{S}\beta^{(sm)} \right)^T \left(M -\overline{S}\beta^{(sm)} \right) - (\overline{S}\beta^{(sm)})^T(\overline{S}\beta^{(sm)}) \right) \\
    \notag
    \Sigma_{y|x,s} &= \frac{1}{N} \left( (\beta^{(sy)})^T(\overline{S^T S})\beta^{(sy)} + (Y-X\beta^{(xy)}-\overline{S}\beta^{(sy)})^T(Y-X\beta^{(xy)}-\overline{S}\beta^{(sy)}) - (\overline{S} \beta^{sy})^T(\overline{S} \beta^{sy}) \right)
\end{align}
\end{subequations}
\noindent
where $P_S := \overline{S}\big(\overline{S^T S} \big)^{-1} \overline{S}^T$ is the estimated latent projection-like matrix.

\subsubsection{A, B, C}
\begin{align*}
    p(S,X,M,Y| \theta) &:= e^{-\frac{1}{2}\langle S^T S, A\rangle+\langle S, B \rangle + C} \\
    A &:= \Sigma_{s|x}^{-1} + \beta^{(sm)}\Sigma_{m|s}^{-1}(\beta^{(sm)})^T + \beta^{(sy)}\Sigma_{y|x,s}^{-1}(\beta^{(sy)})^T \\
    B &:= Y\Sigma_{y|x,s}^{-1}(\beta^{(sy)})^T + X\beta^{(xs)}\Sigma_{s|x}^{-1} -  X\beta^{(xy)}\Sigma_{y|x,s}^{-1}(\beta^{(sy)})^T + M\Sigma_{m|s}^{-1}(\beta^{(sm)})^T \\
    C &:= - \frac{1}{2}\langle X^TX, \left( \Sigma_x^{-1} + \beta^{(xs)}\Sigma_{s|x}^{-1} (\beta^{(xs)})^T + \beta^{(xy)}\Sigma_{y|x,s}^{-1}(\beta^{(xy)})^T \right)\rangle \\
    &- \frac{1}{2} \langle M^TM, \Sigma_{m|s}^{-1} \rangle - \frac{1}{2} \langle Y^T Y, \Sigma_{y|x,s}^{-1} \rangle \\
    &+ \langle X^T Y, \beta^{(xy)}\Sigma_{y|x,s}^{-1} \rangle - \frac{N}{2} \log \left( (2 \pi)^{I+J+K+L} |\Sigma_x||\Sigma_{s|x}||\Sigma_{m|s}||\Sigma_{y|x,s}| \right)
\end{align*}

Let $\bar \ell_{\theta}$ denote the expected incomplete log-likelihood (see the main text):
\begin{align*}
    \bar \ell_\theta= \int_S q_{\theta,X,M,Y}(S) \log \frac{p(S,X,M,Y| \theta)}{q_{\theta,X,M,Y}(S)} dS
\end{align*}

\subsubsection{$x \rightarrow s$}
For $\beta^{(xs)}$, $\Sigma_{s|x}$:
\begin{align*}
    \partial_{\beta^{(xs)}} \bar \ell_\theta &= \sum_i^N \int_{\mathcal S} q_{\theta,x_i,m_i,y_i}(s_i) \partial_{\beta^{(xs)}}\left(-\frac{1}{2}(s_i-x_i\beta^{(xs)}) \Sigma_{s|x}^{-1} (s_i-x_i\beta^{(xs)})^T \right)ds_i = 0\\
    &= \sum_i^N \int_{\mathcal S} q_{\theta,x_i,m_i,y_i}(s_i) \partial_{\beta^{(xs)}}\left((s_i-x_i\beta^{(xs)}) \Sigma_{s|x}^{-1} (s_i-x_i\beta^{(xs)})^T \right)ds_i = 0\\
    &= \sum_i^N \int_{\mathcal S} q_{\theta,x_i,m_i,y_i}(s_i) \left( (-x_i^T )(s_i-x_i\beta^{(xs)})\Sigma_{s|x}^{-1} \right) ds_i = 0\\
    \Leftrightarrow \beta^{(xs)} &= \left(\sum_i^N x_i^T x_i \right)^{-1}\sum_i^N x_i^T \overline{s_i} \\
    &= (X^TX)^{-1} X^T \overline{S} \\
    \partial_{\Sigma_{s|x}} \bar \ell_\theta &= \sum_i^N \int_{\mathcal S} q_{\theta,x_i,m_i,y_i}(s_i) \partial_{\Sigma_{s|x}}\left(-\frac{1}{2}(s_i-x_i\beta^{(xs)}) \Sigma_{s|x}^{-1} (s_i-x_i\beta^{(xs)})^T - \frac{1}{2}\ln(|\Sigma_{s|x}|) \right)ds_i = 0  \\
    &= \sum_i^N \int_{\mathcal S} q_{\theta,x_i,m_i,y_i}(s_i) \left(\Sigma_{s|x}^{-1}(s_i-x_i\beta^{(xs)})^T (s_i-x_i\beta^{(xs)})\Sigma_{s|x}^{-1} - \Sigma_{s|x}^{-1} \right)ds_i = 0  \\
    \Leftrightarrow \Sigma_{s|x} &= \frac{1}{N} \sum_i^N \int_S q_{\theta,x_i,m_i,y_i}(s_i) \left((s_i-x_i\beta^{(xs)})^T (s_i-x_i\beta^{(xs)}) \right) ds_i \\
    &= \frac{1}{N} \sum_i^N \overline{s_i^Ts_i} - \overline{s_i}^T(x_i\beta^{(xs)}) - (x_i\beta^{(xs)})^T\overline{s_i} + (x_i\beta^{(xs)})^T(x_i\beta^{(xs)}) \\
    &= \frac{1}{N} \left( \overline{S^T S} - \overline{S}^T X\beta^{(xs)} - (X\beta^{(xs)})^T\overline{S} + (X\beta^{(xs)})^T(X\beta^{(xs)})  \right) \\
    &= \frac{1}{N} \left( \overline{S^T S} + \left( \overline{S} - X\beta^{(xs)} \right)^T \left( \overline{S} - X\beta^{(xs)}\right) - \overline{S}^T\overline{S} \right)
\end{align*}

\subsubsection{$s \rightarrow m$}
For $\beta^{(sm)}$, $\Sigma_{m|s}$:
\begin{align*}
    \partial_{\beta^{(sm)}} \bar \ell_\theta &= \sum_i^N \int_{\mathcal S} q_{\theta,x_i,m_i,y_i}(s_i) (-2 s_i^T )(m_i-s_i\beta^{(sm)})\Sigma_{m|s}^{-1} ds_i = 0  \\
    \Leftrightarrow \beta^{(sm)} &= \left( \sum_i^N \overline{s_i^Ts_i} \right)^{-1} \sum_i^N \overline{s_i}^T m_i \\
    &= \left( \overline{S^T S} \right)^{-1} \overline{S}^T M \\
    \partial_{\Sigma_{m|s}} \bar \ell_\theta &= \sum_i^N \int_{\mathcal S} q_{\theta,x_i,m_i,y_i}(s_i) \left(-\Sigma_{m|s}^{-1}(m_i-s_i\beta^{(sm)})^T (m_i-s_i\beta^{(sm)})\Sigma_{m|s}^{-1} + \Sigma_{m|s}^{-1} \right)ds_i = 0  \\
    \Leftrightarrow \Sigma_{m|s} &= \frac{1}{N} \sum_i^N \int_S q_{\theta,x_i,m_i,y_i}(s_i)(m_i-s_i\beta^{(sm)})^T (m_i-s_i\beta^{(sm)}) ds_i \\
    &= \frac{1}{N}\sum_i^N m_i^T m_i - m_i^T(\overline{s_i}\beta^{(sm)}) - (\overline{s_i}\beta^{(sm)})^Tm_i + \int_{\mathcal S} q_{\theta,x_i,m_i,y_i}(s_i)(s_i \beta^{(sm)}))^T (s_i\beta^{(sm)})ds \\
    &= \frac{1}{N}\sum_i^N m_i^T m_i - m_i^T(\overline{s_i}\beta^{(sm)}) - (\overline{s_i}\beta^{(sm)})^T m_i + (\beta^{(sm)})^T \overline{S^2_i}\beta^{(sm)} \\
    &= \frac{1}{N} \left(M^T M - M^T\overline{S}\beta^{(sm)} - (\overline{S}\beta^{(sm)})^T M + (\beta^{(sm)})^T (\overline{S^T S})\beta^{(sm)} \right) \\
    &= \frac{1}{N} \left((\beta^{(sm)})^T (\overline{S^T S})\beta^{(sm)} + \left(M -\overline{S}\beta^{(sm)} \right)^T \left(M -\overline{S}\beta^{(sm)} \right) - (\overline{S}\beta^{(sm)})^T(\overline{S}\beta^{(sm)}) \right)
\end{align*}

\subsubsection{$x, s \rightarrow y$}
For $\beta^{(xy)}$, $\beta^{(sy)}$:
\begin{align*}
    \partial_{\beta^{(xy)}} \bar \ell_\theta &= \sum_i^N \int_{\mathcal S} q_{\theta,x_i,m_i,y_i}(s_i) \left(  -2x_i^T(Y_i-x_i\beta^{(xy)}-s_i\beta^{(sy)})\Sigma_{y|x,s}^{-1} \right)ds_i = 0  \\
    \Leftrightarrow \beta^{(xy)} &= \left( \sum_i^N x_i^Tx_i \right)^{-1}\left(\sum_i^N x_i^T Y_i - x_i^T\overline{s_i}\beta^{(sy)} \right) \\
    &= \left( X^TX \right)^{-1}X^T Y - \left( X^TX \right)^{-1}\left( X^T\overline{S}\right)\beta^{(sy)} \\
    \partial_{\beta^{(sy)}} \bar \ell_\theta &= \sum_i^N \int_{\mathcal S} q_{\theta,x_i,m_i,y_i}(s_i) \left(  -2s_i^T(Y_i-x_i\beta^{(xy)}-s_i\beta^{(sy)})\Sigma_{y|x,s}^{-1} \right)ds_i = 0  \\
    \Leftrightarrow \beta^{(sy)} &= \left( \sum_i^N \overline{s_i^Ts_i} \right)^{-1}\left(\sum_i^N \overline{s_i}^T Y_i - \overline{s_i}^Tx_i \beta^{(xy)} \right) \\
    &= \left( \overline{S^T S} \right)^{-1}\overline{S}^T Y - \left( \overline{S^T S} \right)^{-1}\left(\overline{S}^TX \right) \beta^{(xy)} \\
\end{align*}

Thus:
\begin{align*}
    &\underbrace{\left(
    \begin{aligned}
        &X^TX & X^T\overline{S} \\
        &\overline{S}^TX  & \overline{S^T S}
    \end{aligned}
    \right)}_{A}
    \underbrace{\left(
    \begin{aligned}
        &\beta^{(xy)}\\
        &\beta^{(sy)}
    \end{aligned}
    \right)}_{\Theta}
    =
    \underbrace{\left(
    \begin{aligned}
        &X^TY\\
        &\overline{S}^T Y
    \end{aligned}
    \right)}_{r}
\end{align*}
The solution is $\Theta = A^{-1}r$.

Finally for $\Sigma_{y|x,s}$:
\begin{align*}
    &\partial_{\Sigma_{y|x,s}} \bar \ell_\theta = \\
    &\sum_i^N \int_{\mathcal S} q_{\theta,x_i,m_i,y_i}(s_i) \left( \Sigma_{y|x,s}^{-1}(Y_i-x_i\beta^{(xy)}-s_i\beta^{(sy)})^T(Y_i-x_i\beta^{(xy)}-s_i\beta^{(sy)}) \Sigma_{y|x,s}^{-1}  -  \Sigma_{y|x,s}^{-1}  \right)ds_i \\
    &= 0  \\
    \Leftrightarrow \Sigma_{y|x,s} &= \frac{1}{N} \sum_i^N \int_S q_{\theta,x_i,m_i,y_i}(s_i)(Y_i-x_i\beta^{(xy)}-s_i\beta^{(sy)})^T(Y_i-x_i\beta^{(xy)}-s_i\beta^{(sy)}) ds_i \\
    &= \frac{1}{N} ( (Y-X\beta^{(xy)}-\overline{S}\beta^{(sy)})^T(Y-X\beta^{(xy)}-\overline{S}\beta^{(sy)}) - (\overline{S} \beta^{(sy)})^T(\overline{S} \beta^{(sy)})) \\
    &+ \frac{1}{N} (\beta^{(sy)})^T \overline{S^T S}\beta^{(sy)}
\end{align*}

\subsection{LaMe-vM}
\subsubsection{A, B, C}
\begin{align*}
    p(S,X,M,Y| \theta) &:= e^{-\frac{1}{2}\langle S^T S, A\rangle+\langle S, B \rangle + C} \\
    A &:= \Sigma_{s|m}^{-1} + \beta^{(sx)}\Sigma_{x|s}^{-1}(\beta^{(sx)})^T + \beta^{(sy)}\Sigma_{y|x,s}^{-1}(\beta^{(sy)})^T \\
    B &:= M\beta^{(ms)}\Sigma_{s|m}^{-1} + X\Sigma_{x|s}^{-1}(\beta^{(sx)})^T - X\beta^{(xy)} \Sigma_{y|x,s}^{-1} (\beta^{(sy)})^T + Y \Sigma_{y|x,s}^{-1} (\beta^{(sy)})^T \\
    C &:= -\frac{1}{2} \langle X^TX, \left( \Sigma_{x|s}^{-1} + \beta^{(xy)}\Sigma_{y|x,s}^{-1}(\beta^{(xy)})^T \right)\rangle \\
    & -\frac{1}{2} \langle M^TM, \left( \Sigma_m^{-1} + \beta^{(ms)}\Sigma_{s|m}^{-1}(\beta^{(ms)})^T \right) \rangle-\frac{1}{2} \langle Y^TY, \Sigma_{y|x,s}^{-1} \rangle \\
    & + \langle X^T Y, \beta^{(xy)}\Sigma_{y|x,s}^{-1} \rangle - \frac{N}{2} \log \left( (2 \pi)^{I+J+K+L} |\Sigma_m||\Sigma_{s|m}||\Sigma_{x|s}||\Sigma_{y|x,s}| \right)
\end{align*} 

\subsubsection{$m \rightarrow s$}
For $\beta^{(ms)}$, $\Sigma_{s|m}$:
\begin{align*}
    \partial_{\beta^{(ms)}} \bar \ell_\theta(O) &= \sum_i^N \int_{\mathcal S} q_{\theta,o_i}(s_i) \partial_{\beta^{(ms)}}\left(-\frac{1}{2}(s_i-m_i\beta^{(ms)}) \Sigma_{s|m}^{-1} (s_i-m_i\beta^{(ms)})^T \right)ds_i = 0\\
    &= \sum_i^N \int_{\mathcal S} q_{\theta,o_i}(s_i) \left( (m_i^T)(s_i-m_i\beta^{(ms)})\Sigma_{s|m}^{-1} \right) ds_i = 0\\
    &= (\sum_i^N m_i^T m_i) \beta^{(ms)} - \sum_i^N m_i^T (\overline{s_i}) = 0 \\
    \Leftrightarrow \beta^{(ms)} &= \left( \sum_i^N m_i^T m_i \right)^{-1}\sum_i^N m_i^T (\overline{s_i}) \\
    &= (M^TM)^{-1}M^T \overline{S} \\
    \partial_{\Sigma_{s|m}} \bar \ell_\theta(O) &= \sum_i^N \int_{\mathcal S} q_{\theta,o_i}(s_i) \partial_{\Sigma_{s|m}}\left(-(s_i-m_i\beta^{(ms)}) \Sigma_{s|m}^{-1} (s_i-m_i\beta^{(ms)})^T - \ln (|\Sigma_{s|m}|) \right)ds_i = 0  \\
    &= \sum_i^N \int_{\mathcal S} q_{\theta,o_i}(s_i) \left(-\Sigma_{s|m}^{-1}(s_i-m_i\beta^{(ms)})^T (s_i-m_i\beta^{(ms)})\Sigma_{s|m}^{-1} + \Sigma_{s|m}^{-1} \right)ds_i = 0  \\
    \Leftrightarrow \Sigma_{s|m} &= \frac{1}{N} \sum_i^N \int_S q_{\theta,o_i}(s_i) \left( (s_i-m_i\beta^{(ms)})^T (s_i-m_i\beta^{(ms)}) \right) ds_i \\
    &= \frac{1}{N} \sum_i^N \overline{s_i^Ts_i} + (\overline{s_i}-m_i\beta^{(ms)})^T (\overline{s_i}-m_i\beta^{(ms)}) - \overline{s_i}^T \overline{s_i}  \\
    &= \frac{1}{N} \left( \overline{S^T S} + (\overline{S}-M\beta^{(ms)})^T (\overline{S}-M\beta^{(ms)}) - \overline{S}^T \overline{S}\right)
\end{align*}

\subsubsection{$s \rightarrow x$}
For $\beta^{(sx)}$, $\Sigma_{x|s}$:
\begin{align*}
    \partial_{\beta^{(sx)}} \bar \ell_\theta(O) &= \sum_i^N \int_{\mathcal S} q_{\theta,o_i}(s_i) (-2 s_i^T )(x_i-s_i\beta^{(sx)})\Sigma_{x|s}^{-1} ds_i = 0  \\
    \Leftrightarrow \beta^{(sx)} &= \left( \sum_i^N \overline{s_i^Ts_i} \right)^{-1}\sum_i^N \overline{s_i}^T (x_i ) \\
    &= \left( \overline{S^T S} \right)^{-1} \overline{S}^T X \\
    \\
    \partial_{\Sigma_{x|s}} \bar \ell_\theta(O) &= \sum_i^N \int_{\mathcal S} q_{\theta,o_i}(s_i) \left(-\Sigma_{x|s}^{-1}(x_i-s_i\beta^{(sx)})^T (x_i-s_i\beta^{(sx)})\Sigma_{x|s}^{-1} + \Sigma_{x|s}^{-1} \right)ds_i = 0  \\
    \Leftrightarrow \Sigma_{x|s} &= \frac{1}{N} \sum_i^N \int_S q_{\theta,o_i}(s_i)(x_i-s_i\beta^{(sx)})^T (x_i-s_i\beta^{(sx)}) ds_i \\
    &=  \frac{1}{N}\sum_i^N (\beta^{(sx)})^T\overline{s_i^T s_i}\beta^{(sx)} + (x_i-\overline{s_i}\beta^{(sx)})^T (x_i-\overline{s_i}\beta^{(sx)}) - (\overline{s_i}\beta^{(sx)})^T (\overline{s_i}\beta^{(sx)}) \\
    &=  \frac{1}{N} \left( (\beta^{(sx)})^T (\overline{S^T S}) \beta^{(sx)} + (X-\overline{S}\beta^{(sx)})^T (X-\overline{S}\beta^{(sx)}) - (\overline{S}\beta^{(sx)})^T (\overline{S}\beta^{(sx)}) \right) \\
\end{align*}

\subsubsection{$x, s \rightarrow y$}

For $\beta^{(xy)}$, $\beta^{(sy)}$:
\begin{align*}
    \partial_{\beta^{(xy)}} \bar \ell_\theta(O) &= \sum_i^N \int_{\mathcal S} q_{\theta,o_i}(s_i) \left(  -2x_i^T(Y_i-x_i\beta^{(xy)}-s_i\beta^{(sy)})\Sigma_{y|x,s}^{-1} \right)ds_i = 0  \\
    \Leftrightarrow \beta^{(xy)} &= \left( \sum_i^N x_i^Tx_i \right)^{-1}\left(\sum_i^N x_i^T Y_i - x_i^T\overline{s_i}\beta^{(sy)} \right) \\
    &= \left( X^TX \right)^{-1}X^T Y- \left( X^TX \right)^{-1}\left( X^T\overline{S}\right)\beta^{(sy)} \\
    \partial_{\beta^{(sy)}} \bar \ell_\theta(O) &= \sum_i^N \int_{\mathcal S} q_{\theta,o_i}(s_i) \left(  -2s_i^T(Y_i-x_i\beta^{(xy)}-s_i\beta^{(sy)} )\Sigma_{y|x,s}^{-1} \right)ds_i = 0  \\
    \Leftrightarrow \beta^{(sy)} &= \left( \sum_i^N \overline{s_i^Ts_i} \right)^{-1}\left(\sum_i^N \overline{s_i}^T Y_i - \overline{s_i}^Tx_i \beta^{(xy)} \right) \\
    &= \left( \overline{S^T S} \right)^{-1}\overline{S}^T Y - \left( \overline{S^T S} \right)^{-1}\left(\overline{S}^TX \right) \beta^{(xy)}
\end{align*}

Thus:
\begin{align*}
    &\underbrace{\left(
    \begin{aligned}
        &X^TX & X^T\overline{S}\\
        &\overline{S}^TX  & \overline{S^T S}\\
    \end{aligned}
    \right)}_{A}
    \underbrace{\left(
    \begin{aligned}
        &\beta^{(xy)}\\
        &\beta^{(sy)}
    \end{aligned}
    \right)}_{\Theta}
    =
    \underbrace{\left(
    \begin{aligned}
        &X^TY\\
        &\overline{S}^T Y
    \end{aligned}
    \right)}_{r}
\end{align*}

The solution is $\Theta = A^{-1}r$.

Finally for $\Sigma_{y|x,s}$:
\begin{align*}
    &\partial_{\Sigma_{y|x,s}} \bar \ell_\theta(O) \\
    &= \sum_i^N \int_{\mathcal S} q_{\theta,o_i}(s_i) \left( \Sigma_{y|x,s}^{-1}(Y_i-x_i\beta^{(xy)}-s_i\beta^{(sy)})^T(Y_i-x_i\beta^{(xy)}-s_i\beta^{(sy)}) \Sigma_{y|x,s}^{-1}  -  \Sigma_{y|x,s}^{-1}  \right)ds_i \\
    &= 0  \\
    \Leftrightarrow \Sigma_{y|x,s} &= \frac{1}{N} \sum_i^N \int_S q_{\theta,o_i}(s_i)(Y_i-x_i\beta^{(xy)}-s_i\beta^{(sy)})^T(Y_i-x_i\beta^{(xy)}-s_i\beta^{(sy)}) ds_i \\
    &= \frac{1}{N} ( (Y-X\beta^{(xy)}-\overline{S}\beta^{(sy)})^T(Y-X\beta^{(xy)}-\overline{S}\beta^{(sy)}))\\
    &-\frac{1}{N} ( (\overline{S} \beta^{(sy)})^T(\overline{S} \beta^{(sy)}) + (\beta^{(sy)})^T \overline{S^T S}\beta^{(sy)} )
\end{align*}

\section{Model's features and remarks}
\label{sec-SM:model-features}
\subsubsection{Rotational invariance}
Latent mediation models exhibit rotational invariance: for any invertible linear transformation $Q \in \mathbb{R}^{K \times K}$, the sets $(S$, $\beta^{(xs)}$, $\beta^{(sm)}$, $\beta^{(sy)}$, $\Sigma_{s|x})$ and $(SQ$, $\beta^{(xs)}Q$, $Q^{-1}\beta^{(sm)}$, $Q^{-1} \beta^{(sy)}$, $Q^{T} \Sigma_{s|x} Q )$ induce exactly the same likelihood. Although ordering conventions can fix a canonical rotation post-hoc (e.g. in PCA), the above rotational invariance remains. Note that, despite the name, rotational invariance is not restricted to orthogonal rotations; any invertible linear transform $Q$ used in the transformations above leaves the likelihood unchanged.

This implies that latent components (and their corresponding parameters) are not uniquely defined. Importantly, once the noise-signal invariance has been fixed by choosing a canonical representation, this non-identifiability is confined to the latent representation itself. The induced mappings between observed variables, such as the path coefficients $\beta^{(xm)}=\beta^{(xs)}\beta^{(sm)}$, remain identifiable. Consequently, the marginalized mediation model $$p(X,M,Y)=\int_S p(S,X,M,Y)dS$$ remains identifiable.

\subsubsection{Regularized versions}
Regularization can be introduced by changing the prior $p(\theta)$ on the parameters in the log-likelihood, which amounts to adding penalty terms. 
Here we present $\ell_1$ and $\ell_2$ regularizations, which both modify the definition of $C$ in the E-step as well as the optimization problem solved in the M-step. We describe below how such penalties can be applied to any coefficient matrix $\beta^{(uv)} \in \{ \beta^{(xs)}, \beta^{(sm)}, \beta^{(sy)}, \beta^{(xy)} \}$:

\textbf{$\ell_2$ (Ridge) regularization} corresponds to placing a zero-mean Gaussian prior with covariance $\Sigma_{v|u}$ on $\beta^{(uv)}$, which amounts to setting $p(\beta^{(uv)}) \propto e^{-\frac{\lambda_2}{2} \langle (\beta^{(uv)})^T\beta^{(uv)}, \Sigma_{v|u}^{-1} \rangle }$. As a result, the term $C$ in the log-likelihood becomes $C_{reg} = C - \frac{\lambda_2}{2} \langle (\beta^{(uv)})^T\beta^{(uv)}, \Sigma_{v|u}^{-1} \rangle$, and the update equations become $\beta^{(uv)}_{reg} = (U^T U + \lambda_2 I)^{-1}U^T V$ with corresponding updates of the noise covariance matrices. Importantly, the M-step remains available in closed form under $\ell_2$ regularization. Note that this is equivalent to simply adding a penalty term $-\frac{\lambda_2}{2} \langle (\beta^{(uv)})^T\beta^{(uv)}, \Sigma_{v|u}^{-1} \rangle$ to the log-likelihood.

\textbf{$\ell_1$ (LASSO) regularization} corresponds to placing a zero-mean Laplace prior on $\beta^{(uv)}$, which amounts to setting $p(\beta^{(uv)}) \propto e^{-\lambda_1 \sum_{ij} |\beta^{(uv)}_{ij}|}$. As a result, the term $C$ in the log-likelihood becomes $C_{reg} = C - \lambda_1 \sum_{ij} |\beta^{(uv)}_{ij}|$. In contrast to the $\ell_2$ case, the update equations have no analytical solution and must be solved numerically. 
Notably, $\ell_1$ regularization can partially break rotational invariance by favouring sparse orientations of the latent space. Note that this is equivalent to simply adding a penalty term $-\lambda_1 \sum_{ij} |\beta^{(uv)}_{ij}|$ to the log-likelihood.

\subsubsection{Log-normally distributed variables}
Real-world data are not always well approximated by Gaussian distributions. 
When outcomes are strictly positive and exhibit right skewness, as in survival analysis for example, the log-normal distribution can be a more appropriate modelling choice.
The LaMe framework naturally accommodates log-normally distributed variables through a simple reparametrization. Let $Z \in \{ X, M, Y\}$ denote a strictly positive variable assumed to follow a log-normal distribution. The associated transformations are then $Z \rightarrow \log Z$ throughout the model, and $\log p(X,M,Y| \theta) \rightarrow \log p(X,M,Y| \theta) - \sum_{i,j} \log Z_{ij}$, where the summation runs over all the elements of $Z$.




\section{Likelihood-equivalent reparametrizations}
\label{sec-SM:exhaustive-reparam}
Starting from the observed moments parametrization from Section~\ref{sec-SM:bottleneck-condition-derivation}, we allow for arbitrary reparametrizations of the parameters related to $S$, namely $(\beta^{(xs)}, \beta^{(sm)}, \beta^{(sy)}, \Sigma_{s|x})$, noted $(\beta^{(xs)}_*, \beta^{(sm)}_*, \beta^{(sy)}_*, \Sigma_{s|x}^*)$.
These transformed parameters describe every possible reparametrization, including translations and changes of rank. Remaining parameters $\beta^{(xy)}$, $\Sigma_x$, $\Sigma_{m|s}$ and $\Sigma_{y|x,s}$ must be transformed accordingly to preserve the observed moments. 

Now, we assume that the bottleneck assumption holds (see Section~\ref{sec-SM:bottleneck-condition-derivation}), meaning:
\begin{align}
    \label{eq:rank-condition-moments-bis}
    \mathrm{rank}(C) = \mathrm{rank}\left(
    \begin{aligned}
        \Sigma_{XX}^{-1} \Sigma_{XM}  \\
        \Sigma_{MY}^T - \Sigma_{XY}^T \Sigma_{XX}^{-1} \Sigma_{XM}
    \end{aligned}
    \right)
    &\geq 
    \mathrm{rank}\left(
    \begin{aligned}
        \beta^{(xs)} \beta^{(sm)} \\
        (\beta^{(sy)})^T \Sigma_{s|x}\beta^{(sm)}
    \end{aligned}
    \right) = K
\end{align}
This constraint means that all latent dimensions $K$ are used to approximate the observed moments in $C$. Therefore, both factors $(\beta^{(xs)}, (\beta^{(sy)})^T \Sigma_{s|x})^T$ and $\beta^{(sm)}$ above must have rank $K$. Any other factorization of the same $C$ through $K$ dimensions must consequently differ only by an invertible transformation. 

In particular, $\beta^{(sm)}_*$ must also have rank $K$ from the condition above. It follows that $(\beta^{(sm)}_* (\beta^{(sm)}_*)^T)$ is invertible, meaning $R:= (\beta^{(sm)}_*)^T (\beta^{(sm)}_* (\beta^{(sm)}_*)^T)^{-1}$ exists and satisfies $\beta^{(sm)}_* R = I_K$. Therefore, right-multiplying the parametric approximation of $C$ by $R$ and reminding that likelihood equivalence implies that this representation is preserved by equivalent transformations, then:
\begin{align*}
    \left(
    \begin{aligned}
        \beta^{(xs)}  \\
        (\beta^{(sy)})^T \Sigma_{s|x}
    \end{aligned}
    \right) \beta^{(sm)} R
    = \left(
    \begin{aligned}
        \beta^{(xs)}_*  \\
        (\beta^{(sy)}_*)^T \Sigma_{s|x}^*
    \end{aligned}
    \right) \beta^{(sm)}_* R
    = \left(
    \begin{aligned}
        \beta^{(xs)}_*  \\
        (\beta^{(sy)}_*)^T \Sigma_{s|x}^*
    \end{aligned}
    \right)
\end{align*}
Defining $H_M = \beta^{(sm)} R$, which must therefore be an invertible $K \times K$ matrix to preserve the rank $K$ of the transformed expression of $C$, it follows parametrizations preserving observed moments when the bottleneck condition is satisfied are:
\begin{align*}
    \beta^{(xs)}_* &= \beta^{(xs)} H_M \\
    \beta^{(sm)}_* &= H_M^{-1} \beta^{(sm)}  \\
    (\beta^{(sy)}_*)^T \Sigma_{s|x}^* &= (\beta^{(sy)})^T \Sigma_{s|x} H_M
\end{align*}
The last equation preserved the covariance between $M$ and $Y$ given $X$, and defines a condition up to an invertible linear transformation $K \times K$ matrix noted $H_Y$:
\begin{align*}
    \Sigma_{s|x}^* \beta^{(sy)}_* &=H_M^T \Sigma_{s|x} \beta^{(sy)}
\end{align*}
Since $\Sigma_{s|x}^*$ must remain positive-definite, we can invert it, which yields the transformation of $\beta^{(sy)}$ implied by the conservation of observed moments parametrization:
\begin{align*}
    \beta^{(sy)}_* &= \underbrace{(\Sigma_{s|x}^*)^{-1} H_M^T \Sigma_{s|x}}_{:=H_Y^{-1}} \beta^{(sy)}
\end{align*}
Noting that covariances are positive-definite and $H_M$ is invertible, $H_Y$ is also guaranteed to be invertible.
Finally, from the definition of $H_Y$:
\begin{align*}
    \Sigma_{s|x}^* &= H_M^T \Sigma_{s|x} H_Y
\end{align*}
Remaining expressions for the transformed parameters follows as in the main text.

\section{Supporting derivations for identifiability}
\label{sec-SM:add-deriv-id}
\subsection{Transformed covariances remain positive semi-definite}
We recall:
\begin{align*}
    H_Y H_M^{-1} &= \Sigma_{s|x}^{-1} \left( (\beta^{(sm)}\Sigma_{m}^{-1} (\beta^{(sm)})^T)^{-1} - (\beta^{(xs)})^T \Sigma_{x} \beta^{(xs)} \right)  \\
    \Sigma_{s|x}^* &= H_M^{T} \Sigma_{s|x} H_Y  \\  \notag
    \Sigma_{m|s}^* &= \Sigma_{m|s} + (\beta^{(sm)})^T \Sigma_{s|x} \beta^{(sm)} - (\beta^{(sm)}_{*})^T \Sigma_{s|x}^* \beta^{(sm)}_{*}  \\  \notag
    \Sigma_{y|x,s}^* &= \Sigma_{y|x,s} + (\beta^{(sy)})^T \Sigma_{s|x} \beta^{(sy)} - (\beta^{(sy)}_{*})^T \Sigma_{s|x}^* \beta^{(sy)}_{*}  \\  \notag
\end{align*}
The equations above guarantee positive semi-definite conditional covariances after transformation. 
Note that in general, $A^T G A \succeq 0$ when $G \succeq 0$ for any transformation of $A$.

For convenience of notation, note that the noise covariance of $M$ given $X$ is $\Sigma_{m|x} = \Sigma_{m} - (\beta^{(xs)} \beta^{(sm)})^T \Sigma_{x} \beta^{(xs)} \beta^{(sm)}$. Therefore:
\begin{align*}
    H_Y H_M^{-1} &= \Sigma_{s|x}^{-1} \left( (\beta^{(sm)}\Sigma_{m}^{-1} (\beta^{(sm)})^T)^{-1} - (\beta^{(xs)})^T \Sigma_{x} \beta^{(xs)} \right)  \\
    &=\Sigma_{s|x}^{-1} (\beta^{(sm)}\Sigma_{m|x}^{-1} (\beta^{(sm)})^T)^{-1}  \\
    H_M H_Y^{-1} &= \beta^{(sm)}\Sigma_{m|x}^{-1} (\beta^{(sm)})^T\Sigma_{s|x}
\end{align*}

The result is immediate since $\Sigma_{s|x}^*$ since $\Sigma_{m|x} \succeq 0$:
\begin{align*}
    \Sigma_{s|x}^* &= H_M^{T} (\beta^{(sm)}\Sigma_{m|x}^{-1} (\beta^{(sm)})^T)^{-1} H_M \\
    &\succeq 0 \\
\end{align*}

Developing $\Sigma_{m|s}^*$, we can show it also is a form $A^T G A \succeq 0$:
\begin{align*}
    \Sigma_{m|s}^* &= \Sigma_{m|x} - (\beta^{(sm)})^T ((\beta^{(sm)}\Sigma_{m|x}^{-1} (\beta^{(sm)})^T)^{-1}) \beta^{(sm)} \succeq 0  \\
    &= (I - (\beta^{(sm)})^T ((\beta^{(sm)}\Sigma_{m|x}^{-1} (\beta^{(sm)})^T)^{-1}) \beta^{(sm)}\Sigma_{m|x}^{-1}) \\
    &\times \Sigma_{m|x} \\
    &\times (I - (\beta^{(sm)})^T ((\beta^{(sm)}\Sigma_{m|x}^{-1} (\beta^{(sm)})^T)^{-1}) \beta^{(sm)}\Sigma_{m|x}^{-1})^T  \\
    &\succeq 0  \\
\end{align*}

Finally, let $\xi := (\Sigma_{s|x}^{-1} - \beta^{(sm)}\Sigma_{m|x}^{-1} (\beta^{(sm)})^T)$ and note that:
\begin{align*}
    \xi &= \xi \Sigma_{s|x}\xi^T + \beta^{(sm)}\Sigma_{m|x}^{-1} \Sigma_{m|s} \Sigma_{m|x}^{-1} (\beta^{(sm)})^T \\
    \Leftrightarrow \xi &\succeq 0
\end{align*}
Then positive semi-definiteness of $\Sigma_{y|x,s}^*$ immediately follows:
\begin{align*}
    \Sigma_{y|x,s}^* &= \Sigma_{y|x,s} + (\beta^{(sy)})^T \Sigma_{s|x}(\Sigma_{s|x}^{-1} - \beta^{(sm)}\Sigma_{m|x}^{-1} (\beta^{(sm)})^T ) \Sigma_{s|x} \beta^{(sy)}  \\
    &= \Sigma_{y|x,s} + (\beta^{(sy)})^T \Sigma_{s|x} \xi  \Sigma_{s|x} \beta^{(sy)}  \\
    &\succeq 0
\end{align*}

\subsection{Invariance of parametric covariance}
\begin{align*}
    \Sigma_{m}^* &= \Sigma_{m|s}^* + (\beta^{(sm)}_*)^T \Sigma_{s}^* \beta^{(sm)}_* \\
    &= \Sigma_{m|s} + (\beta^{(sm)})^T \Sigma_{s|x} \beta^{(sm)} - (\beta^{(sm)}_{*})^T \Sigma_{s|x}^* \beta^{(sm)}_{*} + (\beta^{(sm)}_*)^T (\Sigma_{s|x}^* + (\beta^{(xs)}_*)^T \Sigma_x^* \beta^{(xs)}_*) \beta^{(sm)}_* \\
    &= \Sigma_{m|s} + (\beta^{(sm)})^T \Sigma_{s|x} \beta^{(sm)} + (\beta^{(xs)}_* \beta^{(sm)}_*)^T \Sigma_x^* \beta^{(xs)}_* \beta^{(sm)}_* \\
    &= \Sigma_{m|s} + (\beta^{(sm)})^T \Sigma_{s|x} \beta^{(sm)} + (\beta^{(xs)} H_M H_M^{-1} \beta^{(sm)})^T \Sigma_x^* \beta^{(xs)} H_M H_M^{-1} \beta^{(sm)} \\
    &= \Sigma_{m|s} + (\beta^{(sm)})^T (\Sigma_{s|x} + (\beta^{(xs)})^T \Sigma_x^* \beta^{(xs)}) \beta^{(sm)} \\
    &= \Sigma_{m|s} + (\beta^{(sm)})^T (\Sigma_{s|x} + (\beta^{(xs)})^T \Sigma_x \beta^{(xs)}) \beta^{(sm)} \\
    &= \Sigma_{m|s} + (\beta^{(sm)})^T \Sigma_{s} \beta^{(sm)} \\
    &= \Sigma_{m}
\end{align*}

\subsection{Relation under exact recoverability}
\begin{align*}
    \Sigma_{m|s} \Sigma_{m}^{-1} (\beta^{(sm)})^T \Sigma_{s} &= 0  \\
    \Leftrightarrow (\Sigma_{m} - (\beta^{(sm)})^T \Sigma_{s} \beta^{(sm)}) \Sigma_{m}^{-1} (\beta^{(sm)})^T \Sigma_{s} &= 0  \\
    \Leftrightarrow (\beta^{(sm)})^T \Sigma_{s} - (\beta^{(sm)})^T \Sigma_{s} \beta^{(sm)}\Sigma_{m}^{-1} (\beta^{(sm)})^T \Sigma_{s} &= 0  \\
    \Leftrightarrow (\beta^{(sm)})^T \Sigma_{s} (I - \beta^{(sm)}\Sigma_{m}^{-1} (\beta^{(sm)})^T \Sigma_{s}) &= 0  \\
    \Leftrightarrow \beta^{(sm)}\Sigma_{m}^{-1} (\beta^{(sm)})^T \Sigma_{s} &= I
\end{align*}

\section{Inverting the model's parameters}
\label{sec-SM:Inverting-parameters}

Parameter estimates for any factorization of the joint linear Gaussian distribution $p(X,S,M,Y)$ can be recovered from the corresponding covariance representation. 
For instance, if the optimal parameter set~$\theta_{vM}$ is obtained from the LaMe-vM factorization, we can easily derive the equivalent parameter set~$\theta_{vX} = f(\theta_{vM})$ used in the LaMe-vX factorization. For illustration, consider the expectation of $X$ under the LaMe-vM factorization $\mathbb{E}[X|M] = M \beta^{(ms)} \beta^{(sx)}:= M\beta^{(mx)}$, which has no causal interpretation. Instead, under the LaMe-vX factorization, $\mathbb{E}[M|X] = X\beta^{(xm)}$ depends on the causal path coefficient $\beta^{(xm)}$, and can be recovered as follows \cite{Bishop2006PatternReco}:
\begin{align}
    \label{eq:parameters-reversal}
    \beta^{(xm)} &= \Sigma_{xx}^{-1} \Sigma_{xm} \\  \notag
    &= \Sigma_{xx}^{-1} \Sigma_{mx}^T \\  \notag
    &= \Big( \underbrace{(\beta^{(sx)})^T (\beta^{(ms)})^T \Sigma_{m} \beta^{(ms)} \beta^{(sx)}}_{\text{Noise propagating from $M$ to $X$}} + \underbrace{(\beta^{(sx)})^T \Sigma_{s|m} \beta^{(sx)}}_{\text{Noise from $S$ to $X$}}+ \underbrace{\Sigma_{x|s}}_{\text{Intrinsic noise}} \Big)^{-1}  \\  \notag
    &\times \Big(   \underbrace{\Sigma_{m} \beta^{(ms)}\beta^{(sx)}}_{\text{Signal from $M$ to $X$}} \Big)^T
\end{align}


For any generic unconditioned linear coefficient $\beta^{(ab)}$, the following relation holds:
\begin{align}
    \beta^{(ab|a)} = \Sigma_{aa}^{-1} \Sigma_{ab}
\end{align}

When conditioning on additional variables, the coefficient reads:
\begin{align}
    \beta^{(ab|a,c)} = (\Sigma_{aa}-\Sigma_{ac}\Sigma_{cc}^{-1}\Sigma_{ca})^{-1} (\Sigma_{ab}-\Sigma_{ac}\Sigma_{cc}^{-1}\Sigma_{cb})
\end{align}

Finally, we recall below all covariances and cross-covariances needed to compute the regression parameters in various models and factorizations.
Note that in the main text, we defined the shorthand notation $\beta^{(my)} := \beta^{(my|m,x)} = (\Sigma_{mm}-\Sigma_{mx}\Sigma_{xx}^{-1}\Sigma_{xm})^{-1} (\Sigma_{my}-\Sigma_{mx}\Sigma_{xx}^{-1}\Sigma_{xy})$ and $\beta^{(xm)}:=\beta^{(xm|x)} = \Sigma_{xx}^{-1} \Sigma_{xm}$.

\subsection{Non-latent model}
Standard mediation models can be expressed as:
\begin{align*}
    p(x,m,y) = p(y|x,m)p(m|x)p(x)
\end{align*}

We have:
\begin{align*}
    \Sigma_{xx} &= \Sigma_x \\
    \Sigma_{mm} &= (\beta^{(xm)})^T \Sigma_{xx} \beta^{(xm)} + \Sigma_m \\
    \Sigma_{mx} &= (\beta^{(xm)})^T \Sigma_{xx} \\
    \Sigma_{xm} &= \Sigma_{xx} \beta^{(xm)} \\
    \Sigma_{my} &= (\beta^{(xm)})^T \Sigma_x \beta^{(xy)} + \Sigma_{mm}\beta^{(my)}\\
    \Sigma_{xy} &= \Sigma_{xx} (\beta^{(xy)} + \beta^{(xm)}\beta^{(my)}) \\
\end{align*}

\subsection{Latent model}
Latent mediation models can be expressed as:
\begin{align*}
    p(x,m,y,s) = p(y|x,s)p(s|x)p(m|s)p(x) = p(y|x,s)p(s|m)p(x|s)p(m)
\end{align*}

The equations for inverting the parameters remain the same, but covariance matrices are now expressed as:
\paragraph{Model vX}
\begin{align*}
    \Sigma_{xx} &= \Sigma_x \\
    \Sigma_{ss} &= \Sigma_{s|x} + (\beta^{(xs)})^T \Sigma_{xx} \beta^{(xs)}\\
    \Sigma_{mm} &= \Sigma_{m|s} + (\beta^{(sm)})^T \Sigma_{ss} \beta^{(sm)} \\
    \Sigma_{xm} &= \Sigma_{xx} \beta^{(xs)} \beta^{(sm)} \\
    \Sigma_{mx} &= (\beta^{(sm)})^T (\beta^{(xs)})^T \Sigma_{xx} \\
    \Sigma_{my} &= (\beta^{(sm)})^T (\beta^{(xs)})^T \Sigma_{xx} \beta^{(xy)} + (\beta^{(sm)})^T \Sigma_{ss} \beta^{(sy)}\\
    \Sigma_{xy} &= \Sigma_{xx} (\beta^{(xy)} + \beta^{(xs)} \beta^{(sy)}) \\
\end{align*}

\paragraph{Model vM}
\begin{align*}
    \Sigma_{mm} &= \Sigma_m \\
    \Sigma_{ss} &= \Sigma_{s|m} + (\beta^{(ms)})^T \Sigma_{mm} \beta^{(ms)}\\
    \Sigma_{xx} &= \Sigma_{x|s} + (\beta^{(sx)})^T \Sigma_{ss} \beta^{(sx)} \\
    \Sigma_{mx} &= \Sigma_{mm} \beta^{(ms)} \beta^{(sx)}\\
    \Sigma_{xm} &= (\beta^{(sx)})^T (\beta^{(ms)})^T \Sigma_{mm} \\
    \Sigma_{my} &= \Sigma_{mm} \beta^{(ms)} (\beta^{(sx)}\beta^{(xy)} + \beta^{(sy)})\\
    \Sigma_{xy} &= \Sigma_{xx} \beta^{(xy)} + (\beta^{(sx)})^T \Sigma_{ss} \beta^{(sy)} \\
\end{align*}

\section{Signal-to-noise ratios}
\label{sec-SM:Force-SNR}
\subsection{LaMe}
\subsubsection{LaMe-vX}
By definition, signal-to-noise ratios are:
\begin{align*}
    SNR(S|X) &= \frac{Tr((\beta^{(xs)})^T \Sigma_x \beta^{(xs)})}{Tr(\Sigma_{s|x})} = \frac{||\Sigma_x^{\frac{1}{2}} \beta^{(xs)} ||_F^2}{||\Sigma_{s|x}^{\frac{1}{2}}||_F^2}\\
    SNR(M|S) &= \frac{Tr((\beta^{(sm)})^T ((\beta^{(xs)})^T \Sigma_{x} \beta^{(xs)} + \Sigma_{s|x}) \beta^{(sm)})}{Tr(\Sigma_{m|s})} = \frac{||((\beta^{(xs)})^T \Sigma_{x} \beta^{(xs)} + \Sigma_{s|x})^{\frac{1}{2}} \beta^{(sm)} ||_F^2}{||\Sigma_{m|s}^{\frac{1}{2}}||_F^2}\\
    SNR(Y|X,S) &= \frac{Tr((\beta^{(xy)} + \beta^{(xs)}\beta^{(sy)})^T\Sigma_x (\beta^{(xy)} + \beta^{(xs)}\beta^{(sy)}) + (\beta^{(sy)})^T \Sigma_{s|x} \beta^{(sy)})}{Tr(\Sigma_{y|x,s})} \\
    &= \frac{||\Sigma_x^{\frac{1}{2}} (\beta^{(xy)} + \beta^{(xs)}\beta^{(sy)})||_F^2 + || \Sigma_{s|x}^{\frac{1}{2}}\beta^{(sy)}||_F^2 }{||\Sigma_{y|x,s}^{\frac{1}{2}}||_F^2}
\end{align*}

The scaling factor can be obtained by simple algebra so that $\Sigma_A' = s_A \Sigma_A$ so that the signal-to-noise ratio of $A$ given its parent nodes is $R_A$, for $A \in [X,S,M,Y]$. Thus introducing $s_m$, $s_y$ and $s_s$:
\begin{align*}
    SNR(S|X) := R_s &= \frac{||\Sigma_x^{\frac{1}{2}} \beta^{(xs)} ||_F^2}{s_s||\Sigma_{s|x}^{\frac{1}{2}}||_F^2} \\
    \Leftrightarrow s_s &= \frac{||\Sigma_x^{\frac{1}{2}} \beta^{(xs)} ||_F^2}{R_s||\Sigma_{s|x}^{\frac{1}{2}}||_F^2}\\
    SNR(M|S) := R_m &= \frac{||((\beta^{(xs)})^T \Sigma_{x} \beta^{(xs)} + s_s\Sigma_{s|x})^{\frac{1}{2}} \beta^{(sm)} ||_F^2}{s_m||\Sigma_{m|s}^{\frac{1}{2}}||_F^2} \\
    \Leftrightarrow s_m &= \frac{||((\beta^{(xs)})^T \Sigma_{x} \beta^{(xs)} + s_s\Sigma_{s|x})^{\frac{1}{2}} \beta^{(sm)} ||_F^2}{R_m||\Sigma_{m|s}^{\frac{1}{2}}||_F^2}\\
    SNR(Y|X,S) := R_y &= \frac{||\Sigma_x^{\frac{1}{2}} (\beta^{(xy)} + \beta^{(xs)}\beta^{(sy)})||_F^2 + s_s ||\Sigma_{s|x}^{\frac{1}{2}}\beta^{(sy)}||_F^2 }{s_y||\Sigma_{y|x,s}^{\frac{1}{2}}||_F^2} \\
    \Leftrightarrow s_y &= \frac{||\Sigma_x^{\frac{1}{2}} (\beta^{(xy)} + \beta^{(xs)}\beta^{(sy)})||_F^2 + s_s ||\Sigma_{s|x}^{\frac{1}{2}}\beta^{(sy)}||_F^2 }{R_y||\Sigma_{y|x,s}^{\frac{1}{2}}||_F^2}
\end{align*}

\subsubsection{LaMe-vM}
By definition, signal-to-noise ratios are:
\begin{align*}
    SNR(S|M) &= \frac{Tr((\beta^{(ms)})^T \Sigma_m \beta^{(ms)})}{Tr(\Sigma_{s|m})} = \frac{||\Sigma_m^{\frac{1}{2}} \beta^{(ms)} ||_F^2}{||\Sigma_{s|m}^{\frac{1}{2}}||_F^2}\\
    SNR(X|S) &= \frac{Tr((\beta^{(sx)})^T ((\beta^{(ms)})^T \Sigma_m \beta^{(ms)} + \Sigma_{s|m}) \beta^{(sx)})}{Tr(\Sigma_{x|s})} = \frac{||((\beta^{(ms)})^T \Sigma_m \beta^{(ms)} + \Sigma_{s|m})^{\frac{1}{2}} \beta^{(sx)} ||_F^2}{||\Sigma_{x|s}^{\frac{1}{2}}||_F^2}\\
    SNR(Y|X,S) &= \frac{Tr((\beta^{(sy)})^T \Sigma_s \beta^{(sy)} + (\beta^{(xy)})^T \Sigma_x \beta^{(xy)} + 2 (\beta^{(sx)}\beta^{(xy)})^T \Sigma_s \beta^{(sy)})}{Tr(\Sigma_{y|x,s})} \\
    &= \frac{||((\beta^{(ms)})^T \Sigma_m \beta^{(ms)} + \Sigma_{s|m})^{\frac{1}{2}} (\beta^{(sx)} \beta^{(xy)} + \beta^{(sy)})||_F^2 + || \Sigma_{x|s}^{\frac{1}{2}} \beta^{(xy)}||_F^2}{||\Sigma_{y|x,s}^{\frac{1}{2}}||_F^2}  \\
\end{align*}
where the unconditional variances are $\Sigma_s = ((\beta^{(ms)})^T \Sigma_m \beta^{(ms)} + \Sigma_{s|m})$ and $\Sigma_x = \Sigma_{x|s} + (\beta^{(sx)})^T \Sigma_s \beta^{(sx)}$ in the LaMe-vM parametrization.

The scaling factor can be obtained by simple algebra so that $\Sigma_A' = s_A \Sigma_A$ so that the signal-to-noise ratio of $A$ given its parent nodes is $R_A$, for $A \in [X,S,M,Y]$. Thus introducing $s_x$, $s_y$ and $s_s$:
\begin{align*}
    SNR(S|M) := R_s &= \frac{||\Sigma_m^{\frac{1}{2}} \beta^{(ms)} ||_F^2}{s_s||\Sigma_{s|m}^{\frac{1}{2}}||_F^2}  \Leftrightarrow s_s = \frac{||\Sigma_m^{\frac{1}{2}} \beta^{(ms)} ||_F^2}{R_s||\Sigma_{s|m}^{\frac{1}{2}}||_F^2}  \\
    SNR(X|S) := R_x &= \frac{||((\beta^{(ms)})^T \Sigma_m \beta^{(ms)} + s_s \Sigma_{s|m})^{\frac{1}{2}} \beta^{(sx)} ||_F^2}{s_x||\Sigma_{x|s}^{\frac{1}{2}}||_F^2} \\
    \Leftrightarrow s_x &= \frac{||((\beta^{(ms)})^T \Sigma_m \beta^{(ms)} + s_s \Sigma_{s|m})^{\frac{1}{2}} \beta^{(sx)} ||_F^2}{R_x||\Sigma_{x|s}^{\frac{1}{2}}||_F^2}\\
    SNR(Y|X,S) := R_y &= \frac{||\Sigma_{ss}^{\frac{1}{2}}(s_s) (\beta^{(sx)} \beta^{(xy)} + \beta^{(sy)})||_F^2 + s_x||\Sigma_{x|s}^{\frac{1}{2}}\beta^{(xy)}||_F^2}{s_y||\Sigma_{y|x,s}^{\frac{1}{2}}||_F^2} \\
    \Leftrightarrow s_y &= \frac{||\Sigma_{ss}^{\frac{1}{2}}(s_s) (\beta^{(sx)} \beta^{(xy)} + \beta^{(sy)})||_F^2 + s_x||\Sigma_{x|s}^{\frac{1}{2}}\beta^{(xy)}||_F^2}{R_y||\Sigma_{y|x,s}^{\frac{1}{2}}||_F^2}  \\
    &= \frac{||((\beta^{(ms)})^T \Sigma_m \beta^{(ms)} + s_s \Sigma_{s|m})^{\frac{1}{2}} (\beta^{(sx)} \beta^{(xy)} + \beta^{(sy)})||_F^2 + s_x||\Sigma_{x|s}^{\frac{1}{2}}\beta^{(xy)}||_F^2}{R_y||\Sigma_{y|x,s}^{\frac{1}{2}}||_F^2}
\end{align*}
where $\Sigma_{ss}(s_s) = (\beta^{(ms)})^T \Sigma_m \beta^{(ms)} + s_s\Sigma_{s|m}$ and $\Sigma_{xx}(s_x, s_s) = (\beta^{(sx)})^T(\beta^{(ms)})^T \Sigma_m \beta^{(ms)}\beta^{(sx)} + (\beta^{(sx)})^T s_s \Sigma_{s|m}\beta^{(sx)} + s_x \Sigma_{x|s}$.

\subsection{Non-latent models}
By definition, signal-to-noise ratios are:
\begin{align*}
    SNR(M|X) &= \frac{Tr((\beta^{(xm)})^T \Sigma_x \beta^{(xm)})}{Tr(\Sigma_{m|x})} = \frac{||\Sigma_x^{\frac{1}{2}} \beta^{(xm)} ||_F^2}{||\Sigma_{m|x}^{\frac{1}{2}}||_F^2}\\
    SNR(Y|X,M) &= \frac{Tr((\beta^{(xy)} + \beta^{(xm)}\beta^{(my)})^T\Sigma_x (\beta^{(xy)} + \beta^{(xm)}\beta^{(my)}) + (\beta^{(my)})^T \Sigma_{m|x} \beta^{(my)})}{Tr(\Sigma_{y|x,m})} \\
    &= \frac{||\Sigma_x^{\frac{1}{2}} (\beta^{(xy)} + \beta^{(xm)}\beta^{(my)})||_F^2 + || \Sigma_{m|x}^{\frac{1}{2}}\beta^{(my)}||_F^2 }{||\Sigma_{y|x,m}^{\frac{1}{2}}||_F^2}
\end{align*}

The scaling factor can be obtained by simple algebra so that $\Sigma_A' = s_A \Sigma_A$ so that the signal-to-noise ratio of $A$ given its parent nodes is $R_A$, for $A \in [X,M,Y]$. Thus introducing $s_m$ and $s_y$:
\begin{align*}
    SNR(M|X) := R_m &= \frac{||\Sigma_x^{\frac{1}{2}} \beta^{(xm)} ||_F^2}{s_m||\Sigma_{m|x}^{\frac{1}{2}}||_F^2} \\
    \Leftrightarrow s_m &= \frac{||\Sigma_x^{\frac{1}{2}} \beta^{(xm)} ||_F^2}{R_m||\Sigma_{m|x}^{\frac{1}{2}}||_F^2}\\
    SNR(Y|X,M) := R_y &= \frac{||\Sigma_x^{\frac{1}{2}} (\beta^{(xy)} + \beta^{(xm)}\beta^{(my)})||_F^2 + s_m ||\Sigma_{m|x}^{\frac{1}{2}}\beta^{(my)}||_F^2 }{s_y||\Sigma_{y|x,m}^{\frac{1}{2}}||_F^2} \\
    \Leftrightarrow s_y &= \frac{||\Sigma_x^{\frac{1}{2}} (\beta^{(xy)} + \beta^{(xm)}\beta^{(my)})||_F^2 + s_m ||\Sigma_{m|x}^{\frac{1}{2}}\beta^{(my)}||_F^2 }{R_y||\Sigma_{y|x,m}^{\frac{1}{2}}||_F^2}
\end{align*}

\section{Baselines implementation details}
\label{sec-SM:BaselinesDerkLind}
\paragraph{JAOG} We extended \cite{Lindquist2023MLHighDimMediation}'s model to handle multivariate latent variables ($K\geq1$). In the article, the authors posit a deterministic dimension reduction model $S = \varphi(M)$, whose output, $S$, is used in a standard LASSO mediation model. To do so, the authors define an optimization target both for inferring the parameters of $\varphi(M)$ and for inferring the structural parameters of the mediation model. Adapting their notation, we define $E = Y - X \beta^{(xy)}$ and $H = X \beta^{(xs)}$. The target of the multivariate extension of \cite{Lindquist2023MLHighDimMediation} then becomes $||\varphi(M)-(E (\beta^{(sy)})^T+H)(\beta^{(sy)}(\beta^{(sy)})^T+I_K)^{-1}||^2$. Finally, we choose $\varphi(M)$ to be a multivariate linear Gaussian regression model so that \cite{Lindquist2023MLHighDimMediation}'s approach is the closest to the data generation process used in our experiments. 

\paragraph{LVMA} 
We considered a direct comparison with \cite{Derkach2019HighDimMed}, whose model assumes a diagonal manifest residual covariance $\Sigma_{m|s}$, whereas LaMe and other baselines impose the perfect-measurement restriction $\Sigma_{m|s}\beta^{(ms)}=0$ for effect identification. In preliminary experiments, an adaptation accommodating multivariate exposures and outcomes performed poorly when the data-generating process violated the diagonal residual-covariance assumption. Such a comparison would therefore combine differences in latent dimensionality with differences in covariance restrictions, making it difficult to isolate the contribution of dimension reduction. 
Moreover, a direct benchmark of \cite{Derkach2019HighDimMed} would require extending its scalar exposure-outcome formulation while preserving its covariance restrictions, penalization, and model-selection procedure. The resulting comparison would depend on additional methodological choices beyond the scope of the present evaluation. We therefore avoid attributing the performance of a substantially modified implementation to the original method.
Instead, we use an overparametrized LaMe comparator with $K=\frac{I+J+L}{2}$, satisfying $K \geq \min(I+L,J)$. This comparator retains the same covariance flexibility while removing the rank restriction responsible for compression, which closely relates to the point we discussed about \cite{Derkach2019HighDimMed} in the background section.

All remaining baselines were implemented using scikit-learn.

\section{Scalability}
\label{sec-SM:scalability}
\begin{figure}[H]
    \centering
    \includegraphics[width=\textwidth]{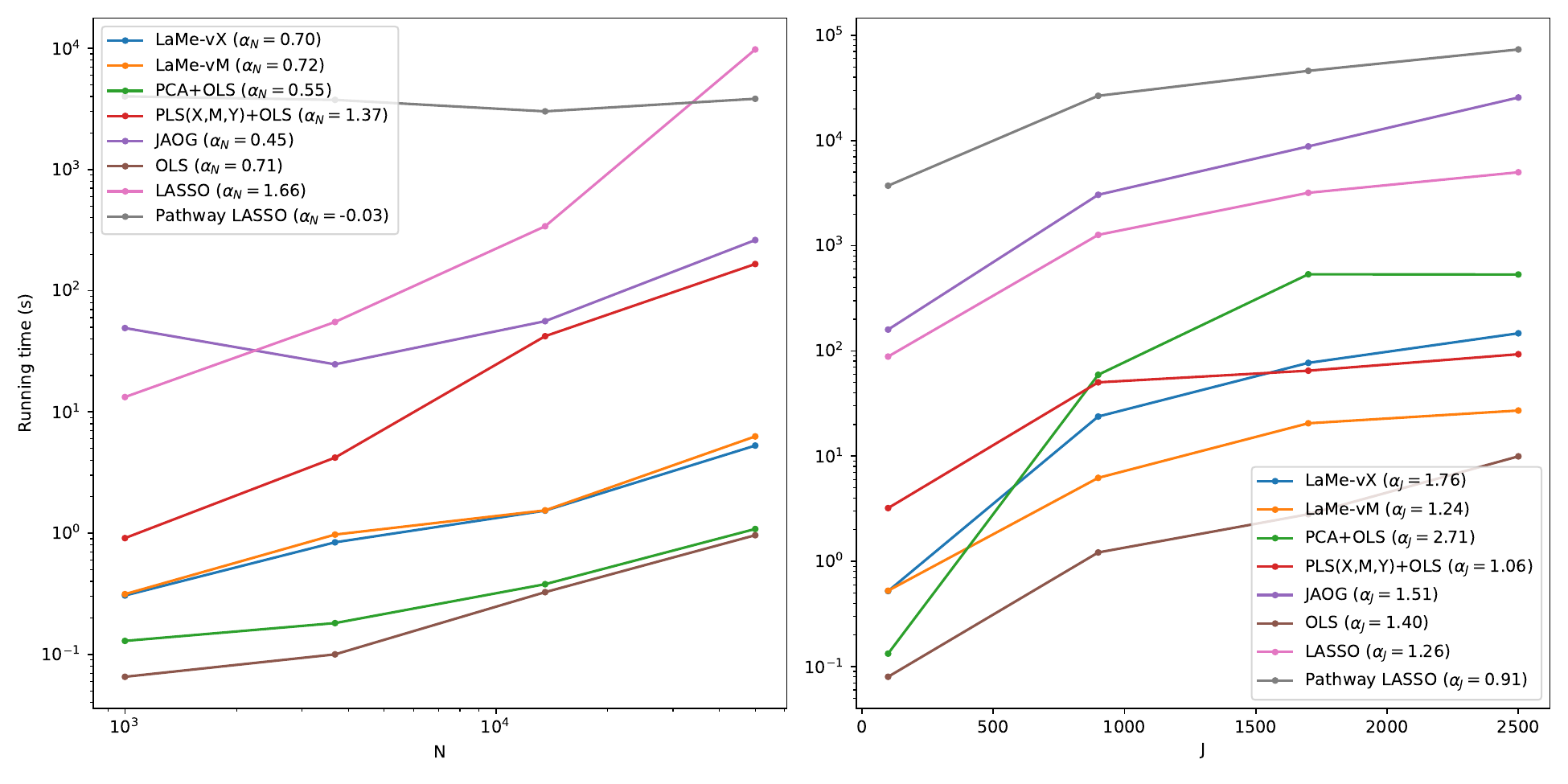}
    \caption{\textbf{Scalability of the models} --- For a system with $I=100$, $J=100$, $L=1$, $N=3000$, with one full inference cycle per model. Each curve is normalized by its smallest entry to show how running time evolves as a function of $N$ (left) and $J$ (right). $\alpha$ denotes the power-law exponent.
    }
    \label{fig-XP1}
\end{figure}
In Fig.~\ref{fig-XP1}, we plot the normalized running time of LaMe and the baseline methods for a system with $I=100$, $J=100$, $L=1$, $N=3000$, unless specified otherwise. The models are run for a full inference procedure, meaning LaMe's EM algorithm is run 100 times, LASSO, PLS and PCA select their hyperparameters by cross-validation, etc. Refer to the main text for more details.

We see that LaMe scales similarly to other methods with respect to $N$, and better than most with respect to $J$. Pathway LASSO seems to scale well, but has the longest absolute convergence time. This figure also highlights the benefits of choosing the LaMe-vM factorization over the LaMe-vX factorization when $J$ is large. Absolute running times must be interpreted with caution due to their dependence on experimental parameters (number of runs, number of cross-validation folds, convergence criterion, etc.).
Overall, this further supports the versatility of the LaMe framework, which can handle a range of situations efficiently and accurately.

\section{Correlation and identifiability issues in retrieving the observed measurement set}
\label{sec-SM:issues-mediating-biomarkers}
In \cite{Derkach2019HighDimMed}, the authors define the mediating set between exposure $X_i$ and outcome $Y_l$ as: $\mathcal{M}^{il}_{LVMA} := \{ M_j | C_{ijl}^{LVMA} := \sum_k |\beta^{(xs)}_{ik} \beta^{(sm)}_{kj} \beta^{(sy)}_{kl}| \neq 0 \}$, where $C_{ijl}^{LVMA}$ is a shorthand notation. This definition, however, does not account for candidate mediators' correlation and non-identifiability, which raises interpretation issues. 

Correlation matters for interpretation, because $\beta^{(sm)}_{kj} \neq 0$ in $\mathcal{M}^{il}_{LVMA}$ does not necessarily imply that $M_j$ informs latent variables $S$ acting as mediators between $X_i$ and $Y_l$. For instance, $\beta^{(sm)}_{kj} \neq 0$ could reflect variation in $M_j$ along a latent component associated with $X_i$, without implying that this component contributes to $Y_l$. 
For $M_j$ to be interpreted as a mediator, a proper definition should therefore account for both a non-zero $\beta^{(xs)}\beta^{(sm)}$ ($M$ varies with $X$ through $S$) and a non-zero $\beta^{(ms)}\beta^{(sy)}$ ($M$ informs $Y$ through $S$); the expression of $\beta^{(ms)}$ in particular involves $\Sigma_{m|s}$, thus accounting for correlation between manifest variables. 

The non-identifiability issue arises in two cases when rotational and scaling invariance exists. Scaling invariance can be removed by forcing $\Sigma_{s|x}=I$, as in \cite{Derkach2019HighDimMed}; rotational invariance however, remains.
Rotational invariance states that $(\beta^{(xs)}, \beta^{(sm)}, \beta^{(sy)}, \theta(I))$ and $(\beta^{(xs)}R, R^T\beta^{(sm)}, R^T\beta^{(sy)}, \theta(R))$ define an equivalent reparametrization of the latent mediation model, where $\theta(R)$ denotes the set of other parameters related to $S$ after an orthogonal rotation $R$. 
The issue arises because $\mathcal{M}^{il}_{LVMA}$ is not invariant to these likelihood-equivalent parametrizations: factorizing $C_{ijl}^{LVMA} = |\beta^{(xs)}_{i\cdot}diag(\beta^{(sm)}_{\cdot j})\beta^{(sy)}_{\cdot l}| \neq |\beta^{(xs)}_{i\cdot}R \cdot diag(R^T\beta^{(sm)}_{\cdot j})\cdot R^T\beta^{(sy)}_{\cdot l}|$ shows that $R$ does not cancel out.
As such, mediating biomarkers can be arbitrarily added or removed from the set $\mathcal{M}^{il}_{LVMA}$ through likelihood-equivalent rotations of the latent space, hence the need for a rotationally invariant definition of the ``mediating'' manifest set of variables.

\section{Additional synthetic experiments}
\label{sec-SM:Additional-synth-XP}
\begin{figure}[H]
    \centering
    \includegraphics[width=0.9\textwidth]{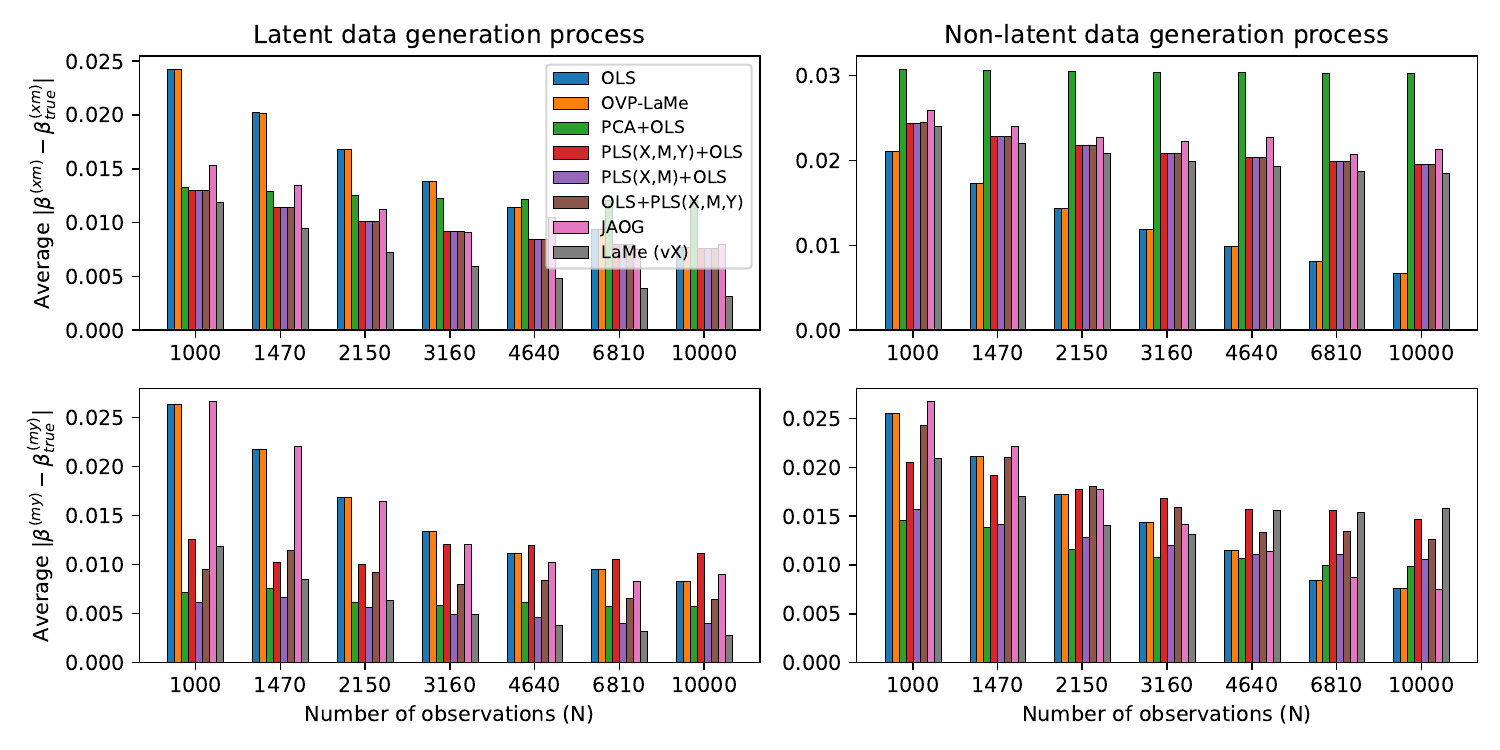}
    \caption{\textbf{LaMe accurately retrieves underlying parameters} --- LaMe and baselines when $I=50$, $J=200$, $L=1$, and a signal-to-noise ratio from $S$ to $Y$ and $M$ equal to $0.1$. \textbf{(Top)} Mean absolute error over all parameters in $\beta^{(xm)}_{inf}$ and $\beta^{(xm)}_{true}$ ; \textbf{(Bottom)} Mean absolute error over all parameters $\beta^{(my)}_{inf}$ and $\beta^{(my)}_{true}$ ; \textbf{(left)} data are generated using a latent process ($K=10$) with the true $K$ provided to all latent models ; \textbf{(right)} data are generated using a non-latent process with $K=\frac{\min(I+L, J)}{2}$ for all latent models except for OVP-LaMe where $K=\frac{I+J}{2}=100$.
    }
    \label{fig-XP2-HighSNR}
\end{figure}

We explore in Fig.~\ref{fig-XP2-HighSNR} how models perform when there is an imbalance between the dimensionality of $X$ and $M$. Our results show that models perform similarly as in the balanced case, although with a notable increase in accuracy of two-step latent approaches (PCA, PLS) for latent data. Increasing $J$ means the mediators carry more information about the whole system, thus alleviating PCA's sequential estimation. On the other hand, lowering $I$ reduces the amount of information flowing through the direct path $\beta^{(xy)}$, which also alleviates PLS's sequential estimation bias.

\begin{figure}[H]
    \centering
    \includegraphics[width=\textwidth]{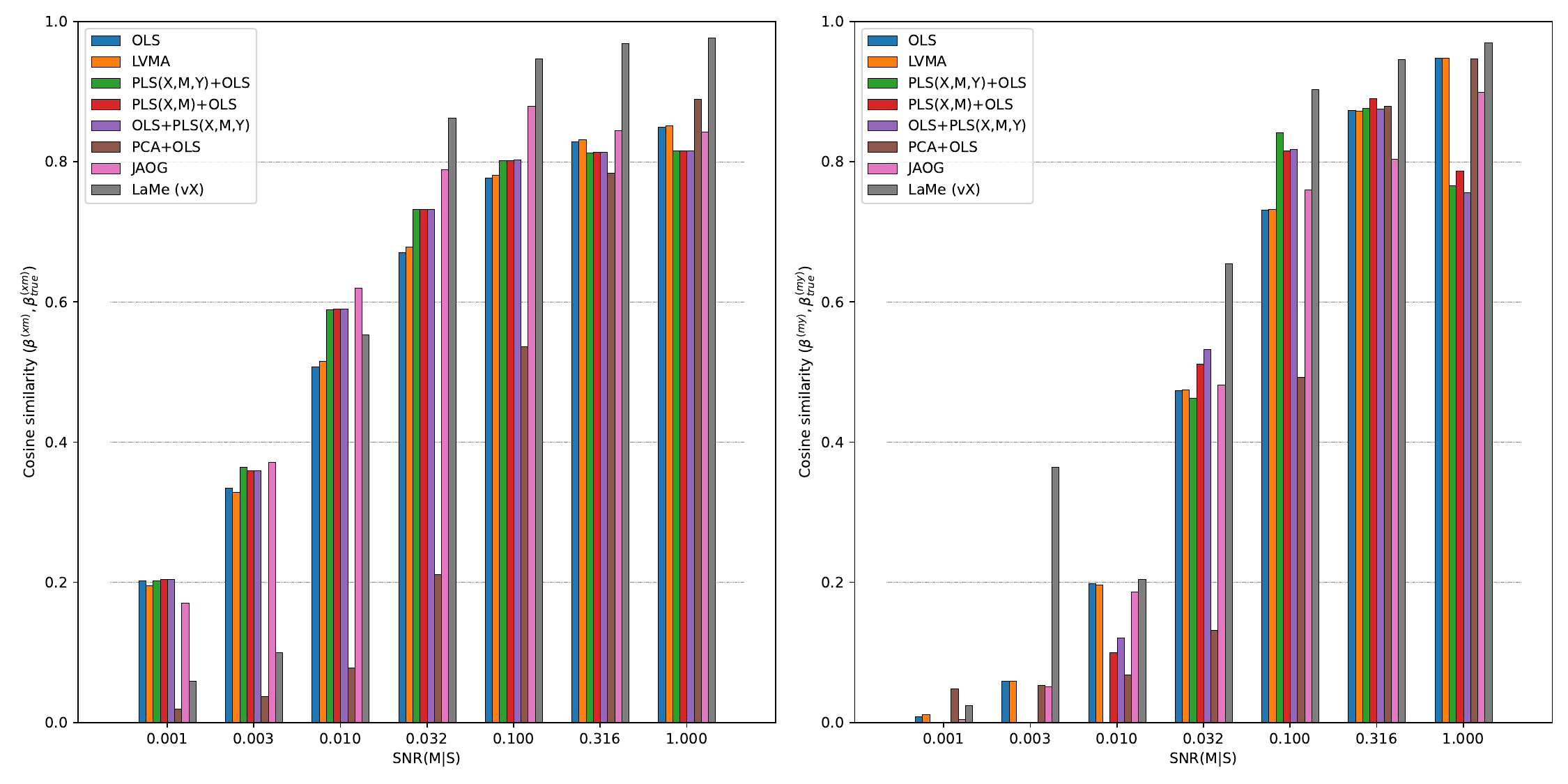}
    \caption{\textbf{LaMe accurately retrieves underlying parameters over a range of signal-to-noise ratios} --- LaMe and baselines when $I=100$, $J=100$, $K=10$, $L=1$, and a varying signal-to-noise ratio from $S$ to $Y$ and $M$. \textbf{(left)} Cosine similarity between $\beta^{(xm)}_{inf}$ and $\beta^{(xm)}_{true}$ ; \textbf{(right)} Cosine similarity between $\beta^{(my)}_{inf}$ and $\beta^{(my)}_{true}$ ; data are generated using a latent process ($K=10$).
    }
    \label{fig-XP4}
\end{figure}

In Fig.~\ref{fig-XP4}, we compare models against extreme signal-to-noise ratios from $S$ to $M$ and $Y$ on latent data. We use cosine similarity as a normalized measure for absolute comparison. Overall, we observe a sharp decrease in accuracy for $\beta^{(my)}$ once $SNR(Y|S)<3\%$ whereas $\beta^{(xm)}$ estimates remain accurate for even smaller $SNR(M|S)$ --which can be explained by a greater amount of information available due to $J\gg L$. Overall, LaMe still outperforms baselines and retains its accuracy for lower values of $SNR(M,Y|S)$.

\begin{figure}[H]
    \centering
    \includegraphics[width=0.75\textwidth]{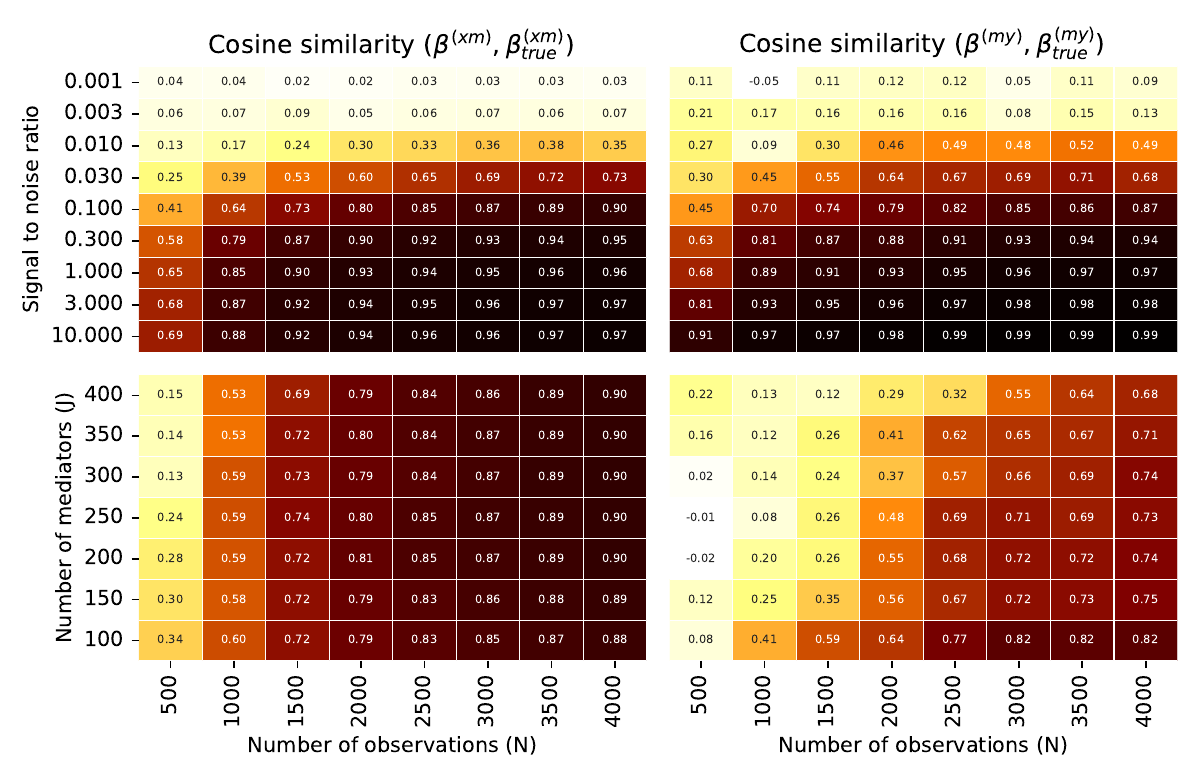}
    \caption{\textbf{LaMe's parameter retrieval in challenging situations} --- LaMe's cosine similarity when $I=100$, $J=100$, $K=10$, $L=1$ and $SNR(Y|S)=SNR(M|S)=0.1$, unless specified otherwise, with a varying number of observations $N$ | (top) with respect to $SNR(M|S)=SNR(Y|S)$ | (bottom) with respect to the number of mediators $J$ ; a dataset is generated once for $J=400$, and models are trained on subsets of $M$, which can cause $SNR(Y|S), SNR(M|S)$ to fluctuate around their target value across different $J$ | \textbf{(left)} between estimated $\beta^{(xm)}$ and $\beta^{(xm)}_{true}$ | \textbf{(right)} between estimated $\beta^{(my)}$ and $\beta^{(my)}_{true}$.
    }
    \label{fig-XP3-full}
\end{figure}

Fig.~\ref{fig-XP3-full} shows that LaMe's performance is unaffected by the dimension of $M$ once $SNR(M|S)$ is fixed. We use cosine similarity as a normalized measure for absolute comparison.

\section{Real-world data - Standard 5-fold CV}
\label{sec-SM:RW-XP-5xCV}
\begin{table}
\caption{Out-of-sample and in-sample performance metrics of LaMe (version $X$ and $M$, with and without $\ell 1$ regularization, assuming full noise covariances) and all baseline models. The last lines report the average improvement of models' metrics over OLS across all datasets --hence the null line for OLS. NLPD is the Negative Log-Predictive Density, $R^2$ the fraction of explained variance and $AIC$ the Akaike Information Criterion. When applicable, the number of latent variables is reported as $K$. Errors correspond to the standard error across five folds. The colour scale indicates relative performance within each dataset and metric ; (orange=best). }
\label{table-res-RW-red}
\centering
\noindent\makebox[\columnwidth]{\resizebox{\columnwidth}{!}{

\begin{tabular}{|l|l|S|S|S|S|S}

    \cline{1-6}

    & &\text{$NLPD(Y|X,M)$ $\downarrow$} & \text{$R^2(Y|X,M)$ $\uparrow$} & \text{AIC $\downarrow$} & \text{$K$ $\downarrow$}\\

    \cline{1-6}

    \multirow{10}{*}{\rotatebox[origin=c]{90}{\footnotesize \text{\textbf{NHANES}}}}
    & LaMe (vM) &  \cellcolor{orange!21} 1.326(51) &  \cellcolor{orange!4} 0.186(10) &  \cellcolor{orange!30} 98.318(320) &  \cellcolor{orange!17} 3.000(1)\\
    & LaMe (vM) (reg) &  \cellcolor{orange!21} 1.326(52) &  \cellcolor{orange!3} 0.186(10) &  \cellcolor{orange!29} 98.374(310) &  \cellcolor{orange!17} 3.000(1)\\
    & LaMe (vX) &  \cellcolor{orange!22} 1.326(51) &  \cellcolor{orange!4} 0.186(10) &  \cellcolor{orange!30} 98.318(320) &  \cellcolor{orange!17} 3.000(1)\\
    & LaMe (vX) (reg) &  \cellcolor{orange!21} 1.326(52) &  \cellcolor{orange!6} 0.187(10) &  \cellcolor{orange!30} 98.322(317) &  \cellcolor{orange!17} 3.000(1)\\
    & PCA+OLS &  \cellcolor{orange!21} 1.326(53) &  \cellcolor{orange!7} 0.187(12) &  \cellcolor{orange!17} 98.809(321) &  \cellcolor{orange!0} 24.000(1)\\
    & PCA+LASSO &  \cellcolor{orange!30} 1.320(51) &  \cellcolor{orange!30} 0.194(9) &  \cellcolor{orange!15} 98.893(322) &  \cellcolor{orange!0} 24.000(1)\\
    & PLS(X,M,Y)+OLS &  \cellcolor{orange!0} 1.484(23) &  \cellcolor{orange!0} -0.139(70) &  \cellcolor{orange!22} 98.610(368) &  \cellcolor{orange!0} 6.000(1)\\
    & PLS(X,M)+OLS &  \cellcolor{orange!22} 1.325(51) &  \cellcolor{orange!2} 0.186(9) &  \cellcolor{orange!30} 98.319(320) &  \cellcolor{orange!0} 6.000(1)\\
    & OVP-LaMe &  \cellcolor{orange!21} 1.326(52) &  \cellcolor{orange!6} 0.187(12) &  \cellcolor{orange!22} 98.612(321) &  \cellcolor{orange!0} 19.000(1)\\
    & JAOG &  \cellcolor{orange!23} 1.325(52) &  \cellcolor{orange!13} 0.189(11) &  \cellcolor{orange!7} 99.219(320) &  \cellcolor{orange!30} 1.800(179)\\

    \cline{1-6}

    \multirow{10}{*}{\rotatebox[origin=c]{90}{\footnotesize \text{\textbf{EPIC CAN}}}}
    & LaMe (vM) &  \cellcolor{orange!19} 0.701(7) &  \cellcolor{orange!22} 0.762(1) &  \cellcolor{orange!30} 861.197(607) &  \cellcolor{orange!21} 4.000(1)\\
    & LaMe (vM) (reg) &  \cellcolor{orange!19} 0.701(7) &  \cellcolor{orange!22} 0.762(1) &  \cellcolor{orange!30} 861.029(495) &  \cellcolor{orange!18} 5.000(1)\\
    & LaMe (vX) &  \cellcolor{orange!19} 0.701(7) &  \cellcolor{orange!22} 0.762(1) &  \cellcolor{orange!30} 861.197(607) &  \cellcolor{orange!21} 4.000(1)\\
    & LaMe (vX) (reg) &  \cellcolor{orange!19} 0.701(7) &  \cellcolor{orange!22} 0.762(1) &  \cellcolor{orange!13} 866.853(463) &  \cellcolor{orange!30} 1.000(1)\\
    & PCA+OLS &  \cellcolor{orange!0} 0.747(7) &  \cellcolor{orange!0} 0.746(2) &  \cellcolor{orange!0} 933.025(613) &  \cellcolor{orange!0} 377.600(219)\\
    & PCA+LASSO &  \cellcolor{orange!30} 0.692(8) &  \cellcolor{orange!30} 0.766(1) &  \cellcolor{orange!0} 935.368(635) &  \cellcolor{orange!0} 377.600(219)\\
    & PLS(X,M,Y)+OLS &  \cellcolor{orange!0} 0.776(18) &  \cellcolor{orange!0} 0.723(9) &  \cellcolor{orange!30} 861.128(611) &  \cellcolor{orange!0} 15.000(1)\\
    & PLS(X,M)+OLS &  \cellcolor{orange!0} 0.720(9) &  \cellcolor{orange!5} 0.752(4) &  \cellcolor{orange!30} 861.032(610) &  \cellcolor{orange!0} 15.000(1)\\
    & OVP-LaMe &  \cellcolor{orange!0} 0.747(6) &  \cellcolor{orange!0} 0.746(2) &  \cellcolor{orange!0} 897.439(602) &  \cellcolor{orange!0} 206.000(1)\\
    & JAOG &  \cellcolor{orange!0} 0.736(13) &  \cellcolor{orange!1} 0.750(3) &  \cellcolor{orange!23} 863.432(638) &  \cellcolor{orange!23} 3.400(358)\\

    \cline{1-6}

    \multirow{10}{*}{\rotatebox[origin=c]{90}{\footnotesize \text{\textbf{PISA - I}}}}
    & LaMe (vM) &  \cellcolor{orange!30} 1.066(4) &  \cellcolor{orange!30} 0.506(4) &  \cellcolor{orange!30} 102.216(25) &  \cellcolor{orange!3} 10.000(1)\\
    & LaMe (vM) (reg) &  \cellcolor{orange!30} 1.066(4) &  \cellcolor{orange!30} 0.506(4) &  \cellcolor{orange!30} 102.217(27) &  \cellcolor{orange!3} 10.000(1)\\
    & LaMe (vX) &  \cellcolor{orange!30} 1.066(4) &  \cellcolor{orange!29} 0.506(4) &  \cellcolor{orange!30} 102.216(25) &  \cellcolor{orange!3} 10.000(1)\\
    & LaMe (vX) (reg) &  \cellcolor{orange!30} 1.066(4) &  \cellcolor{orange!29} 0.506(4) &  \cellcolor{orange!29} 102.232(23) &  \cellcolor{orange!3} 10.000(1)\\
    & PCA+OLS &  \cellcolor{orange!13} 1.075(4) &  \cellcolor{orange!1} 0.497(4) &  \cellcolor{orange!30} 102.227(25) &  \cellcolor{orange!0} 19.000(1)\\
    & PCA+LASSO &  \cellcolor{orange!12} 1.075(4) &  \cellcolor{orange!0} 0.497(4) &  \cellcolor{orange!29} 102.231(25) &  \cellcolor{orange!0} 19.000(1)\\
    & PLS(X,M,Y)+OLS &  \cellcolor{orange!0} 1.197(5) &  \cellcolor{orange!0} 0.358(7) &  \cellcolor{orange!22} 102.496(25) &  \cellcolor{orange!0} 12.000(1)\\
    & PLS(X,M)+OLS &  \cellcolor{orange!0} 1.189(3) &  \cellcolor{orange!0} 0.368(5) &  \cellcolor{orange!22} 102.477(24) &  \cellcolor{orange!0} 12.000(1)\\
    & OVP-LaMe &  \cellcolor{orange!30} 1.066(4) &  \cellcolor{orange!30} 0.506(4) &  \cellcolor{orange!30} 102.213(25) &  \cellcolor{orange!0} 23.000(1)\\
    & JAOG &  \cellcolor{orange!0} 1.162(3) &  \cellcolor{orange!0} 0.402(3) &  \cellcolor{orange!0} 103.830(28) &  \cellcolor{orange!30} 1.000(1)\\

    \cline{1-6}

    \multirow{10}{*}{\rotatebox[origin=c]{90}{\footnotesize \text{\textbf{CRIME}}}}
    & LaMe (vM) &  \cellcolor{orange!15} 0.894(14) &  \cellcolor{orange!15} 0.651(14) &  \cellcolor{orange!30} 72.937(252) &  \cellcolor{orange!14} 15.000(1)\\
    & LaMe (vM) (reg) &  \cellcolor{orange!12} 0.896(16) &  \cellcolor{orange!11} 0.649(15) &  \cellcolor{orange!6} 75.055(330) &  \cellcolor{orange!22} 10.000(1)\\
    & LaMe (vX) &  \cellcolor{orange!14} 0.895(15) &  \cellcolor{orange!14} 0.650(15) &  \cellcolor{orange!30} 72.937(251) &  \cellcolor{orange!14} 15.000(1)\\
    & LaMe (vX) (reg) &  \cellcolor{orange!9} 0.898(17) &  \cellcolor{orange!9} 0.648(16) &  \cellcolor{orange!10} 74.700(246) &  \cellcolor{orange!22} 10.000(1)\\
    & PCA+OLS &  \cellcolor{orange!23} 0.889(15) &  \cellcolor{orange!25} 0.655(14) &  \cellcolor{orange!29} 73.022(246) &  \cellcolor{orange!0} 32.000(1)\\
    & PCA+LASSO &  \cellcolor{orange!30} 0.885(16) &  \cellcolor{orange!30} 0.657(16) &  \cellcolor{orange!23} 73.542(243) &  \cellcolor{orange!0} 32.000(1)\\
    & PLS(X,M,Y)+OLS &  \cellcolor{orange!0} 0.927(14) &  \cellcolor{orange!0} 0.626(13) &  \cellcolor{orange!0} 76.419(222) &  \cellcolor{orange!0} 32.000(1)\\
    & PLS(X,M)+OLS &  \cellcolor{orange!0} 0.925(14) &  \cellcolor{orange!0} 0.627(13) &  \cellcolor{orange!0} 76.401(224) &  \cellcolor{orange!0} 32.000(1)\\
    & OVP-LaMe &  \cellcolor{orange!23} 0.889(15) &  \cellcolor{orange!25} 0.655(14) &  \cellcolor{orange!5} 75.213(247) &  \cellcolor{orange!0} 50.000(1)\\
    & JAOG &  \cellcolor{orange!16} 0.893(14) &  \cellcolor{orange!18} 0.652(15) &  \cellcolor{orange!0} 80.429(255) &  \cellcolor{orange!30} 5.000(1)\\

    \cline{1-6}

\end{tabular}

}}
\end{table}

The results for a standard 5-fold cross-validation procedure are provided in Table~\ref{table-res-RW-red}. The data are split into 80\% training data and 20\% test data with no overlap between testing samples. We stress that PCA+LASSO's overall good predictive performances are consistently achieved at the cost of high-dimensional latent mediators, whereas models such as LaMe or JAOG seem to efficiently compress the relevant information into a smaller number of latent variables.

\section{Full mediation graphs}
\label{sec-SM:full-mediation-graphs}
\begin{figure}[H]
    \centering
    \includegraphics[height=0.85\textheight]{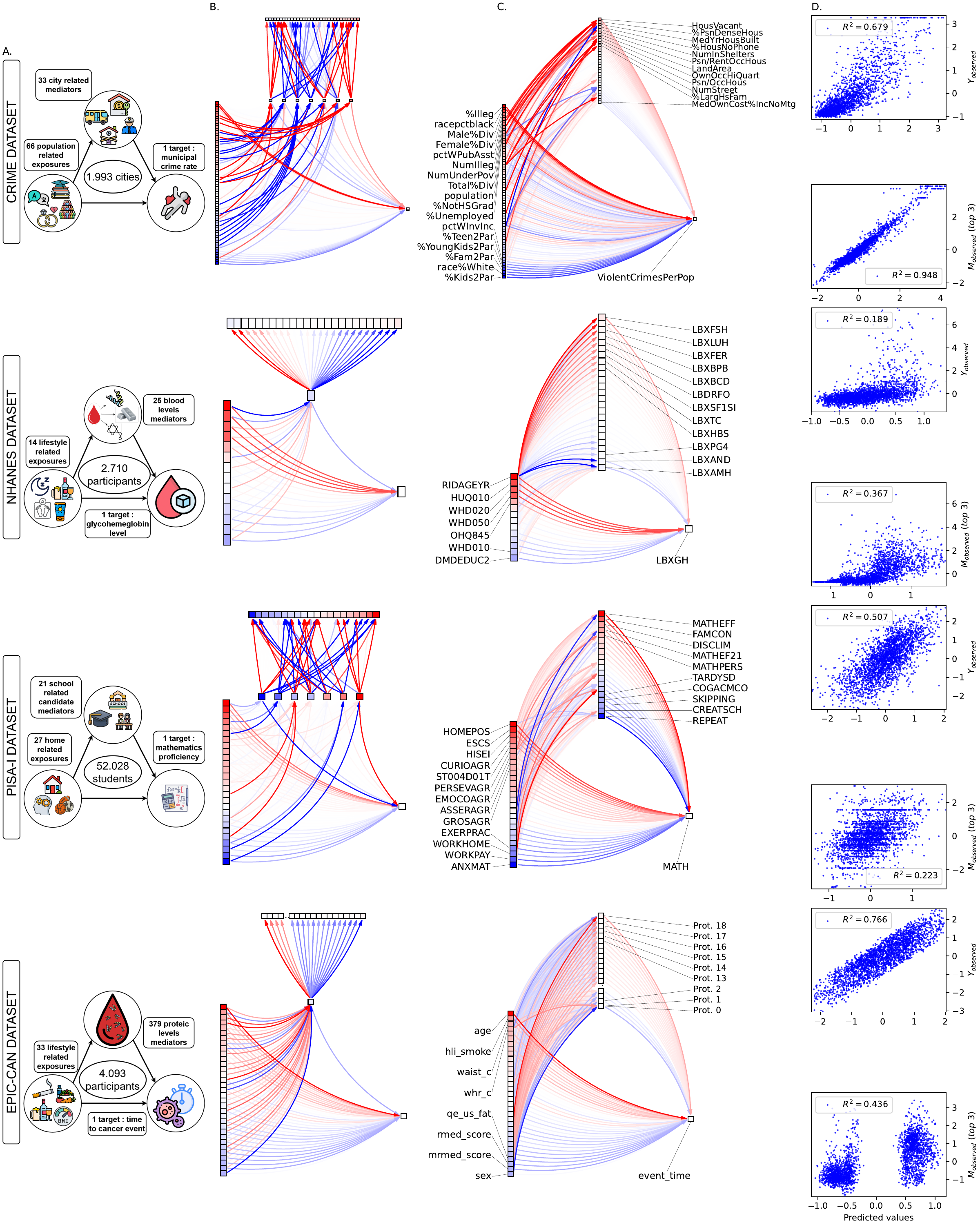}
    \caption{\textbf{LaMe's results on real-world data} --- \textbf{(A)} Schematic illustration of the dataset and of the mediated system at hand. \textbf{(B,C)} Results from the LaMe-vX model, assuming either the causal structural model with latent mediators with $K$ chosen for the lowest BIC (B) or without latent mediators (C). Arrows represent the standardized structural coefficients $\tilde \beta^{(uv)} = \Sigma_{u}^{\frac{1}{2}}\beta^{(uv)}\Sigma_{v}^{-\frac{1}{2}}$; for variables with a causal path to the outcome, squares' colour represents the total causal effect of a unit intervention on the corresponding variable on the outcome~$Y$ under the model specified with latent mediators--in this case, the sum of all causal paths leading to $Y$; in the latent graph, mediator colours represent marginal associations with $Y$, not causal effects. For both squares and arrows, blue represents a negative relation, red represents a positive relation, and colour intensity represents its strength with extremal values at $\pm 0.25$ s.d. Annotations of (C) match those in (B). \textbf{(D)} Plot of the predicted values of $Y$ and entries of $M$ (the three with highest $R^2$) against their true values, along with the associated coefficient of determination $R^2$.
    }
    \label{fig-full_fit_res}
\end{figure}
\begin{figure}[H]
    \centering
    \includegraphics[width=0.45\textwidth]{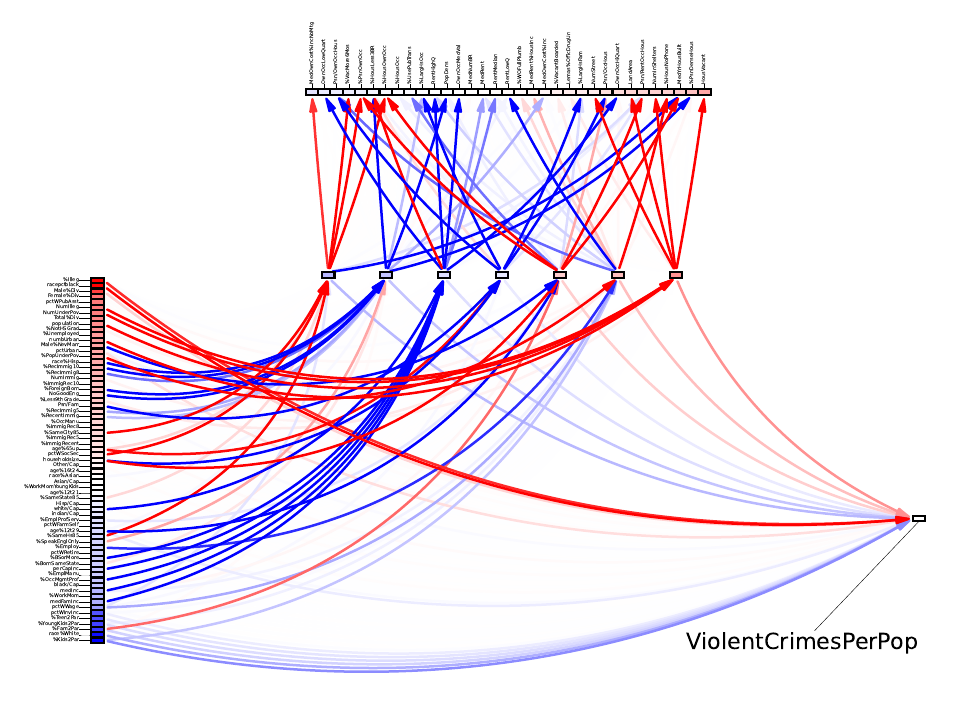}
    \includegraphics[width=0.45\textwidth]{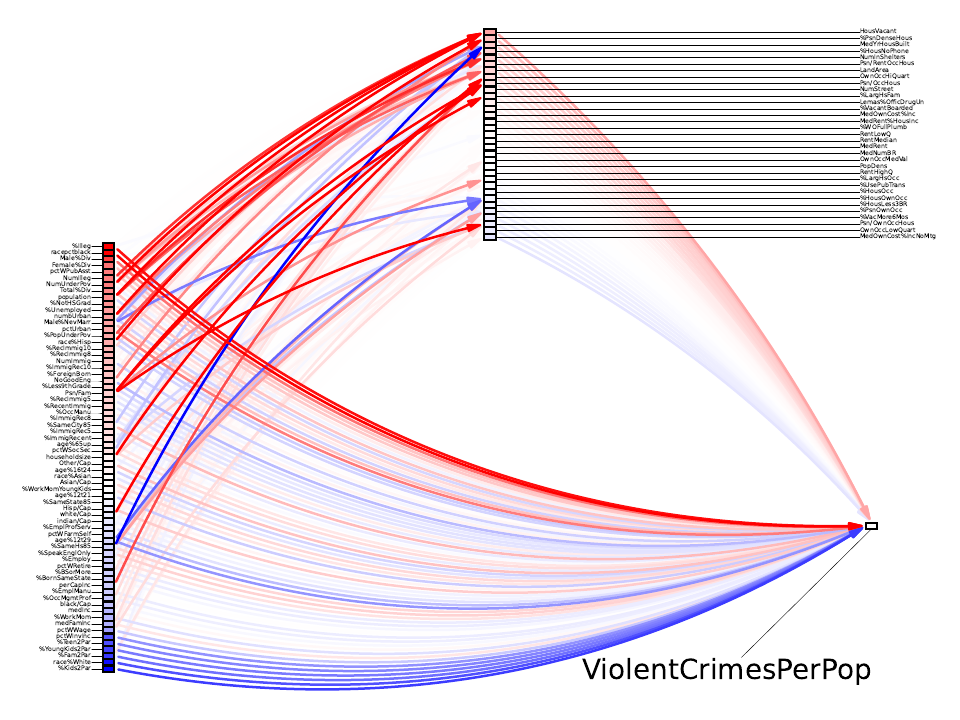}
    \caption{Full graphs for the CRIME dataset}
\end{figure}
\begin{figure}[H]
    \centering
    \includegraphics[width=0.45\textwidth]{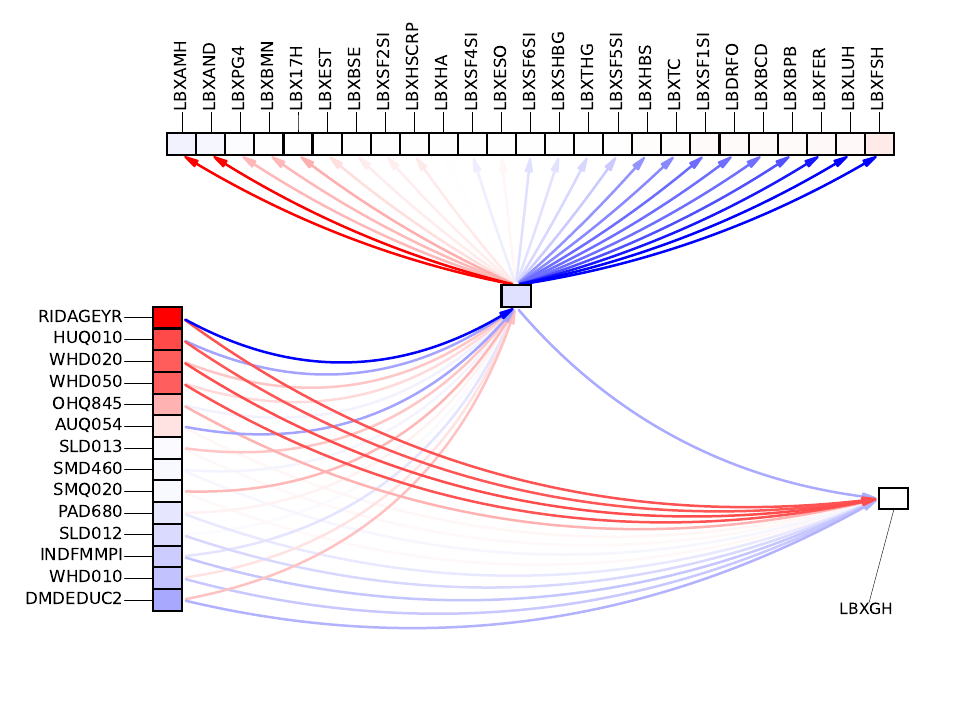}
    \includegraphics[width=0.45\textwidth]{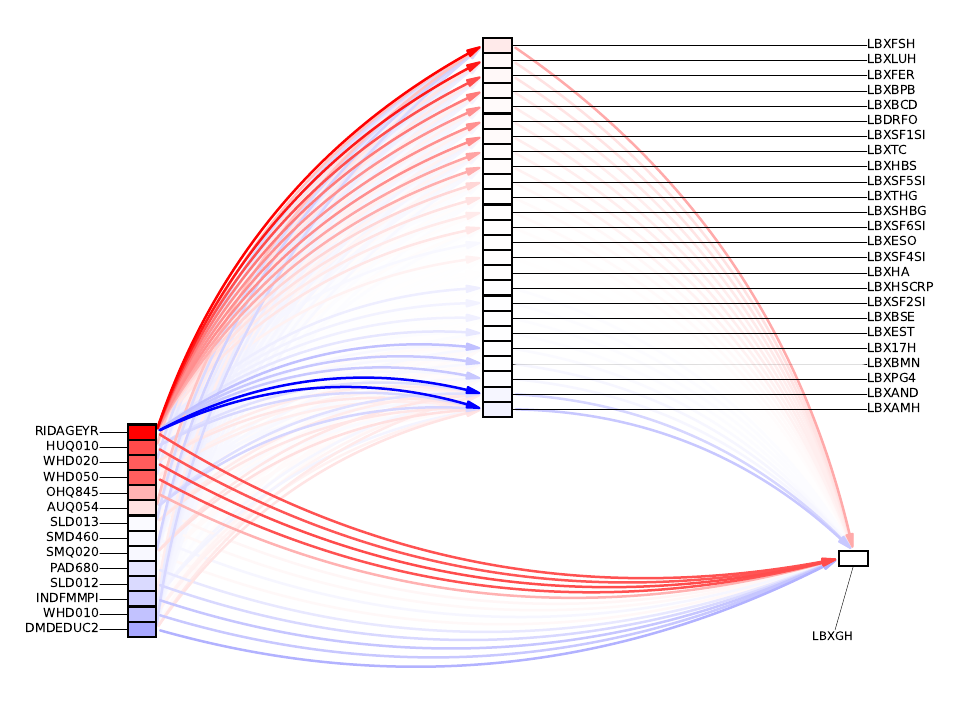}
    \caption{Full graphs for the NHANES dataset}
\end{figure}
\begin{figure}[H]
    \centering
    \includegraphics[width=0.45\textwidth]{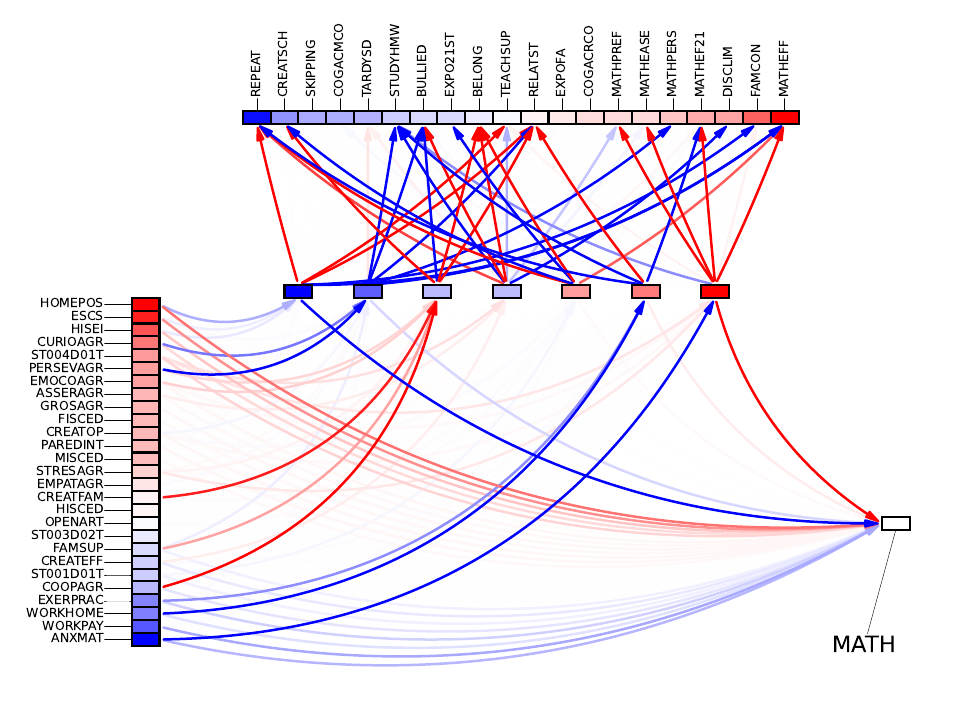}
    \includegraphics[width=0.45\textwidth]{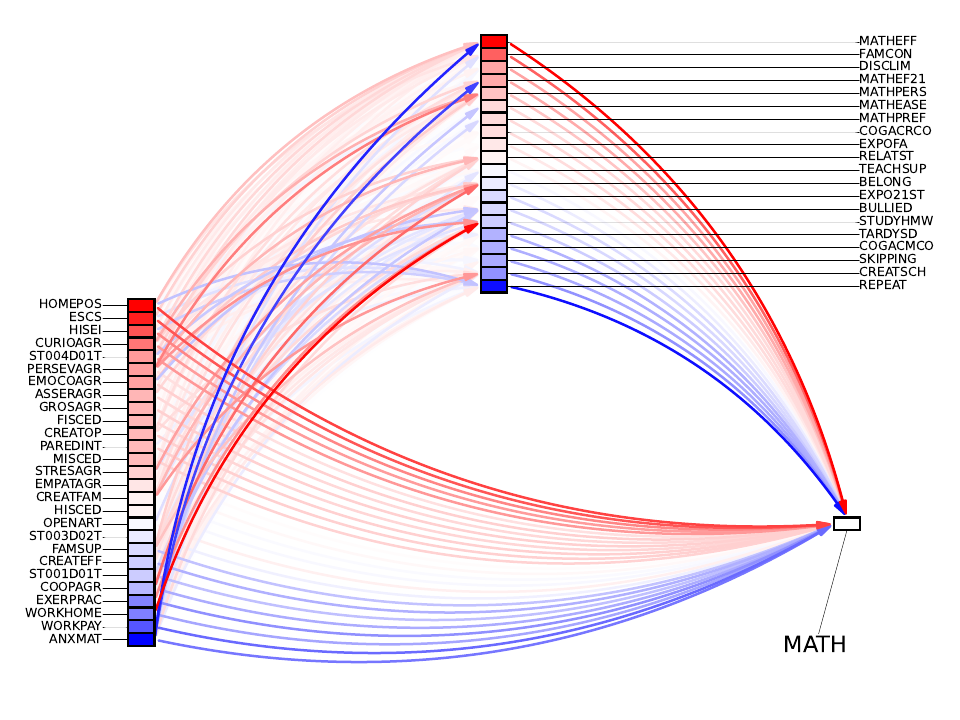}
    \caption{Full graphs for the PISA-I dataset}
\end{figure}
\begin{figure}[H]
    \centering
    \includegraphics[width=0.45\textwidth]{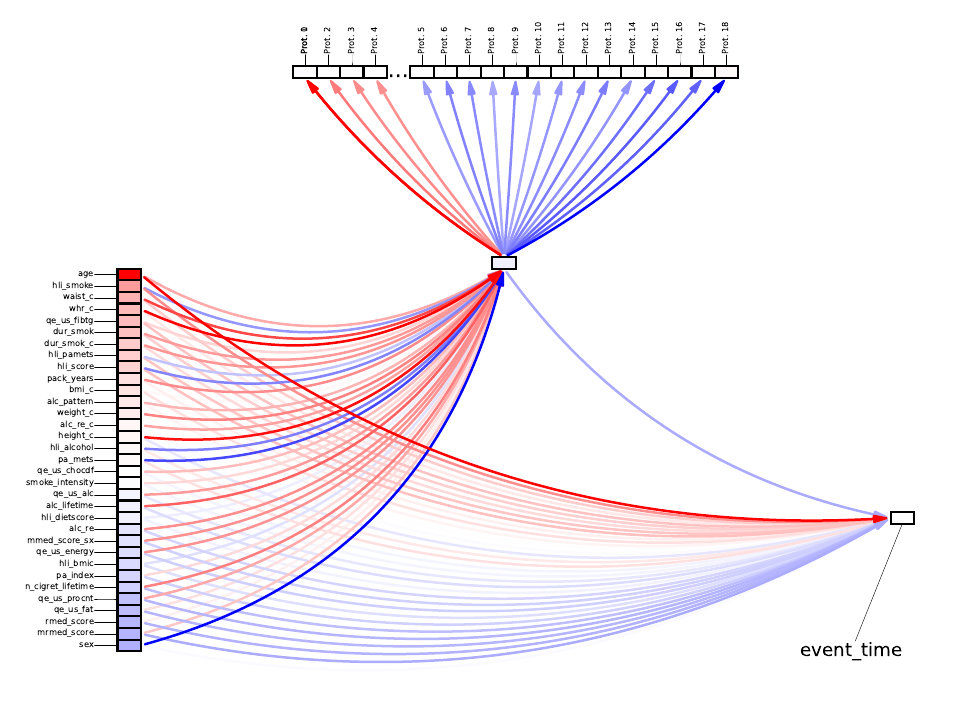}
    \includegraphics[width=0.45\textwidth]{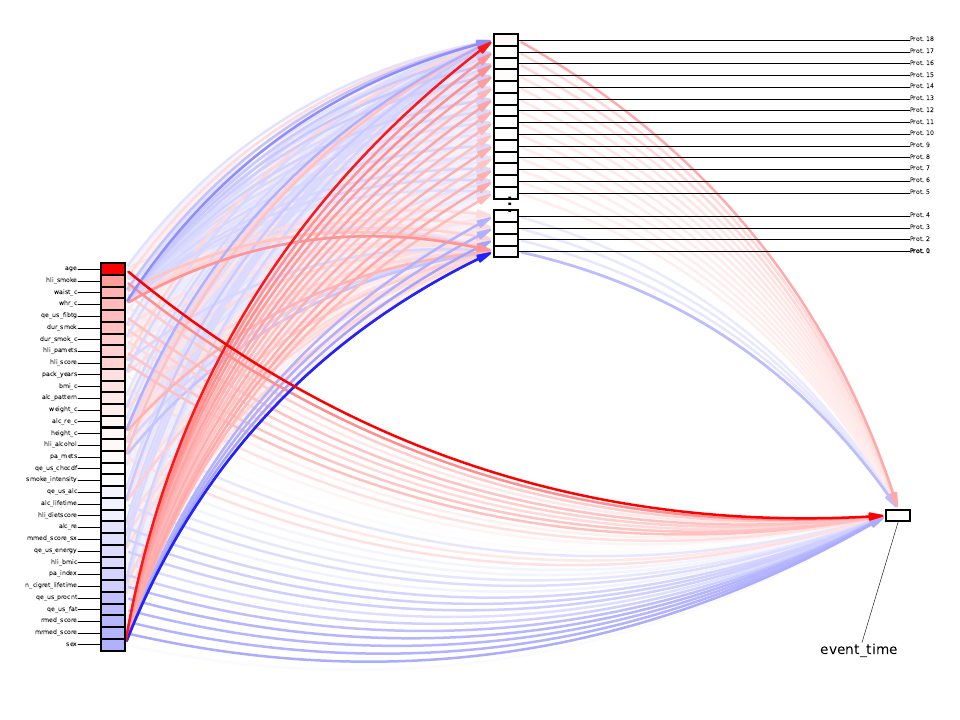}
    \caption{Full graphs for the EPIC-CAN dataset}
\end{figure}

\section{Real-world data labels}
List of all source variables considered in the proposed real-world datasets. Not all variables are being used in the main experiments, some being removed to avoid collinearity, reduce noise and/or due to irrelevance regarding the task at hand.
\label{sec-SM:RW-labels}
\subsection{PISA-I}
\subsubsection{X}
\textbf{ST003D02T}~=~Birth month 1-12
, \textbf{ST004D01T}~=~Gender 1-2
, \textbf{ST001D01T}~=~class level
, \textbf{STUBMI}~=~Body Mass Index (BMI)
, \textbf{MISCED}~=~Mother's level of education (ISCED)
, \textbf{FISCED}~=~Father's level of education (ISCED)
, \textbf{HISCED}~=~Highest level of education of parents (ISCED)
, \textbf{PAREDINT}~=~Index highest parental education (international years of schooling scale)
, \textbf{BMMJ1}~=~Mother's occupational status (ISEI) based on 4-digit human coded ISCO
, \textbf{BFMJ2}~=~Father's occupational status (ISEI) based on 4-digit human coded ISCO
, \textbf{HISEI}~=~Highest parental occupational status (ISEI) based on 4-digit human coded ISCO
, \textbf{ESCS}~=~Index of economic, social and cultural status
, \textbf{FAMSUP}~=~Family support (WLE)
, \textbf{FAMSUPSL}~=~Family support for self-directed learning (WLE)
, \textbf{CURSUPP}~=~Current parental/guardian Support (WLE)
, \textbf{SOCCON}~=~Social Connections: Ease of Communication About Worries and Concerns (WLE)
, \textbf{SOCONPA}~=~Social connection to Parents (WLE)
, \textbf{ICTRES}~=~ICT Resources (WLE)
, \textbf{HOMEPOS}~=~Home possessions (WLE)
, \textbf{ICTHOME}~=~ICT availability outside of school  (WLE)
, \textbf{ICTAVHOM}~=~Availability and Usage of ICT at Home
, \textbf{ICTQUAL}~=~Quality of access to ICT (WLE)
, \textbf{ICTEFFIC}~=~Self-efficacy in digital competencies (WLE)
, \textbf{ICTWKDY}~=~Frequency of ICT activity (weekday) (WLE)
, \textbf{ICTWKEND}~=~Frequency of ICT activity (weekend) (WLE)
, \textbf{PARINVOL}~=~Parental Involvement (WLE)
, \textbf{PQMIMP}~=~Parent Attitudes Toward Mathematics (WLE)
, \textbf{PQMCAR}~=~Mathematics Career (WLE)
, \textbf{PAREXPT}~=~Parent's expectations in child's career
, \textbf{ATTIMMP}~=~Parents' Attitudes towards Immigrants (WLE)
, \textbf{EXERPRAC}~=~Exercise or practice a sport before or after school
, \textbf{WORKPAY}~=~Working for pay before or after school
, \textbf{WORKHOME}~=~Working in household/take care of family members before or after school
, \textbf{PERSEVAGR}~=~Perseverance (agreement) (WLE)
, \textbf{CURIOAGR}~=~Curiosity (agreement) (WLE)
, \textbf{COOPAGR}~=~Cooperation (agreement) (WLE)
, \textbf{EMPATAGR}~=~Empathy (agreement) (WLE)
, \textbf{ASSERAGR}~=~Assertiveness (agreement) (WLE)
, \textbf{STRESAGR}~=~Stress resistance (agreement) (WLE)
, \textbf{EMOCOAGR}~=~Emotional control (agreement) (WLE)
, \textbf{GROSAGR}~=~Growth Mindset (WLE)
, \textbf{INFOSEEK}~=~Information seeking about future careers (WLE)
, \textbf{PROBSELF}~=~Problems with self-directed learning (WLE)
, \textbf{FEELLAH}~=~Feelings about learning at home (WLE)
, \textbf{SDLEFF}~=~Self-directed learning self-efficacy (WLE)
, \textbf{ACCESSFP}~=~Access to money and financial products, sources of money (WLE)
, \textbf{FLFAMILY}~=~Parental involvement in matters of FL (WLE)
, \textbf{FLCONFIN}~=~Confidence about financial matters (WLE)
, \textbf{FLCONICT}~=~Confidence about financial matters using digital devices (WLE)
, \textbf{ACCESSFA}~=~Access to money and financial products, financial activities (WLE)
, \textbf{ATTCONFM}~=~Attitudes towards and confidence about financial matters (WLE)
, \textbf{FRINFLFM}~=~Friends' influence on financial matters (WLE)
, \textbf{FCFMLRTY}~=~Familiarity with concepts of finance
, \textbf{OPENART}~=~Openness to Art and Reflection TBD (WLE)
, \textbf{IMAGINE}~=~Imagination and Adventurousness TBD (WLE)
, \textbf{CREATHME}~=~Creative Home Environment (WLE)
, \textbf{CREATACT}~=~Participation in creative activities outside of school (WLE)
, \textbf{CREATOPN}~=~Creativity and Openness to Intellect (WLE)
, \textbf{CREATOR}~=~Openness to creativity: Other's report (WLE)
, \textbf{CREATEFF}~=~Creative self-efficacy (WLE)
, \textbf{CREATFAM}~=~Creative peers and family environment (WLE)
, \textbf{CREATOOS}~=~Creative Activities outside of school (WLE)
, \textbf{CREATOP}~=~Creativity and Openness to Intellect TBD (WLE)
, \textbf{ICTDISTR}~=~Distress from online content and cyberbullying
, \textbf{ANXMAT}~=~Mathematics Anxiety (WLE)
, \textbf{BODYIMA}~=~Body Image (WLE)
, \textbf{LIFESAT}~=~Students' Life Satisfaction across Domains (WLE)
, \textbf{PSYCHSYM}~=~Psychosomatic Symptoms (WLE)
, \textbf{EXPWB}~=~Experienced Well-being (Previous Day) (WLE)

\subsubsection{M}
\textbf{REPEAT}~=~Grade repetition
, \textbf{MISSSC}~=~Missing school for more than 3 months
, \textbf{SKIPPING}~=~Skipping classes or days of school
, \textbf{TARDYSD}~=~Arriving late for school stricter definition
, \textbf{DURECEC}~=~Duration in early childhood education and care
, \textbf{STUDYHMW}~=~Studying for school or homework before or after school
, \textbf{EXPECEDU}~=~Highest expected educational level
, \textbf{RELATST}~=~Quality of student-teacher relationships (WLE)
, \textbf{BELONG}~=~Sense of belonging (WLE)
, \textbf{BULLIED}~=~Being bullied (WLE)
, \textbf{FEELSAFE}~=~Feeling safe (WLE)
, \textbf{SCHRISK}~=~School safety risks (WLE)
, \textbf{SCHSUST}~=~School actions/activities to sustain learning  (WLE)
, \textbf{LEARRES}~=~Types of learning resources used while school was closed (WLE)
, \textbf{ICTSCH}~=~ICT availability at school  (WLE)
, \textbf{PQSCHOOL}~=~School quality (WLE)
, \textbf{ICTAVSCH}~=~Availability and Usage of ICT at School
, \textbf{PASCHPOL}~=~School Policies for Parental Involvement  (WLE)
, \textbf{MATHPREF}~=~Prefers maths
, \textbf{MATHEASE}~=~Ease at maths
, \textbf{MATHMOT}~=~Motivation in maths
, \textbf{CREATSCH}~=~Creative school and class environment (WLE)
, \textbf{CREATAS}~=~Creative Activities at school (WLE)
, \textbf{ICTSUBJ}~=~Subject-related ICT Use During Lessons (WLE)
, \textbf{ICTENQ}~=~Use of ICT in enquiry-based learning activities (WLE)
, \textbf{ICTFEED}~=~Support or feedback via ICT  (WLE)
, \textbf{ICTOUT}~=~Use of ICT for school activities outside of the classroom (WLE)
, \textbf{ICTREG}~=~Views of regulated ICT use in school (WLE)
, \textbf{ICTINFO}~=~Students' practices regarding online information (WLE)
, \textbf{FLSCHOOL}~=~Financial education in school lessons (WLE)
, \textbf{FLMULTSB}~=~Financial education in school lessons, multiple subjects (WLE)
, \textbf{COGACRCO}~=~Cognitive activation in mathematics: Foster reasoning Version B (WLE)
, \textbf{COGACMCO}~=~Cognitive activation in mathematics: Encourage mathematical thinking Version B (WLE)
, \textbf{EXPOFA}~=~Exposure to Formal and Applied Mathematics Tasks (WLE)
, \textbf{EXPO21ST}~=~Exposure to Mathematical Reasoning and 21st century mathematics tasks (WLE)
, \textbf{MATHEFF}~=~Mathematics self-efficacy: formal and applied mathematics - response options reversed in 2022 (WLE)
, \textbf{MATHEF21}~=~Mathematics self-efficacy: mathematical reasoning and 21st century skills (WLE)
, \textbf{FAMCON}~=~Subjective familiarity with mathematics concepts (WLE)
, \textbf{MATHPERS}~=~Effort and Persistence in Mathematics  (WLE)
, \textbf{TEACHSUP}~=~Mathematics Teacher Support (WLE)
, \textbf{DISCLIM}~=~Disciplinary climate in mathematics (WLE)

\subsubsection{Y}
\textbf{MATH}~=~Mathematics proficiency score

\subsection{NHANES}
\subsubsection{X}
\textbf{RIAGENDR}~=~Gender
, \textbf{RIDAGEYR}~=~Age at screening
, \textbf{DMDEDUC2}~=~Education level [1-5]
, \textbf{AUQ054}~=~General condition of hearing (audiometry) [1-6] from [excellent-deaf]
, \textbf{ALQ130}~=~Avg number alcoholic drinks/day/past 12 mos [1-15]
, \textbf{ALQ280}~=~Number times 8+ drinks in 1 day/past 12 mos
, \textbf{DEQ034A}~=~Stay in the shade when sunny outside [1-5] from [always-never] in the shadows
, \textbf{DEQ034C}~=~Wear a long sleeved shirt [1-5] from [always-never]
, \textbf{DEQ034D}~=~Use sunscreen [1-5] from [always-never]
, \textbf{ECD070A}~=~Weight at birth, pounds
, \textbf{HUQ010}~=~General health condition [1-5] from [good-poor]
, \textbf{HOD051}~=~Number of rooms in home
, \textbf{INDFMMPI}~=~Family monthly poverty level index
, \textbf{KIQ481}~=~How many times wake up and urinate?
, \textbf{OCQ180}~=~Number of hours worked in the last week
, \textbf{OHQ845}~=~Rate the health of your teeth and gums [1-5] from [good-poor]
, \textbf{PAD680}~=~Minutes sedentary activity/day (sitting at school, work, bus, etc)
, \textbf{RXQ050}~=~Number of prescription medicines taken last 30 days [1-5]
, \textbf{SLD012}~=~Sleep hours - weekdays or workdays
, \textbf{SLD013}~=~Sleep hours - weekends
, \textbf{SMQ020}~=~Smoked at least 100 cigarettes in life
, \textbf{SMD460}~=~Number of people who live here smoke tobacco? [0-2]
, \textbf{WHD010}~=~Current self-reported height (inches) [40-83]
, \textbf{WHD020}~=~Current self-reported weight (pounds) [60-530]
, \textbf{WHD050}~=~Self-reported weight - 1 yr ago (pounds) [60-530]
, \textbf{FNQ450}~=~>17yo - Difficulty communicating [1-4] from [no difficulty-cannot at all]
, \textbf{FNQ460}~=~>17yo - Difficulty remembering things/concentrating [1-4] from [no difficulty-cannot at all]
, \textbf{FNQ530}~=~>17yo - How often very sad/depressed [1-5] from [daily-never]

\subsubsection{M}
\textbf{LBXAGP}~=~alpha-1-acid glycoprotein (g/L)
, \textbf{LBXTC}~=~Total Cholesterol (mg/dL)
, \textbf{LBXFER}~=~Ferritin(ng/mL)
, \textbf{LBDRFO}~=~RBC folate (ng/mL)
, \textbf{LBXHSCRP}~=~HS C-Reactive Protein (mg/L) - Linked to risk of CVD
, \textbf{LBXIN}~=~Insulin (uU/mL)
, \textbf{LBXBPB}~=~Blood lead (ug/dL)
, \textbf{LBXBCD}~=~Blood cadmium (ug/L)
, \textbf{LBXTHG}~=~Blood mercury, total (ug/L)
, \textbf{LBXBSE}~=~Blood selenium (ug/L)
, \textbf{LBXBMN}~=~Blood manganese (ug/L)
, \textbf{LBXSF1SI}~=~5-Methyl-tetrahydrofolate (nmol/L)
, \textbf{LBXSF2SI}~=~Folic acid (nmol/L)
, \textbf{LBXSF3SI}~=~5-Formyl-tetrahydrofolate (nmol/L)
, \textbf{LBXSF4SI}~=~Tetrahydrofolate (nmol/L)
, \textbf{LBXSF5SI}~=~5,10-Methenyl-tetrahydrofolate (nmol/L)
, \textbf{LBXSF6SI}~=~Mefox oxidation product (nmol/L)
, \textbf{LBX17H}~=~17alpha-hydroxyprogesterone (ng/dL)
, \textbf{LBXAND}~=~Androstenedione (ng/dL)
, \textbf{LBXAMH}~=~Anti-Mullerian hormone (ng/mL)
, \textbf{LBXDHE}~=~DHEAS (µg/dL)
, \textbf{LBXEST}~=~Estradiol (pg/mL)
, \textbf{LBXESO}~=~Estrone (ng/dL)
, \textbf{LBXFSH}~=~Follicle Stimulating Hormone (mIU/mL)
, \textbf{LBXLUH}~=~Luteinizing Hormone (mIU/mL)
, \textbf{LBXPG4}~=~Progesterone (ng/dL)
, \textbf{LBXTST}~=~Testosterone, total (ng/dL)
, \textbf{LBXSHBG}~=~SHBG (nmol/L)
, \textbf{LBXTFR}~=~Transferrin receptor (mg/L)
, \textbf{LBXVIDMS}~=~25OHD2+25OHD3 (nmol/L)
, \textbf{LBXHA}~=~Hepatitis A antibody [1,2] for [positive,negative]
, \textbf{LBXHBS}~=~Hepatitis B surface antibody [1,2] for [positive,negative]
, \textbf{LBXGLU}~=~Fasting Blood/Plasma Glucose (mg/dL)

\subsubsection{Y}
\textbf{LBXGH}~=~Glycohemoglobin (\%) - Glucose in blood (\%)

\subsection{EPIC-CAN}
Acronyms: whr: waist to height ratio, rmed: Mediterranean diet, alc: alcohol, \_c: category, pa: physical activity, aggr: aggregated, qe\_us: USDA, hli: Healthy Lifestyle Index

\subsubsection{X}
\textbf{age}
, \textbf{sex}
, \textbf{whr\_c}
, \textbf{height\_c}
, \textbf{weight\_c}
, \textbf{waist\_c}
, \textbf{bmi\_c}
, \textbf{l\_school}
, \textbf{mrmed\_score}
, \textbf{mmed\_score\_sx}
, \textbf{rmed\_score}
, \textbf{mrmed\_score}
, \textbf{pa\_mets}
, \textbf{pa\_index}
, \textbf{alc\_re}
, \textbf{alc\_re\_c}
, \textbf{alc\_lifetime}
, \textbf{alc\_pattern}
, \textbf{dur\_smok}
, \textbf{dur\_smok\_c}
, \textbf{n\_cigret\_aggr}
, \textbf{n\_cigret\_lifetime}
, \textbf{smoke\_intensity}
, \textbf{pack\_years}
, \textbf{qe\_us\_alc}
, \textbf{qe\_us\_chocdf}
, \textbf{qe\_us\_energy}
, \textbf{qe\_us\_fat}
, \textbf{qe\_us\_fibtg}
, \textbf{qe\_us\_procnt}
, \textbf{hli\_dietscore}
, \textbf{hli\_pamets}
, \textbf{hli\_smoke}
, \textbf{hli\_alcohol}
, \textbf{hli\_bmic}
, \textbf{hli\_score}
, \textbf{systol}
, \textbf{diastol}

\subsubsection{M}
Protein identifiers

\subsubsection{Y}
Age at cancer event

\subsection{CRIME}
\subsubsection{X}
\textbf{population} population for community
, \textbf{householdsize} mean people per household
, \textbf{racepctblack} percentage of population that is african american
, \textbf{racePctWhite} percentage of population that is caucasian
, \textbf{racePctAsian} percentage of population that is of asian heritage
, \textbf{racePctHisp} percentage of population that is of hispanic heritage
, \textbf{agePct12t21} percentage of population that is 12-21 in age
, \textbf{agePct12t29} percentage of population that is 12-29 in age
, \textbf{agePct16t24} percentage of population that is 16-24 in age
, \textbf{agePct65up} percentage of population that is 65 and over in age
, \textbf{numbUrban} number of people living in areas classified as urban
, \textbf{pctUrban} percentage of people living in areas classified as urban
, \textbf{PctForeignBorn} percent of people foreign born
, \textbf{PctBornSameState} percent of people born in the same state as currently living
, \textbf{PctSameHouse85} percent of people living in the same house as in 1985 (5 years before)
, \textbf{PctSameCity85} percent of people living in the same city as in 1985 (5 years before)
, \textbf{PctSameState85} percent of people living in the same state as in 1985 (5 years before)
, \textbf{MalePctDivorce} percentage of males who are divorced
, \textbf{MalePctNevMarr} percentage of males who have never married
, \textbf{FemalePctDiv} percentage of females who are divorced
, \textbf{TotalPctDiv} percentage of population who are divorced
, \textbf{PersPerFam} mean number of people per family
, \textbf{PctFam2Par} percentage of families (with kids) that are headed by two parents
, \textbf{PctKids2Par} percentage of kids in family housing with two parents
, \textbf{PctYoungKids2Par} percent of kids 4 and under in two parent households
, \textbf{PctTeen2Par} percent of kids age 12-17 in two parent households
, \textbf{PctWorkMomYoungKids} percentage of moms of kids 6 and under in labor force
, \textbf{PctWorkMom} percentage of moms of kids under 18 in labor force
, \textbf{NumIlleg} number of kids born to never married
, \textbf{PctIlleg} percentage of kids born to never married
, \textbf{NumImmig} total number of people known to be foreign born
, \textbf{PctImmigRecent} percentage of immigrants who immigrated within last 3 years
, \textbf{PctImmigRec5} percentage of immigrants who immigated within last 5 years
, \textbf{PctImmigRec8} percentage of immigrants who immigated within last 8 years
, \textbf{PctImmigRec10} percentage of immigrants who immigated within last 10 years
, \textbf{PctRecentImmig} percent of population who have immigrated within the last 3 years
, \textbf{PctRecImmig5} percent of population who have immigrated within the last 5 years
, \textbf{PctRecImmig8} percent of population who have immigrated within the last 8 years
, \textbf{PctRecImmig10} percent of population who have immigrated within the last 10 years
, \textbf{PctSpeakEnglOnly} percent of people who speak only English
, \textbf{PctNotSpeakEnglWell} percent of people who do not speak English well
, \textbf{PctLess9thGrade} percentage of people 25 and over with less than a 9th grade education
, \textbf{PctNotHSGrad} percentage of people 25 and over that are not high school graduates
, \textbf{PctBSorMore} percentage of people 25 and over with a bachelor's degree or higher education
, \textbf{medIncome} median household income
, \textbf{pctWWage} percentage of households with wage or salary income in 1989
, \textbf{pctWFarmSelf} percentage of households with farm or self employment income in 1989
, \textbf{pctWInvInc} percentage of households with investment / rent income in 1989
, \textbf{pctWSocSec} percentage of households with social security income in 1989
, \textbf{pctWPubAsst} percentage of households with public assistance income in 1989
, \textbf{pctWRetire} percentage of households with retirement income in 1989
, \textbf{medFamInc} median family income (differs from household income for non-family households)
, \textbf{perCapInc} per capita income
, \textbf{whitePerCap} per capita income for caucasians
, \textbf{blackPerCap} per capita income for african americans
, \textbf{indianPerCap} per capita income for native americans
, \textbf{AsianPerCap} per capita income for people with asian heritage
, \textbf{OtherPerCap} per capita income for people with 'other' heritage
, \textbf{HispPerCap} per capita income for people with hispanic heritage
, \textbf{NumUnderPov} number of people under the poverty level
, \textbf{PctPopUnderPov} percentage of people under the poverty level
, \textbf{PctUnemployed} percentage of people 16 and over, in the labor force, and unemployed
, \textbf{PctEmploy} percentage of people 16 and over who are employed
, \textbf{PctEmplManu} percentage of people 16 and over who are employed in manufacturing
, \textbf{PctEmplProfServ} percentage of people 16 and over who are employed in professional services
, \textbf{PctOccupManu} percentage of people 16 and over who are employed in manufacturing
, \textbf{PctOccupMgmtProf} percentage of people 16 and over who are employed in management or professional occupations

\subsubsection{M}
\textbf{LemasSwornFT} number of sworn full time police officers
, \textbf{LemasSwFTPerPop} sworn full time police officers per 100K population
, \textbf{LemasSwFTFieldOps} number of sworn full time police officers in field operations (on the street as opposed to administrative etc)
, \textbf{LemasSwFTFieldPerPop} sworn full time police officers in field operations (on the street as opposed to administrative etc) per 100K population
, \textbf{LemasTotalReq} total requests for police
, \textbf{LemasTotReqPerPop} total requests for police per 100K population
, \textbf{PolicReqPerOffic} total requests for police per police officer
, \textbf{PolicPerPop} police officers per 100K population
, \textbf{RacialMatchCommPol} a measure of the racial match between the community and the police force. High values indicate proportions in community and police force are similar
, \textbf{PctPolicWhite} percent of police that are caucasian
, \textbf{PctPolicBlack} percent of police that are african american
, \textbf{PctPolicHisp} percent of police that are hispanic
, \textbf{PctPolicAsian} percent of police that are asian
, \textbf{PctPolicMinor} percent of police that are minority of any kind
, \textbf{OfficAssgnDrugUnits} number of officers assigned to special drug units
, \textbf{NumKindsDrugsSeiz} number of different kinds of drugs seized
, \textbf{PolicAveOTWorked} police average overtime worked  
, \textbf{PolicCars} number of police cars
, \textbf{PolicOperBudg} police operating budget
, \textbf{LemasPctPolicOnPatr} percent of sworn full time police officers on patrol
, \textbf{LemasGangUnitDeploy} gang unit deployed (numeric - decimal - but really ordinal - 0 means NO, 1 means YES,  0.5 means Part Time)
, \textbf{LemasPctOfficDrugUn} percent of officers assigned to drug units
, \textbf{PolicBudgPerPop} police operating budget per population
, \textbf{PctLargHouseFam} percent of family households that are large (6 or more)
, \textbf{PctLargHouseOccup} percent of all occupied households that are large (6 or more people)
, \textbf{PersPerOccupHous} mean persons per household
, \textbf{PersPerOwnOccHous} mean persons per owner occupied household
, \textbf{PersPerRentOccHous} mean persons per rental household
, \textbf{PctPersOwnOccup} percent of people in owner occupied households
, \textbf{PctPersDenseHous} percent of persons in dense housing (more than 1 person per room)
, \textbf{PctHousLess3BR} percent of housing units with less than 3 bedrooms
, \textbf{MedNumBR} median number of bedrooms
, \textbf{HousVacant} number of vacant households
, \textbf{PctHousOccup} percent of housing occupied
, \textbf{PctHousOwnOcc} percent of households owner occupied
, \textbf{PctVacantBoarded} percent of vacant housing that is boarded up
, \textbf{PctVacMore6Mos} percent of vacant housing that has been vacant more than 6 months
, \textbf{MedYrHousBuilt} median year housing units built
, \textbf{PctHousNoPhone} percent of occupied housing units without phone (in 1990, this was rare!)
, \textbf{PctWOFullPlumb} percent of housing without complete plumbing facilities
, \textbf{OwnOccLowQuart} owner occupied housing - lower quartile value
, \textbf{OwnOccMedVal} owner occupied housing - median value
, \textbf{OwnOccHiQuart} owner occupied housing - upper quartile value
, \textbf{RentLowQ} rental housing - lower quartile rent
, \textbf{RentMedian} rental housing - median rent (Census variable H32B from file STF1A)
, \textbf{RentHighQ} rental housing - upper quartile rent
, \textbf{MedRent} median gross rent (Census variable H43A from file STF3A - includes utilities)
, \textbf{MedRentPctHousInc} median gross rent as a percentage of household income
, \textbf{MedOwnCostPctInc} median owners cost as a percentage of household income - for owners with a mortgage
, \textbf{MedOwnCostPctIncNoMtg} median owners cost as a percentage of household income - for owners without a mortgage
, \textbf{NumInShelters} number of people in homeless shelters
, \textbf{NumStreet} number of homeless people counted in the street
, \textbf{LandArea} land area in square miles
, \textbf{PopDens} population density in persons per square mile
, \textbf{PctUsePubTrans} percent of people using public transit for commuting

\subsubsection{Y}
\textbf{ViolentCrimesPerPop} total number of violent crimes per 100K population GOAL attribute (to be predicted)

\end{document}